\documentclass[10pt, letterpaper, twoside]{article} 

\usepackage{amsmath, amsthm, amssymb,bbold,mathrsfs} 
\usepackage{graphicx,hyperref}
\usepackage[toc,page]{appendix}
\usepackage{enumitem}
\setlist[enumerate]{topsep=1pt,itemsep=-0.5pt,parsep=2pt}
\usepackage[numbers]{natbib}
\usepackage{slashbox,caption,subcaption,rotating}
\usepackage{geometry,pdflscape}

\def\re{\mathbb{R}} 
 
\def\e{e}
\def\F{F}

\def\i{i}

\def\1{\mathbf{1}}
\def\0{\mathbf{0}}

\def\E{\mathbb{E}}

\def\H{B}

\def\old#1{}

\begin{document} 

\title{Some Simulation Results for Emphatic Temporal-Difference Learning Algorithms\thanks{This note uses colors in most of the figures to distinguish between different algorithms. Therefore it is better to display the contents on a computer screen than to print them in black and white.}}

\author{Huizhen Yu\thanks{RLAI Lab, Department of Computing Science, University of Alberta, Canada (\texttt{janey.hzyu@gmail.com}). This research was supported by a grant from Alberta Innovates -- Technology Futures.}}
\date{}
\maketitle

\begin{abstract}
This is a companion note to our recent study of the weak convergence properties of constrained emphatic temporal-difference learning (ETD) algorithms from a theoretic perspective.
It supplements the latter analysis with simulation results and illustrates the behavior of some of the ETD algorithms using three example problems.
 \end{abstract}

\bigskip
\bigskip
\bigskip

\tableofcontents

\newpage
\section{About this Note}

This is a companion note to our recent study of the weak convergence properties of constrained ETD algorithms from a theoretical perspective \cite{etd-wkconv}.  
Our purpose here is to supplement that theoretical analysis with simulation results, and to illustrate the behavior of some of the ETD algorithms using examples.

We will consider three test problems: two small grid world-like problems and then the larger Mountain Car problem. 
As to the algorithms, we will focus on the two variant algorithms in \cite{etd-wkconv} (given by Eqs.~(3.3) and (3.4) respectively in Section 3.2 of \cite{etd-wkconv}), as well as their perturbed versions for the constant-stepsize case (given by Eq.~(3.7) in Section 3.3 of \cite{etd-wkconv}).
These algorithms are constrained ETD algorithms that have biases (cf.\ the discussion in Section 3.2 of \cite{etd-wkconv}), but they are more robust than the unbiased algorithms in practice, as we will also explain later in Section~\ref{sec-2prob} of this note.
We will refer to the two variant algorithms as Variant I and Variant II below.

In what follows, we first describe the two small test problems and illustrate the behavior of the trace iterates (Section~\ref{sec-2prob}). 
We then show simulation results of the constrained ETD algorithms just mentioned, for the case of constant stepsize (Section~\ref{sec-conststp}) and for the case of diminishing stepsize (Section~\ref{sec-dimstp}). We use these results in particular to demonstrate some of the convergence properties proved in \cite{etd-wkconv}, and to show that the algorithms are well-behaved despite the high variance issue in off-policy learning. 
Finally, we show simulation results on the Mountain Car example for a chosen target policy (Section~\ref{sec-mountaincar}). This is to demonstrate that ETD can be applied beyond small test problems and is a useful method for off-policy learning.

Before proceeding, we would like to clarify that we do not intend this note to be a stand-alone paper. 
We will thus use the notation given in \cite{etd-wkconv} without redefining it here.
We will also include very few references -- only those needed in order to clarify some experimental setup or results. 
(Please see the paper \cite{etd-wkconv} for important prior works on TD and ETD learning.)  
 
We would also like to mention that we use colors in most of the figures to distinguish between the iterates produced by different algorithms. 
Therefore it is better to view the contents of this note on a computer screen than to have them printed out in black and white.

\section{Two Test Problems} \label{sec-2prob}

We now describe two test problems used in our experiments. For these two problems, it is simpler to describe the system dynamics directly in terms of the state transition probabilities, without dealing with actions explicitly. So this is what we are going to do below. (Readers who wish to make the action space explicit may interpret each state transition in our description below as being caused by a distinct action that results in that particular transition with certainty.)

\begin{figure}[thb]
   \centering
   \includegraphics[width=0.27\linewidth]{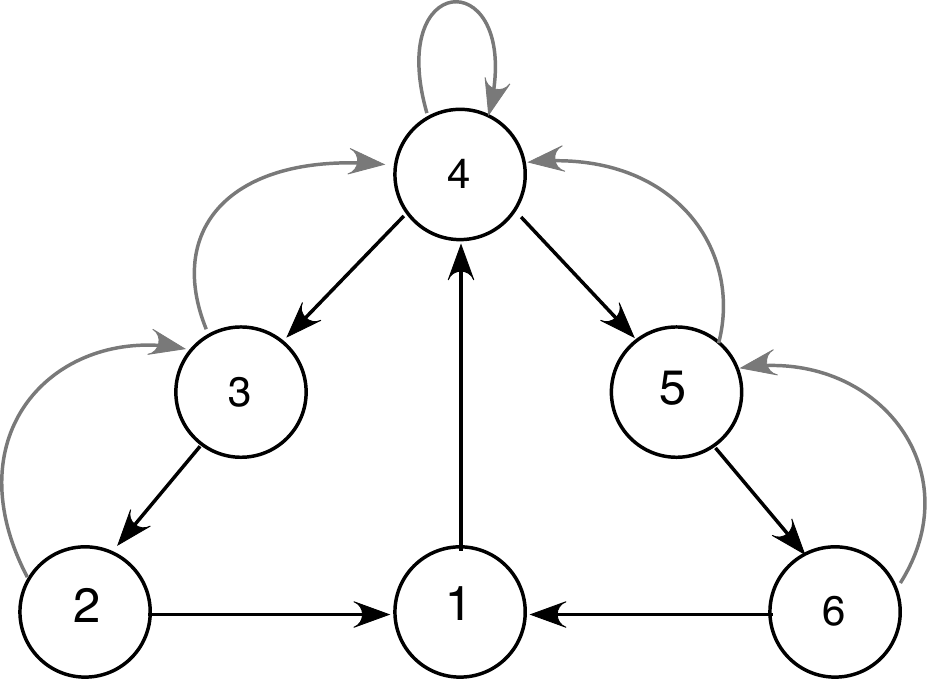} \qquad \qquad \qquad \quad
   \includegraphics[width=0.3\linewidth]{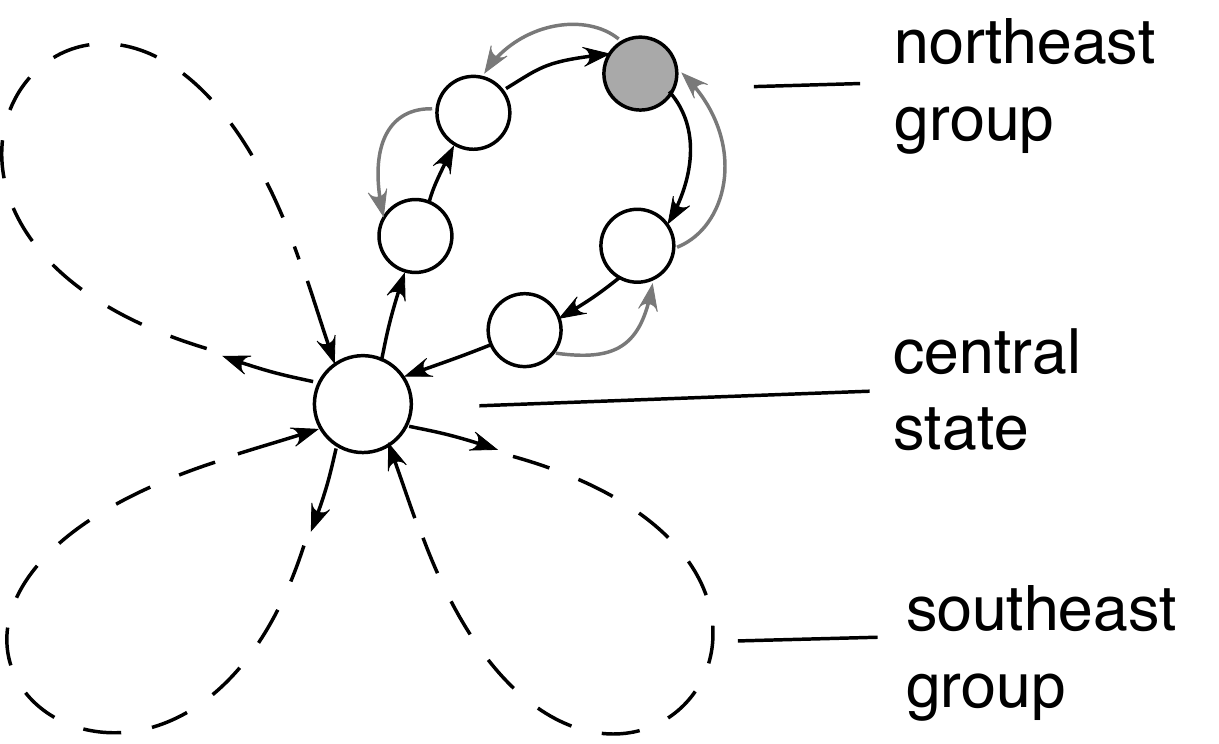}\\*[5pt]
   \caption{The transition graphs of two test problems.}
   \label{fig-trgrph}
\end{figure}

\medskip
\noindent{\bf Problem I:}
Problem I has $6$ states. Let $P_{\pi}$ and $P_{\pi^o}$ be the state transition probability matrices under the target policy $\pi$ and behavior policy $\pi^o$, respectively. These transition matrices are given by
$$ P_{\pi} = \left( 
\begin{array}{cccccc}
  0  &  0 & 0  &  1  & 0  &  0 \\
  0.9 & 0  & 0.1 & 0  & 0  &  0\\
     0 & 0.9 & 0  & 0.1  & 0  & 0\\
     0  & 0  & 0.2  & 0.3  & 0.5 & 0\\
     0  & 0 &  0  & 0.1 & 0  & 0.9\\
     0.9 & 0 &  0 & 0  & 0.1  & 0 
     \end{array} \right), 
     \qquad P_{\pi^o} = \left(
     \begin{array}{cccccc} 
     0  &  0  & 0  & 1 & 0  &  0\\    
     0.5 & 0  & 0.5 & 0  & 0  & 0\\
     0  & 0.5 & 0  & 0.5  & 0  & 0\\
     0  & 0  & 0.4  & 0.2  & 0.4  & 0\\   
      0  & 0  & 0 & 0.5 & 0  & 0.5\\
     0.5  & 0  & 0 & 0 & 0.5 & 0 \end{array} 
     \right).
$$     
Their associated transition graphs have the same topology, which is drawn in Figure~\ref{fig-trgrph} (left).  
The transition from state $6$ to $1$ has reward $1$; all the other transitions have reward zero. 

The rest of the parameters are defined as follows. The discount factors $\gamma(s)$ are state-dependent: $\gamma(1)=0.7$ and $\gamma(s) = 1$ for $s > 1$. 
The interest weights $\i(s)$ and the $\lambda$-parameters are also state-dependent: $\i(s) = 1$ for $s \in \{2, 4, 6\}$ and $\i(s) = 0$ otherwise, $\lambda(s) = 0$ for $s \in \{2, 4, 6\}$ and $\lambda(s)=1$ otherwise. Aggregating states into $3$ groups, $\{1, 4\}$, $\{2,3\}$, and $\{5, 6\}$, we assign $3$ binary features to each state to indicate its membership.

We remark that with the above choices of $\i$ and $\lambda$, the approximate value function $\phi(s)^\top \theta^*$ of ETD equals exactly the true value function $v_{\pi}(s)$ at $s \in \{2, 4, 6\}$, the three states of interest. This serves as an example to show how one can define state-dependent $\i$ and $\lambda$ jointly so that accurate estimates of $v_\pi(s)$ for desired states can be obtained, in spite of capacity limitation in the function approximation architecture.

\noindent{\bf Problem II:}
Problem II has $21$ states, whose interconnections are depicted in Figure~\ref{fig-trgrph} (right). One state is located at the centre, and the rest of the states split evenly into four groups, indicated by the four loops in Figure~\ref{fig-trgrph}. The topology of the transition graph is the same for the target and behavior policies. We have drawn the transition graph only for the northeast group in Figure~\ref{fig-trgrph} (right); the states in each of the other three groups are arranged in the same manner and have the same transition structure. Given this symmetry, to specify the transition probability matrices $P_{\pi}$ and $P_{\pi^o}$, it suffices to specify the submatrices of $P_{\pi}$ and $P_{\pi^o}$ for the central state and one of the groups.
If we label the central state as state $1$ and the states in the northeast group counterclockwise as states $2$-$6$, the submatrices of $P_{\pi}$, $P_{\pi^o}$ for these states are given by
\begin{align*}
\text{target policy:} & \quad \left( 
\begin{array}{cccccc}
  0  &  0.25 & 0  &  0  & 0  &  0\\
  0 &  0 & 1 & 0  & 0  &  0\\
     0 & 0.2 & 0  & 0.8  & 0  & 0\\
     0  & 0  & 0.2  & 0  & 0.8 & 0\\
     0  & 0 &  0  & 0.2 & 0  & 0.8\\
     0.8 & 0 &  0 & 0  & 0.2  & 0 
     \end{array} \right),\\*[0.2cm]     
 \text{behavior policy:} & \quad    
     \left(
    \begin{array}{cccccc}
  0  &  0.25 & 0  &  0  & 0  &  0\\
  0 &  0 & 1 & 0  & 0  &  0\\
     0 & 0.5 & 0  & 0.5  & 0  & 0\\
     0  & 0  & 0.5  & 0  & 0.5 & 0\\
     0  & 0 &  0  & 0.5 & 0  & 0.5\\
     0.5 & 0 &  0 & 0  & 0.5  & 0 
     \end{array} \right).
\end{align*}         
Intuitively speaking, from the central state, the system enters one group of states by moving diagonally in one of the four directions with equal probability. After spending some time in that group, eventually returns to the central state and the process repeats. The behavior policy on average spends more time wandering inside each group than the target policy, while the target policy tends to traverse counterclockwise through the group more quickly. 

All the rewards are zero except for the middle state in each group -- for the northeast group, this is the shaded state in Figure~\ref{fig-trgrph} (right). For the two northern groups, their middle state has reward $1$, while for the two southern groups, that reward is $-1$.

The discount factor is $\gamma=0.9$ for all states. The interest weights and $\lambda$-parameters are set to be $\lambda(s)=0$, $\i(s) =1$ for all states.
As to features, we aggregate states into $5$ groups, the $4$ groups mentioned earlier and the central state forming its own group, and 
we let each state have $5$ binary features indicating its membership.
 
\begin{figure}[htb]
   \centering
      \includegraphics[width=0.27\linewidth]{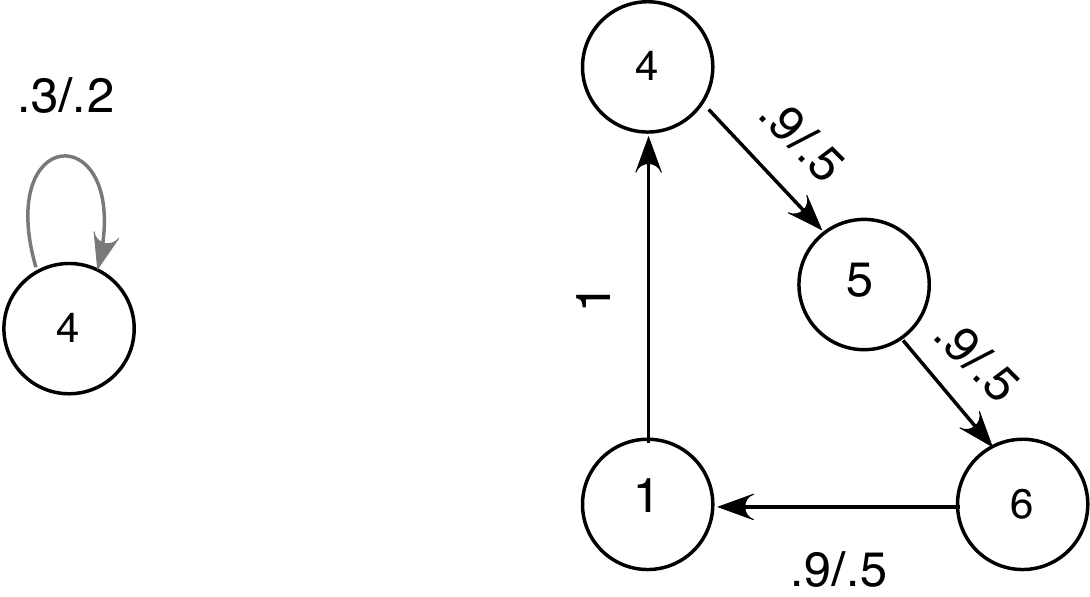} \qquad \qquad \qquad \quad
   \includegraphics[width=0.15\linewidth]{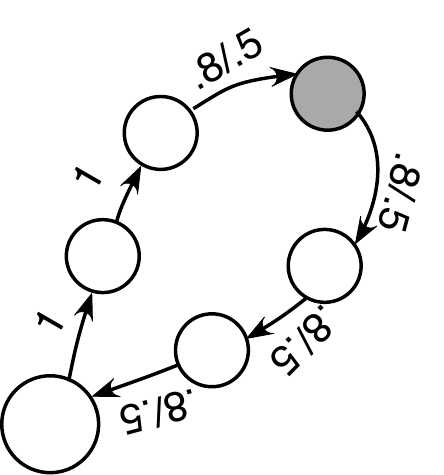}

 \caption{Some cycle patterns in the transition graphs of the test problems.}
   \label{fig-trgrph-cycles}
\end{figure}

\medskip
\noindent{\bf Behavior of traces:} We use the next three figures to illustrate the behavior of traces. (Readers who are interested only in the behavior of the $\theta$-iterates of the ETD algorithms may skip this part and go to the subsequent sections directly.) In general, by identifying certain cycle patterns in the transition graphs, one can infer whether the trace iterates $\{(\e_t, \F_t)\}$ will be unbounded over time almost surely \cite[Section 3.1]{Yu-siam-lstd}.  
Figure~\ref{fig-trgrph-cycles} shows a few examples of such cycles in the two test problems just described. 

The left graph in Figure~\ref{fig-trgrph-cycles} is a cycle of two states, $\{4, 4\}$, in the transition graph of Problem I. The edge of the graph is labeled with the importance sampling weight $0.3/0.2$ for the self-transition $4 \to 4$. (For the two test problems the importance sampling weights are simply given by the ratios between the entries of $P_\pi$ and $P_{\pi^o}$.) If we multiply together the importance sampling weight and the discount factor $\gamma(s)$ along this cycle, we get $ \tfrac{0.3}{0.2} \cdot \gamma(4)   = \tfrac{0.3}{0.2} > 1$,
while the interest weight $\i(4) > 0$ for the only state involved in this cycle. Then from \cite[Prop.\ 3.1]{Yu-siam-lstd} (cf.\ Footnote 3 therein) we can deduce that in Problem I, the follow-on traces $\{\F_t\}$ (which is updated according to $\F_t = \gamma_t \rho_{t-1} \F_{t-1} + \i(S_t)$ in this test problem) will be almost surely unbounded. 

Similarly, the right graph in Figure~\ref{fig-trgrph-cycles} is a cycle of states in the transition graph of Problem II. It consists of the central state and the northeast group of states. The importance sampling weights for each transition are labeled on the edges of the cycle. Traversing through the cycle once from any starting state, and multiplying together the importance sampling weights and the discount factors of each edge and its destination state, we get
$ \left( \tfrac{0.8}{0.5} \right)^4 \cdot \gamma^6 = \left( \tfrac{0.8}{0.5} \right)^4 \cdot 0.9^6 > 1,$
while at least one of the states in the cycle has a positive interest weight (since all the states are of interest in this problem). Then we can deduce as in the previous case that $\{\F_t\}$ will be unbounded almost surely. Hence, the eligibility traces $\e_t$ will also be unbounded in this case (because with $\lambda = 0$ in this problem, we have $\e_t = \F_t \phi(S_t)$).

As another example, suppose in Problem I we let $\lambda = 1$ for all states instead. 
The middle graph in Figure~\ref{fig-trgrph-cycles} exhibits a cycle of states in the transition graph in this case. If we multiply together the importance sampling weights and the $\gamma(s), \lambda(s)$ values on each edge and its destination state in this cycle, we get
$\left( \tfrac{0.9}{0.5} \right)^3 \cdot 0.7 > 1$,
while $\i(s) \phi(s)$ is nonnegative for all states in the cycle and nonzero for at least one (e.g., state $4$ or $6$). Then it can be deduced by using \cite[Prop.\ 3.1]{Yu-siam-lstd} as before that the eligibility traces $\{\e_t\}$ (generated in this case by $\e_t = \lambda_t \gamma_t \rho_{t-1} \e_{t-1} + \i(S_t) \phi(S_t)$) will be almost surely unbounded. 

\begin{figure}[!htb]
   \centering
   \includegraphics[width=0.49\linewidth]{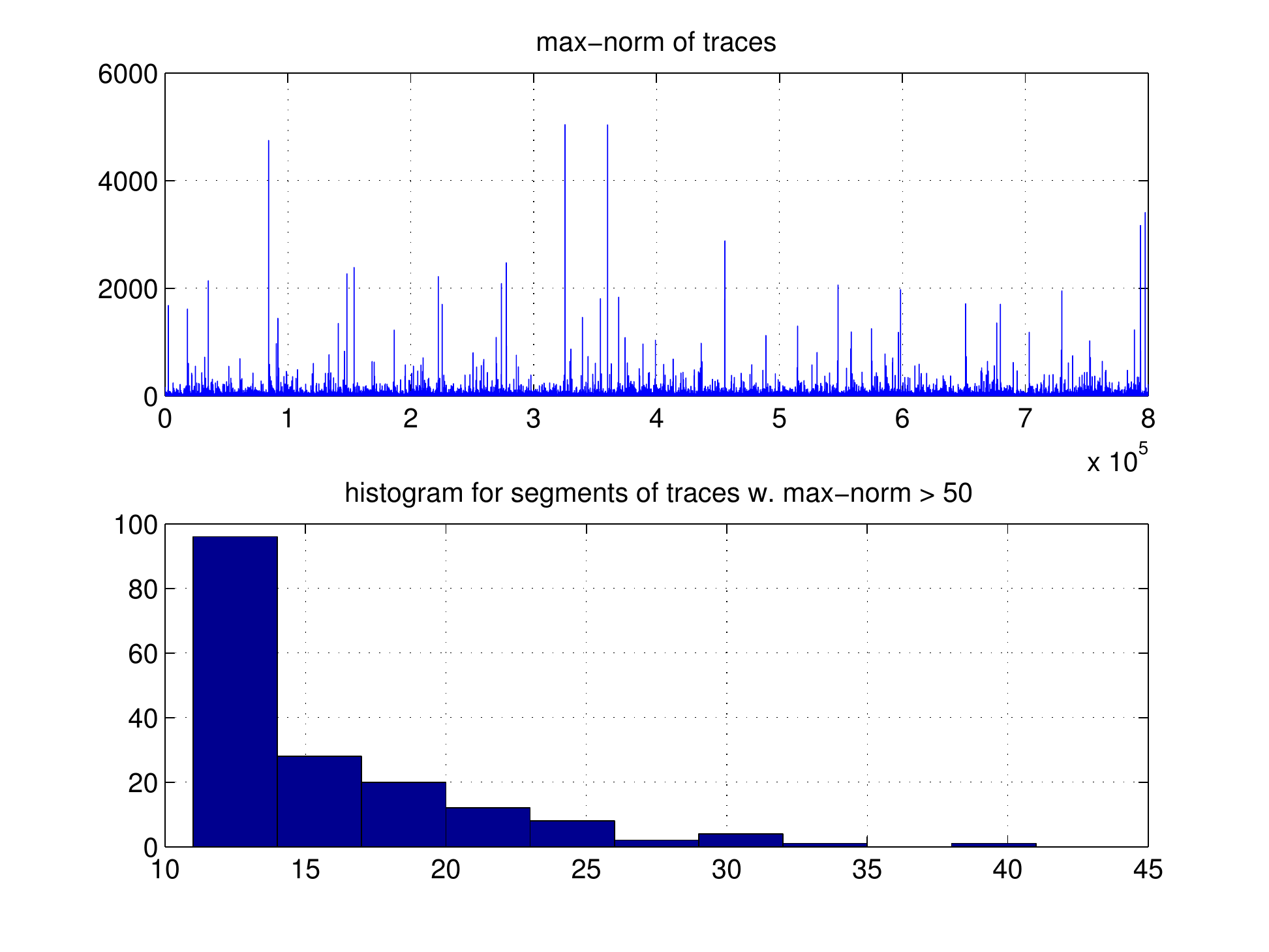} \hfill
   \raisebox{11pt}{\includegraphics[width=0.44\linewidth]{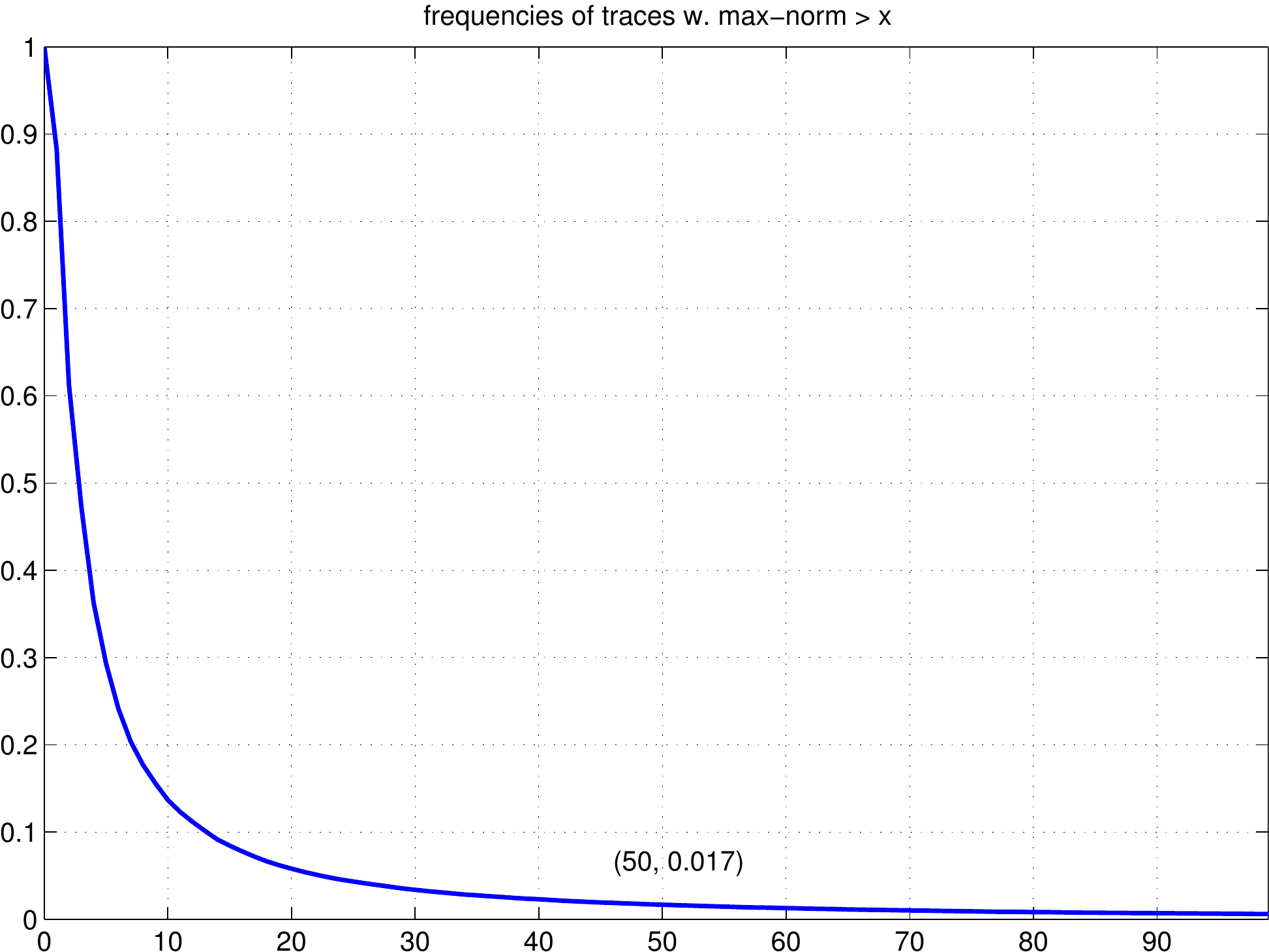}}
   \caption{Statistics of traces for Problem I. See the text for details.} \label{fig-trace1}
\end{figure}

\begin{figure}[!htb] 
   \centering
   \includegraphics[width=0.49\linewidth]{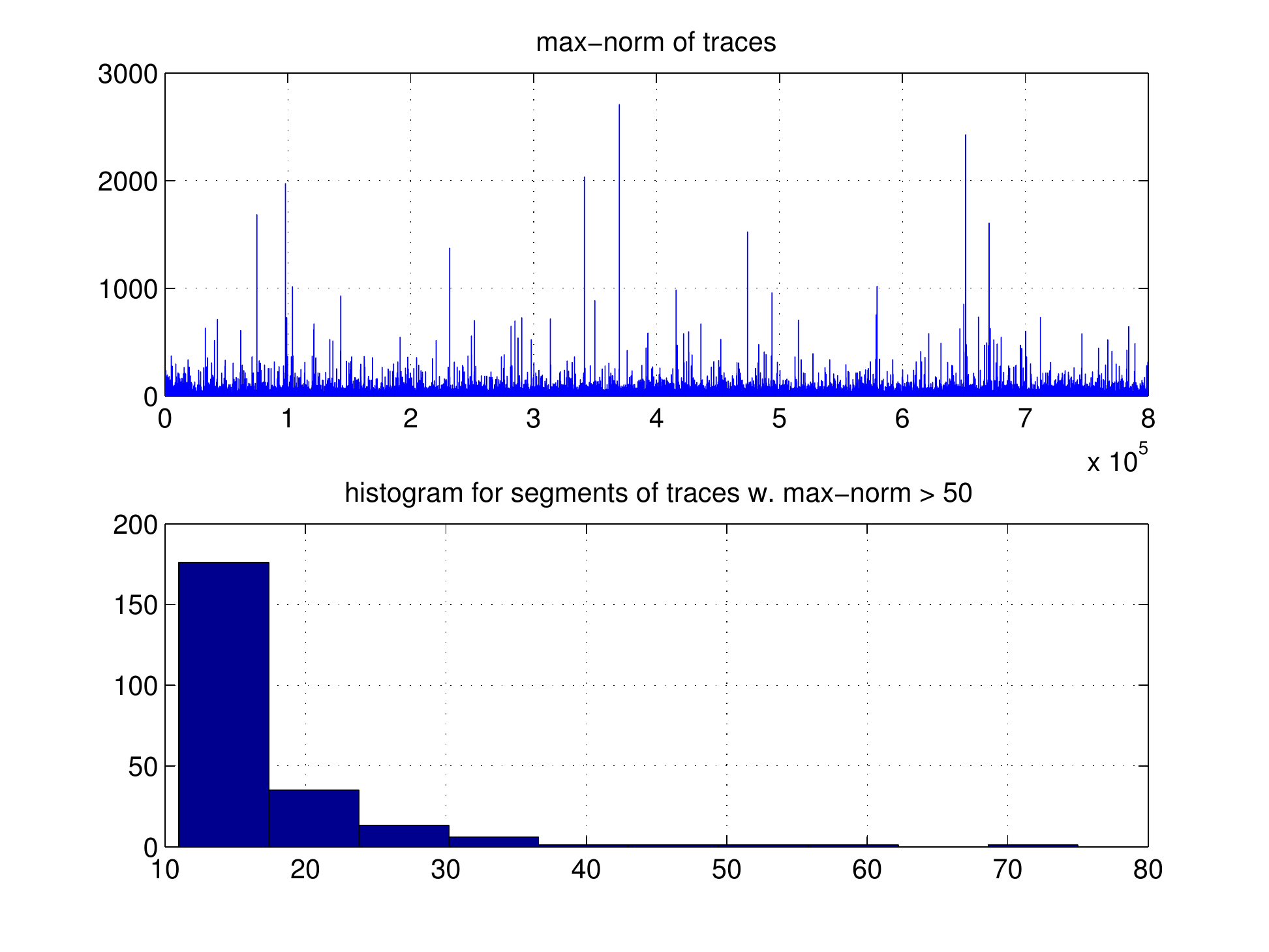} \hfill
   \raisebox{11pt}{\includegraphics[width=0.44\linewidth]{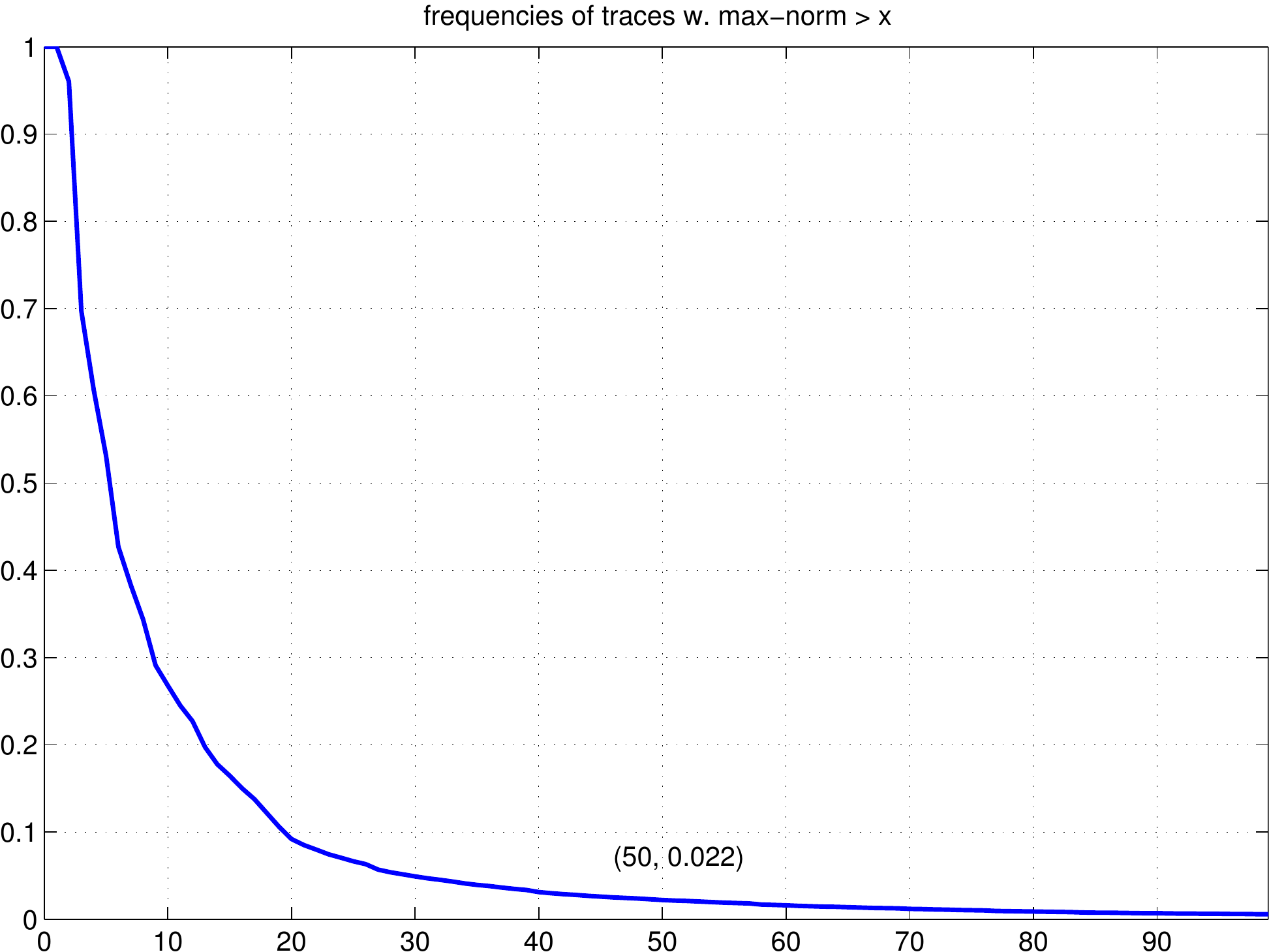}}
   \caption{Statistics of traces for Problem II. See the text for details.} \label{fig-trace2}
\end{figure}

We plotted in the upper left graphs of Figure \ref{fig-trace1} and Figure \ref{fig-trace2} the values of the max-norm $\|(\e_t, \F_t)\|$ over $8 \times 10^5$ iterations for the two test problems, respectively (the $x$-axis indicates the iteration $t$). One can see the recurring spikes in these graphs and the exceptionally large values of some of these spikes. This is consistent with the unboundedness of $\{(\e_t, \F_t)\}$ in the two test problems just discussed.

The unboundedness of $\{(\e_t, \F_t)\}$ tells us that the invariant probability measure $\zeta$ of the Markov chain $\{Z_t\} = \{(S_t, A_t, \e_t, \F_t)\}$ has an unbounded support. 
Despite this unboundedness, $\{(\e_t, \F_t)\}$ is bounded in probability (see the discussion in~\cite[Appendix A]{etd-wkconv}), and under the invariant distribution $\zeta$, $\E_\zeta \big[ \| (\e_0, \F_0)\| \big] < \infty$ (see \cite[Theorem 2.3]{etd-wkconv}). The latter relation implies that under the invariant distribution, the probability of $\| (\e_0, \F_0)\| > x$ decreases as $o(1/x)$ for large $x$.
Since the empirical distribution of $\{Z_t\}$ converges to $\zeta$ almost surely, during a run of many iterations, we expect to see the fraction of traces with $\| (\e_t, \F_t)\| > x$  drop in a similar way as $x$ increases.

The simulation results shown in the right part of Figures~\ref{fig-trace1}-\ref{fig-trace2} agree with the preceding discussion.
Plotted in those two graphs are fractions of traces with $\|(\e_t, \F_t)\| > x$ during $8\times 10^5$ iterations of the ETD algorithm (the vertical axis indicates the fraction, and the horizontal axis indicates $x$).
For instance, the fraction of traces with $\|(\e_t, \F_t)\| > 50$ is less than (abound) $0.02$ for Problem I (Problem II).
It can be seen that despite the recurring spikes in $\|(\e_t, \F_t)\|$ during the entire run, the fraction of traces with large magnitude $x$ drops sharply with the increase in $x$.

Finally, let us discuss yet another behavior of the traces, in connection with the biased constrained ETD algorithms that we will focus on in the rest of this note.  
Although only a small fraction of traces have exceptionally large magnitude, 
they can occur in consecutive iterations. 
We plotted two histograms in the lower left part of Figures~\ref{fig-trace1}-\ref{fig-trace2} to illustrate this type of behavior.
These histograms concern the excursions of the trajectory $(\e_t, \F_t)$, $t \geq 0$, outside of the box $\{  x \in \re^{n+1} \mid  \| x\| \leq 50\}$. 
The x-axis of the histograms indicates how long is such an excursion (i.e., the number of iterations it contains), 
and the y-axis indicates how many excursions of length $x$ occurred during the $8\times 10^5$ iterations of the experimental run. 
We plotted the histograms for length $x > 10$. 

This type of behavior suggests that it is better to apply the biased constrained ETD algorithms instead of the unbiased ones (constrained or unconstrained) in practice. 
This is because when traces with large magnitude occur in consecutive iterations, they can result in large changes in the $\theta$-iterates in a short period of time, if little constraint is put on the size of the change $\theta_{t+1} - \theta_t$ at each iteration. 
The unbiased ETD algorithms tend to be fragile in practice for this reason, despite their superior asymptotic convergence properties.
The biased algorithms take measures to prevent such abrupt changes in the $\theta$-iterates: for example, Variant I truncates the traces, and Variant II truncates the increments in the $\theta$-iterates. Because the fractions of traces with large magnitude are small, these truncations, with proper choices of threshold parameters, make only a small change in the mean update of ETD (cf.\ the discussion in \cite[Section 3.2]{etd-wkconv}). So the biased algorithms can gain much robustness in performance by paying only a small price of bias.

\section{Simulation Results for the Constant-stepsize Case} \label{sec-conststp}

In this section we show simulation results of the biased constrained ETD algorithms with a constant stepsize, for the two test problems described in the previous section. 
Besides the two biased algorithms, Variant I and Variant II, we will also show results for the perturbed versions of these two variants. Our focus will be on the behavior of multiple consecutive $\theta$-iterates and the behavior of a trajectory of $\theta$-iterates or their averaged iterates, under various stepsizes.

In the experiments reported below, the radius parameter $r_\H$ for constraining $\theta$ is set to be $r_\H=100$ (well above the threshold required by \cite[Lemma 2.1]{etd-wkconv}, which is calculated to be $ r_\H > 7.04$ for Problem I and  $r_\H > 5.20$ for Problem II).
The function $\psi_K$ in the variant algorithms (cf.\ Eq.~(3.2) in \cite{etd-wkconv}) is taken to be the componentwise truncation, 
$\psi_K(x) = \min\{K, \max \{ -K, x\}\}$, for $K=50$.
The perturbed versions of the two algorithms use the same $r_\H$ and $\psi_K$, and the perturbation variables $\Delta_{\theta,t}^\alpha$ (which are of the same size as $\theta$) are i.i.d.\ normal random variables with zero mean and covariance matrix $(\tfrac{\alpha}{2})^2 I$.

We will also show results of a modified version of ELSTD, which like Variant I also uses $\psi_K$ to truncate the traces $\e_t$ in its matrix/vector iterates. The limiting $\theta$ produced by this modified ELSTD is indeed the point that Variant I would converge to in the case of diminishing stepsize. Thus by running this version of ELSTD we can get an estimate of the bias in Variant I. To be concise, in what follows, we will often refer to this modified ELSTD algorithm simply as ELSTD.
 
To visualize the behavior of the algorithms, instead of plotting the iterates $\theta_t$ themselves, we will calculate and plot the normalized distances between $\theta_t$ and the desired ETD solution $\theta^*$. Here by the normalized distance we mean $|\theta_t - \theta^*|/|\theta^*|$, normalized by $|\theta^*|$, which is nonzero for both test problems. Correspondingly, we will refer frequently to $\delta$-neighborhoods of $\theta^*$ where $\delta$ are multiples of $|\theta^*|$, 
such as the $0.1 |\theta^*|$-neighborhood of $\theta^*$ or the $x |\theta^*|$-neighborhood of $\theta^*$ for some $x > 0$.

\subsection{Problem I}

The experiments below compare the behavior of the various algorithms in Problem I, for four different stepsizes: $\alpha = 0.01, 0.002, 0.001, 0.0005$.  
First, we did $4$ independent runs of both Variant I and Variant II. Each run lasted for $6\times 10^5$ iterations, during which the same state trajectory is used by both algorithms for all the four stepsizes. We did the same experiments for the perturbed versions of the two variants. To illustrate the steady state behavior, we used only the last $4 \times 10^5$ iterations of each run to obtain the statistics of multiple consecutive iterates shown in Figures~\ref{fig-cnst-ex1b}-\ref{fig-cnst-ex1e} below. 

\begin{figure}[h] 
\centering
\includegraphics[width=0.4\linewidth]{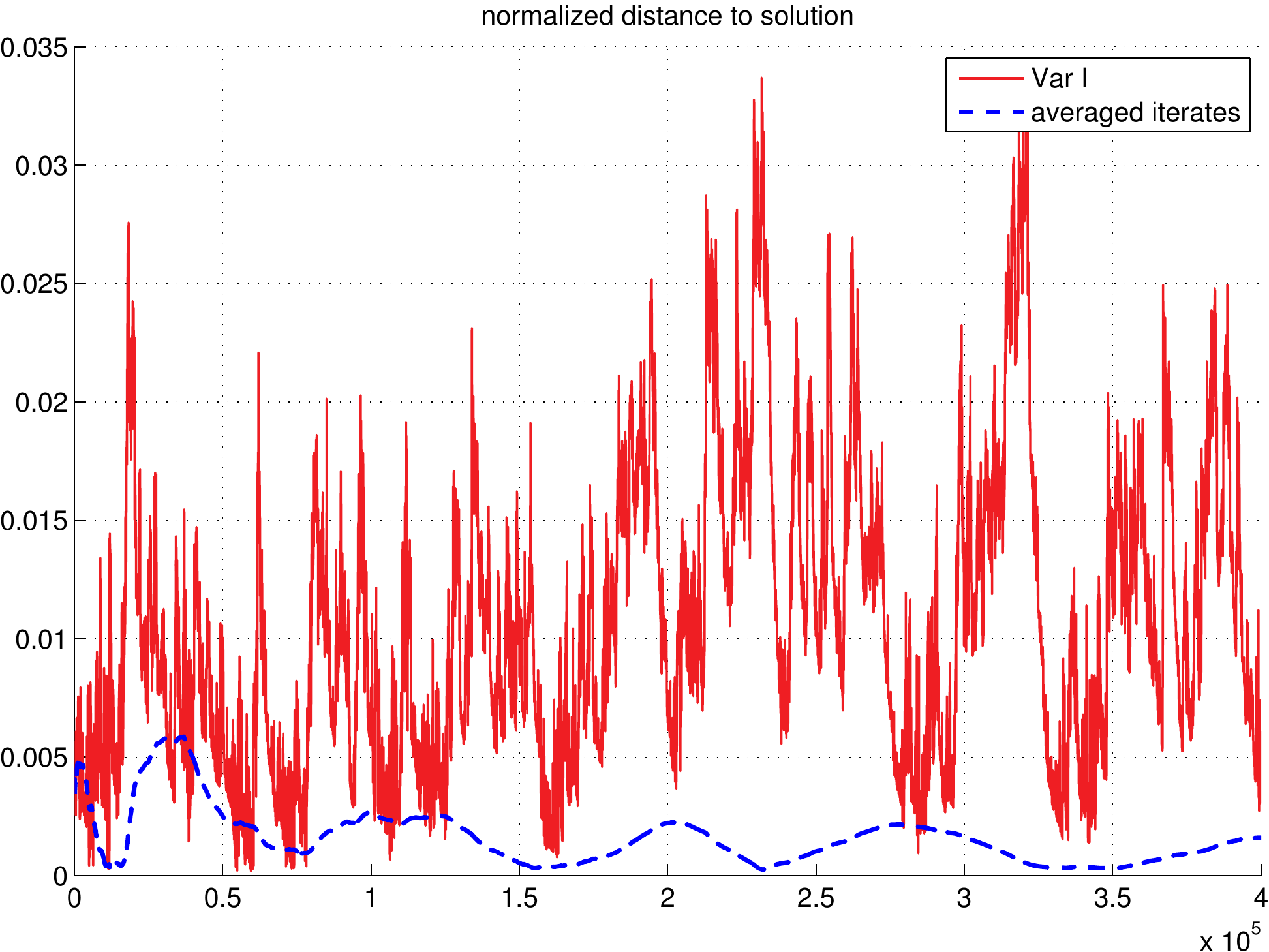} \qquad  
\includegraphics[width=0.4\linewidth]{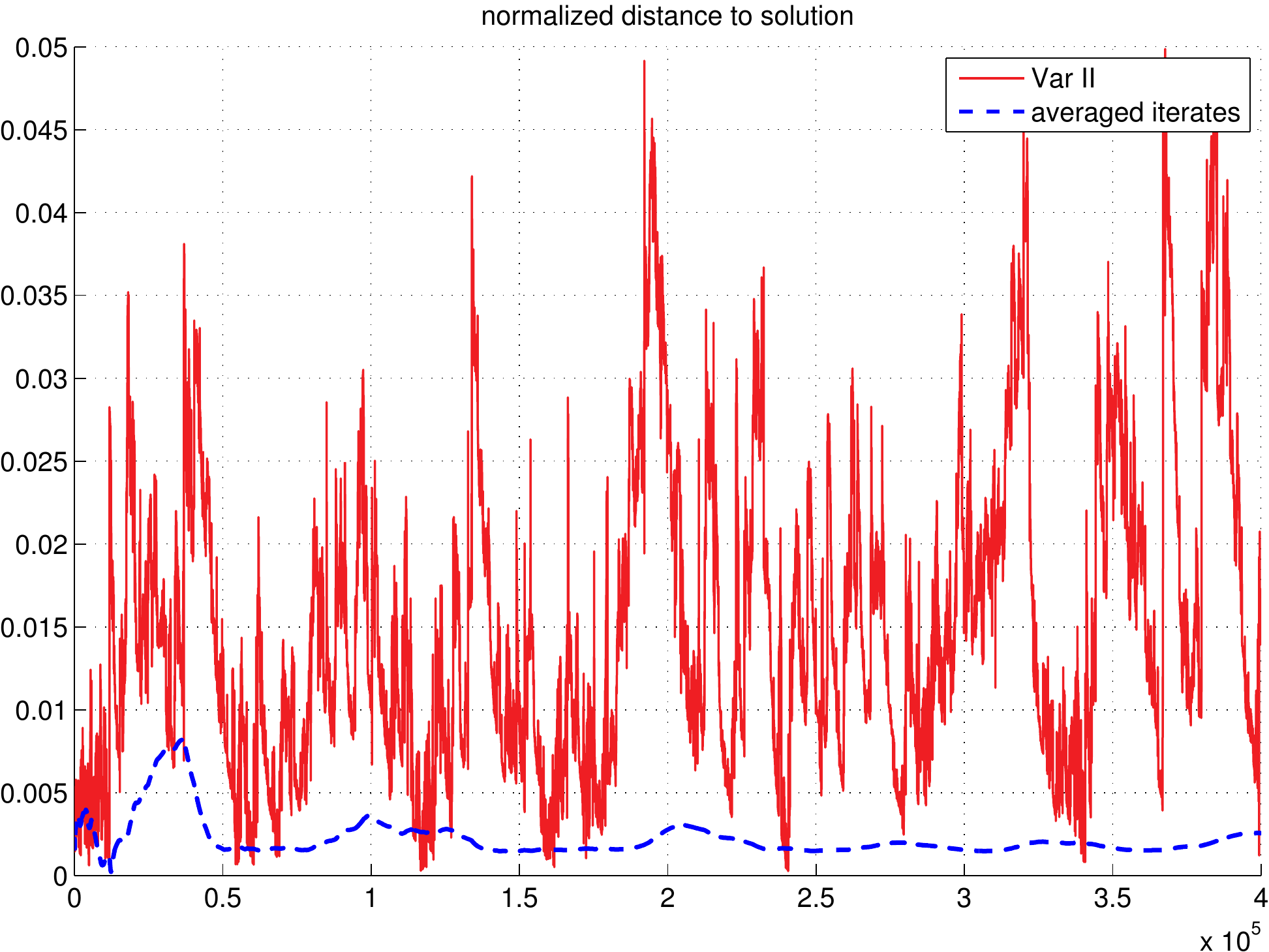}\\*[0.1cm]   
\includegraphics[width=0.4\linewidth]{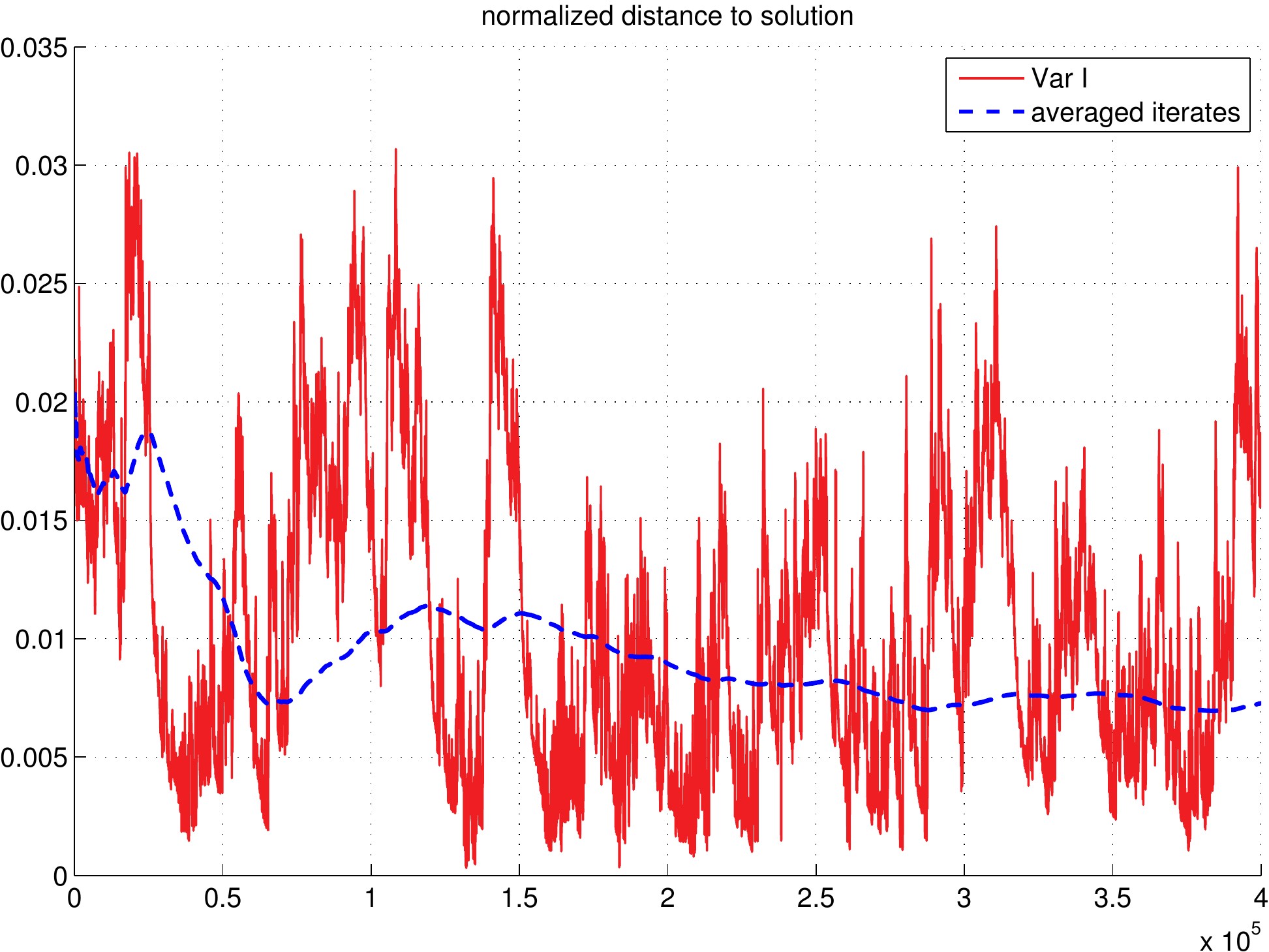} \qquad 
\includegraphics[width=0.4\linewidth]{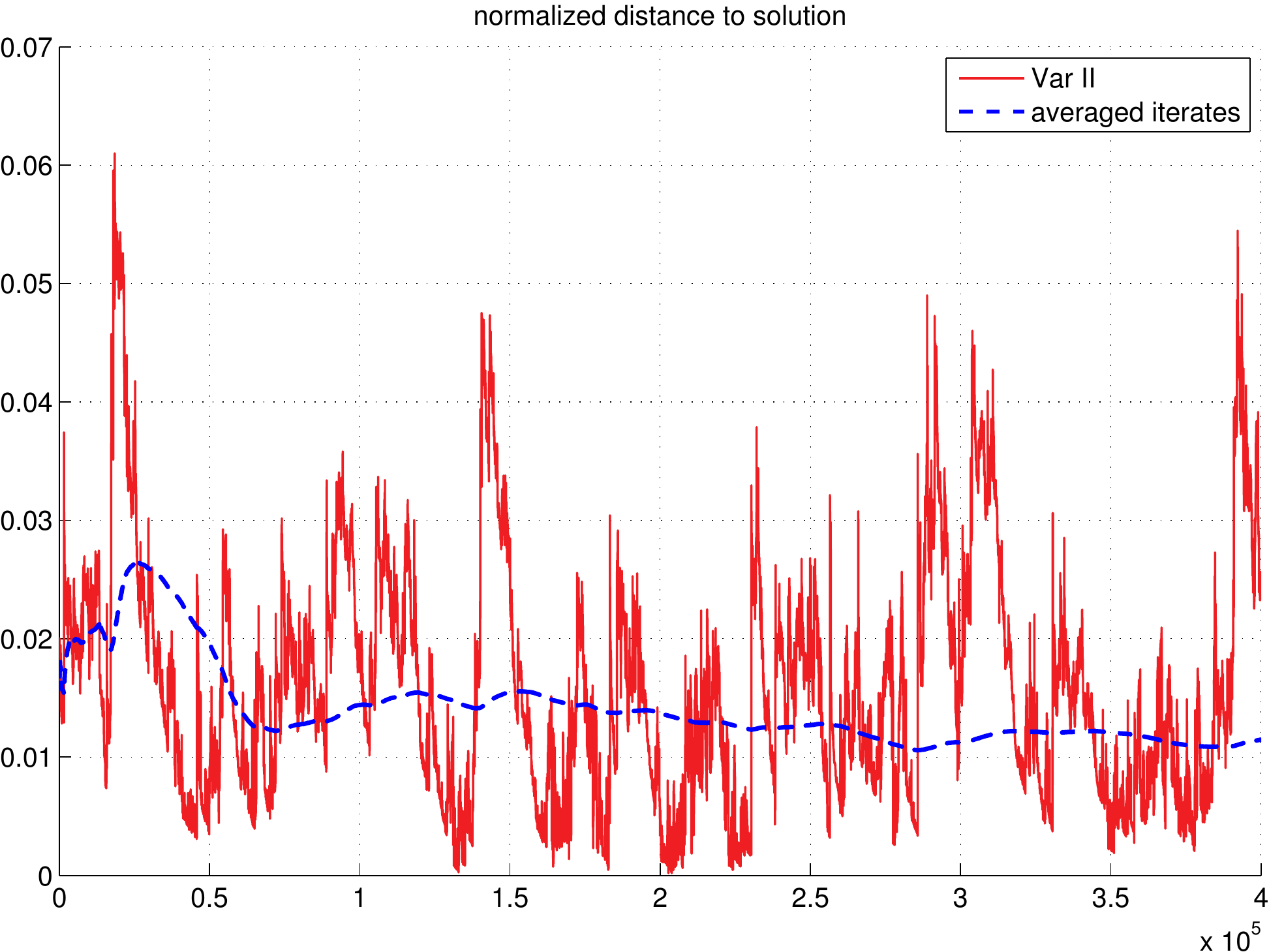}    
\caption{Variants I and II without (top) and with (bottom) perturbation ($\alpha = 0.0005$).} \label{fig-cnst-ex1a}
\end{figure}

Before proceeding to explain these figures, let us first show an example trajectory from a single run (more trajectories of iterates will be shown later). 
Plotted in Figure~\ref{fig-cnst-ex1a} are the normalized distances to $\theta^*$ of the iterates $\theta_t^\alpha$ produced  by Variants I and II (top row) and by their perturbed versions (bottom row) in the last $4 \times 10^5$ iterations of one run, for the smallest stepsize $\alpha = 0.0005$ in our experiments.
The dashed lines correspond to the averaged iterates $\bar \theta_t^\alpha$ (where the averaging also starts with the later portion of the run and neglects its initial portion). It can be seen that compared to the original iterates $\theta_t^\alpha$, the averaged iterates $\bar \theta_t^\alpha$ are much less volatile and, in the case of the unperturbed variant algorithms, approach a smaller neighborhood of $\theta^*$.  

Another note is that since the algorithms are biased, even if we had used smaller stepsizes, the iterates would not be able to approach an arbitrarily small neighborhood of $\theta^*$. To get an estimate of the degree of bias, we ran the modified ELSTD for $8$ independent runs of $8 \times 10^5$ iterations each. Averaged over the $8$ runs, the mean normalized distance (to $\theta^*$) of the ELSTD final solution was $0.0035$ with standard deviation $0.0017$. Consistently, we see from Figure~\ref{fig-cnst-ex1a} that most iterates of Variants I and II are still outside the $0.005 |\theta^*|$-neighborhood of $\theta^*$, although the averaged iterates $\bar \theta_t^\alpha$ seem to approach a smaller neighborhood. Recall also that Variant~II need not converge at all (cf.\ \cite[Section 3.2]{etd-wkconv}). In our experiments we observed it to behave similarly to Variant I and have a comparable bias (albeit slightly larger than that of Variant I for the two test problems).

We now proceed to explain the details of Figures~\ref{fig-cnst-ex1b}-\ref{fig-cnst-ex1e}, which demonstrate the behavior of multiple consecutive $\theta$-iterates for the four stepsizes:

\begin{figure}[!htb]    \centering
\includegraphics[width=0.48\linewidth]{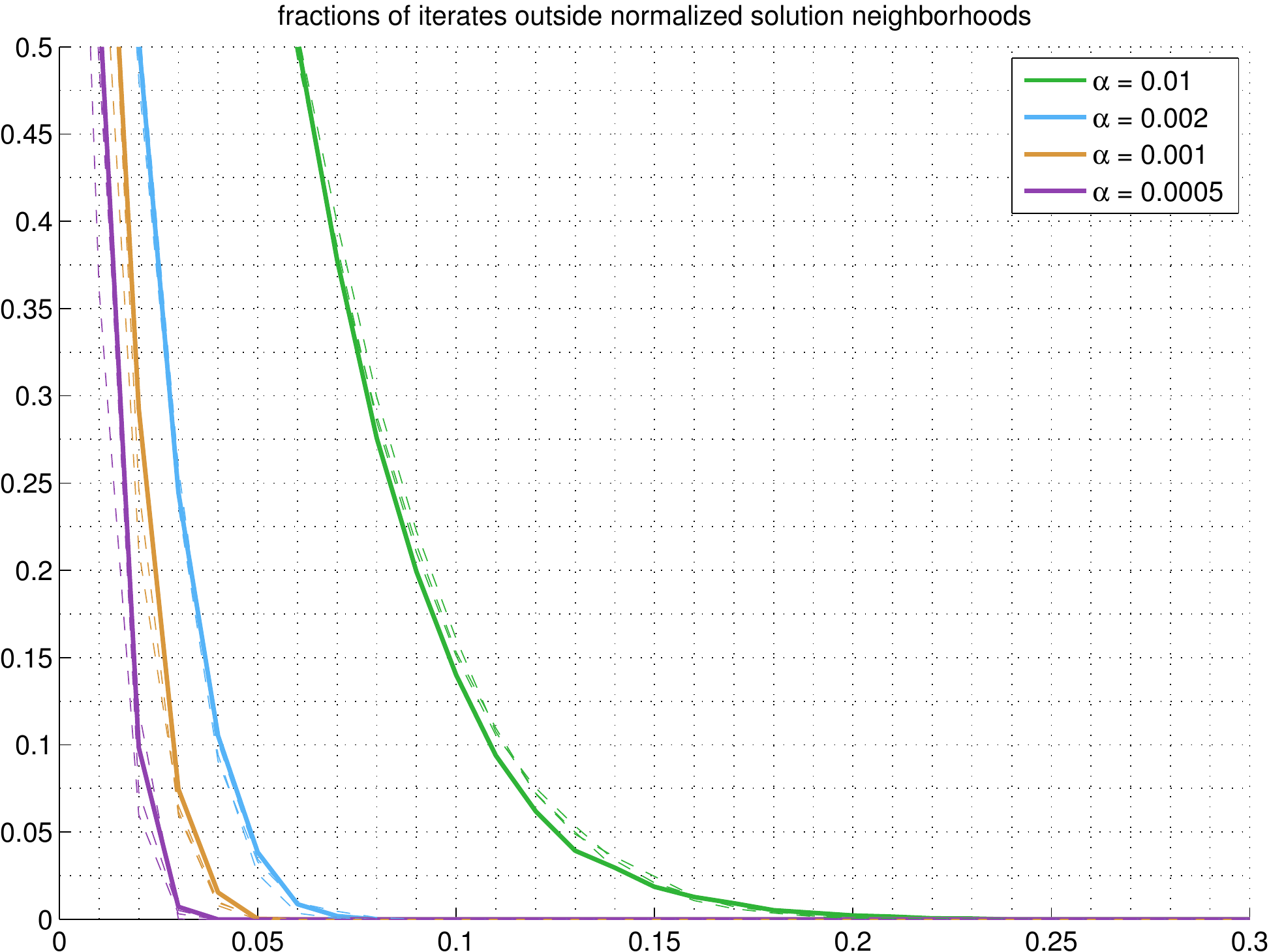} \hfill  
\includegraphics[width=0.48\linewidth]{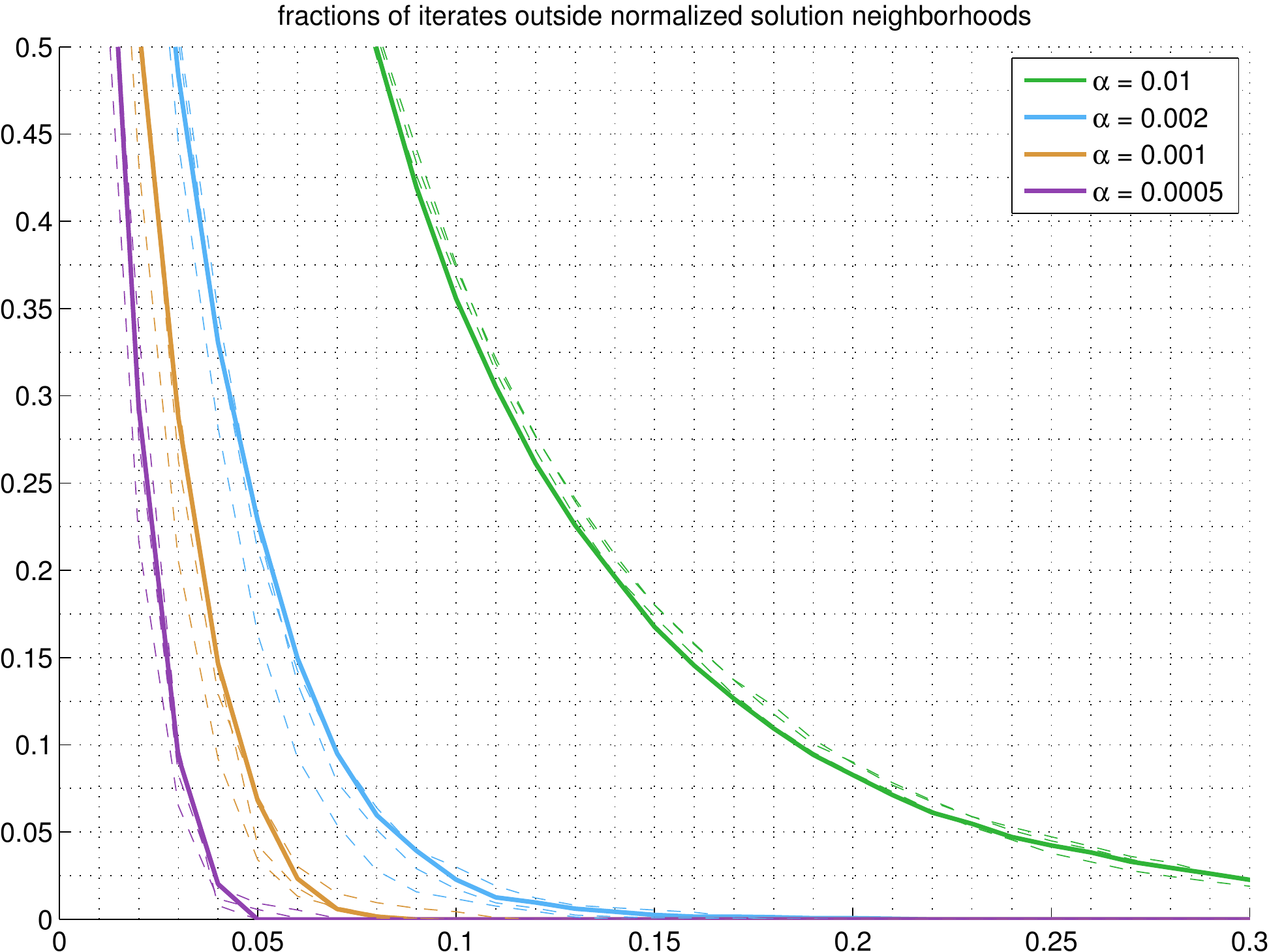}    
\caption{Variant I (left) and Variant II (right) without perturbation. The $x$-axis represents the $x |\theta^*|$-neighborhood of $\theta^*$. The $y$-component of a point $(x,y)$ represents the fraction of times (in a single run) that a segment of $100$ consecutive iterates fails to lie entirely inside the $x|\theta^*|$-neighborhood of $\theta^*$. (See the text for more details.)}\label{fig-cnst-ex1b}
\end{figure}

\begin{figure}[!htb] 
   \centering
\includegraphics[width=0.48\linewidth]{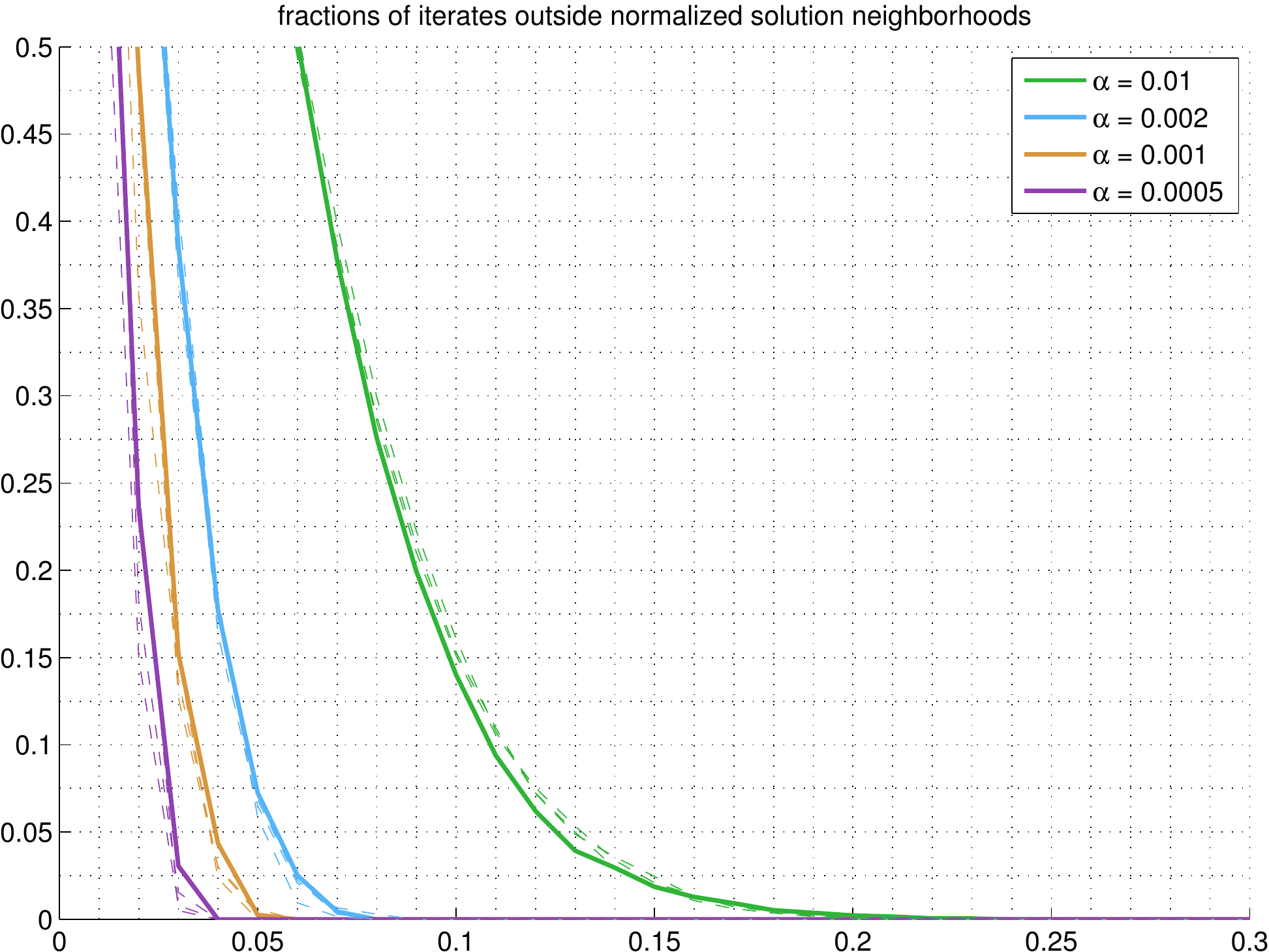} \hfill 
\includegraphics[width=0.48\linewidth]{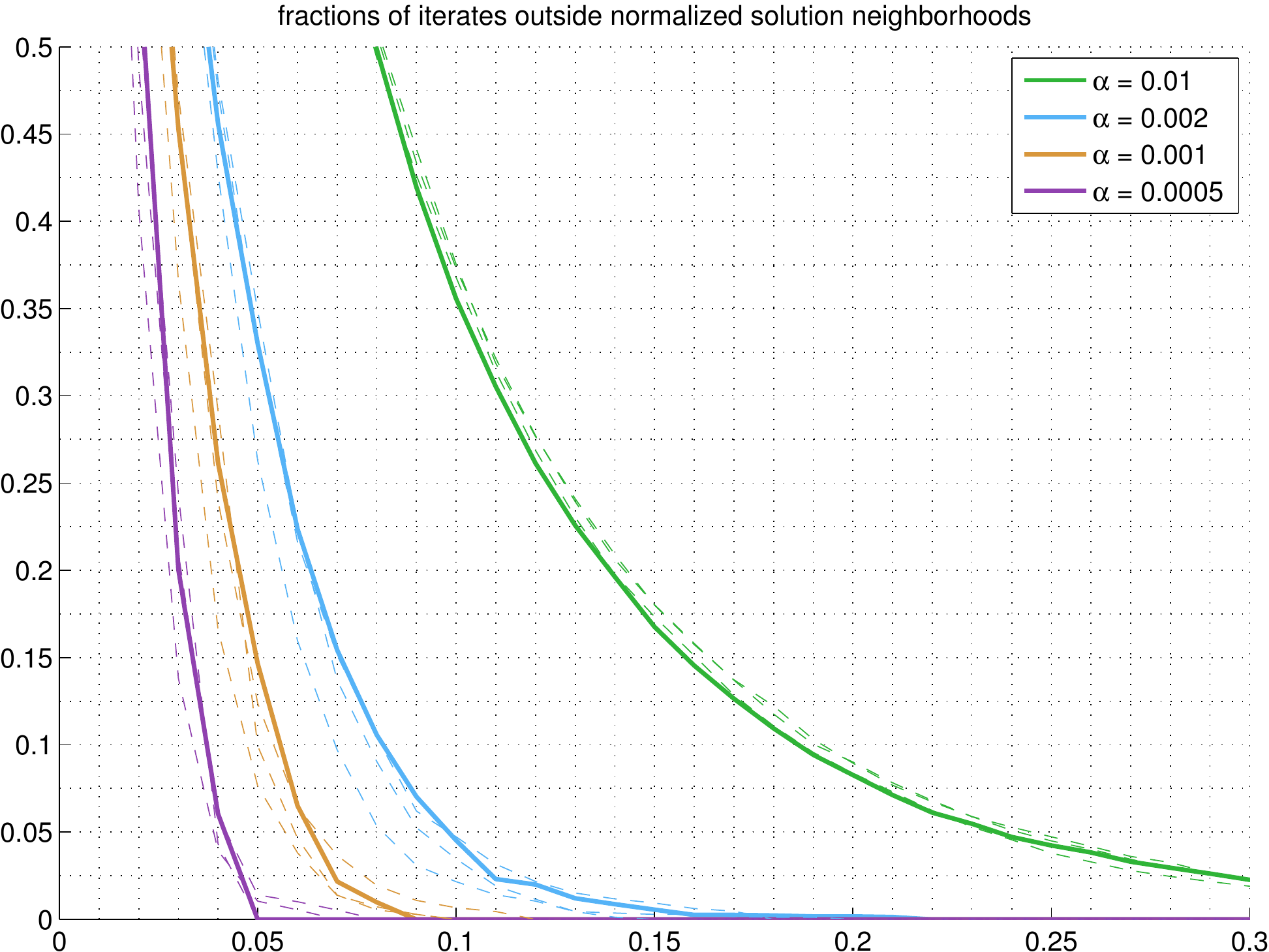}        
\caption{Variant I (left) and Variant II (right) without perturbation. The $x$-axis represents the $x |\theta^*|$-neighborhood of $\theta^*$. The $y$-component of a point $(x,y)$ represents the fraction of times (in a single run) that a segment of $\lfloor \tfrac{1}{\alpha} \rfloor$ consecutive iterates fails to lie entirely inside the $x|\theta^*|$-neighborhood of $\theta^*$. (See the text for more details.)}\label{fig-cnst-ex1c}
\end{figure}%

\medskip
\noindent Figures~\ref{fig-cnst-ex1b}-\ref{fig-cnst-ex1c}: In both figures, the $x$-axis represents the $x|\theta^*|$-neighborhood of $\theta^*$, and the $y$-component of a point $(x,y)$ represents the fraction of times that a certain number of consecutive iterates $\theta_t^\alpha$ fail to lie entirely inside the $x|\theta^*|$-neighborhood of $\theta^*$.
Specifically, for Figure~\ref{fig-cnst-ex1b} we consider every segment of $100$ consecutive iterates, $(\theta_{t}^\alpha, \ldots, \theta_{t+99}^\alpha), t= 0, 1, \ldots$, during each run. Plotted in Figure~\ref{fig-cnst-ex1b} are the fractions of times (during a single run) that such a segment fails to lie entirely inside the $x |\theta^*|$-neighborhood of $\theta^*$.
We then consider segments of $\lfloor \tfrac{1}{\alpha} \rfloor$ consecutive iterates,  $\big(\theta_{t}^\alpha, \ldots, \theta^\alpha_{t+\lfloor 1/\alpha \rfloor-1} \big), t = 0, 1, \ldots$. Plotted in Figure~\ref{fig-cnst-ex1c} are the fractions of times (during a single run) that a segment of length $\lfloor \tfrac{1}{\alpha} \rfloor$ fails to lie entirely inside the $x |\theta^*|$-neighborhood of $\theta^*$. (Note that the smaller the stepsize $\alpha$, the longer the segments used to calculate the fractions of times shown in the figure.)
In both figures, for each color and each algorithm, the solid line corresponds to the results from one of the four runs, while the three dashed lines correspond to the results from the other three runs.
It can be seen that the smaller the stepsize, the smaller the neighborhood of $\theta^*$ inside which a trajectory of iterates spends most of its time.  
The behavior of multiple consecutive iterates shown in these figures can be compared with the assertions in Theorem~3.4(ii) and Theorem 3.6(i) of \cite{etd-wkconv}.

\medskip
\noindent Figures~\ref{fig-cnst-ex1d}-\ref{fig-cnst-ex1e}: We repeated the same experiments for the perturbed versions of Variants I and II.  The results are shown in Figures~\ref{fig-cnst-ex1d}-\ref{fig-cnst-ex1e}, and they show similar behavior of the multiple consecutive iterates generated by these perturbed algorithms. (As in the previous case, in the figures, for each color, the solid lines correspond to the results from one of the four runs, and the dashed lines the other three runs.)
These simulation results can be compared with the assertions in Theorem~3.8 of \cite{etd-wkconv}.

\vfill
\pagebreak

\begin{figure}[!htb] 
   \centering
\includegraphics[width=0.46\linewidth]{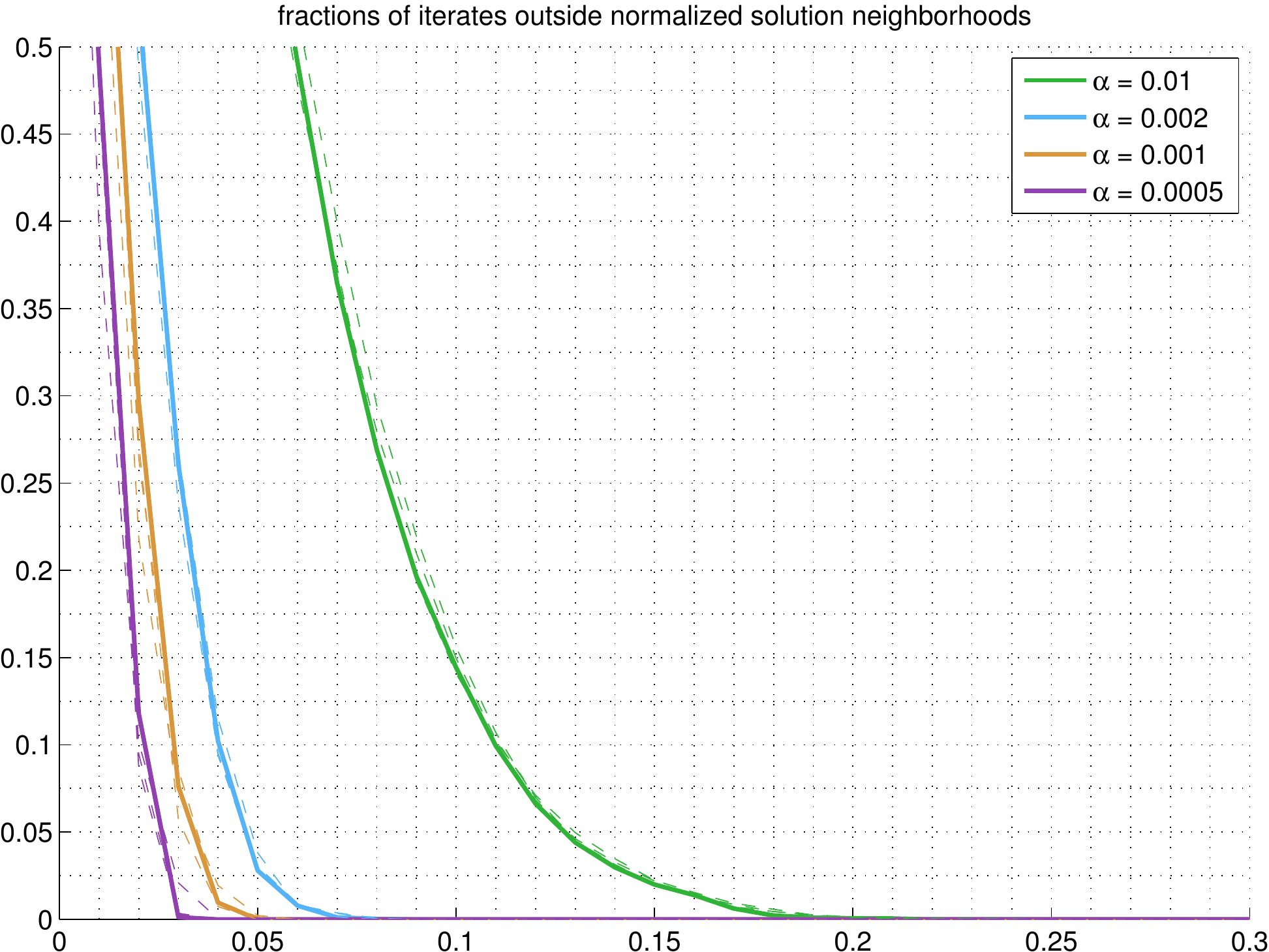} \hfill 
\includegraphics[width=0.46\linewidth]{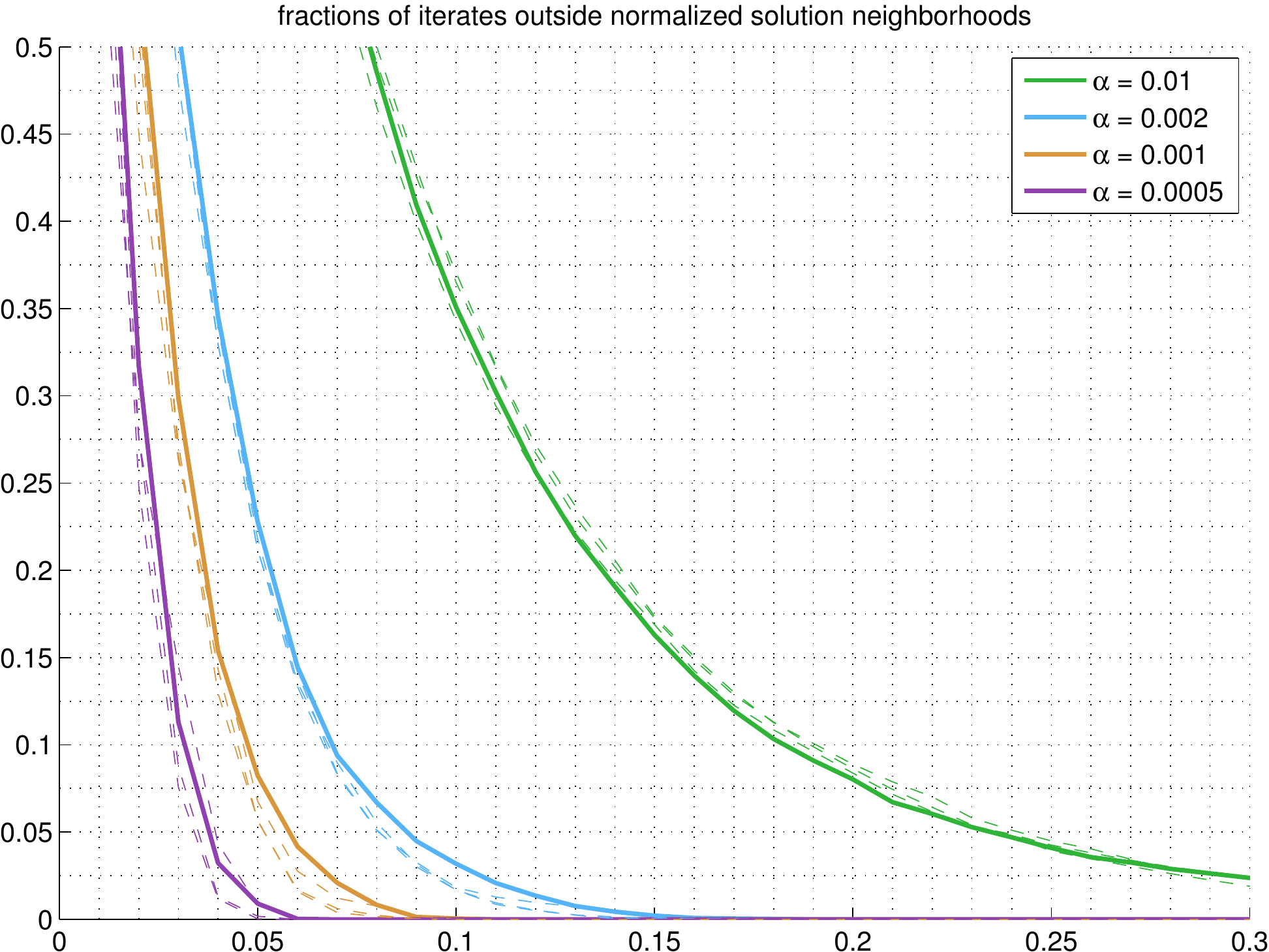}  
\caption{Variant I (left) and Variant II (right) with perturbation. The $x$-axis represents the $x |\theta^*|$-neighborhood of $\theta^*$. The $y$-component of a point $(x,y)$ represents the fraction of times (in a single run) that a segment of $100$ consecutive iterates fails to lie entirely inside the $x|\theta^*|$-neighborhood of $\theta^*$. See the text for more details.}\label{fig-cnst-ex1d}
\end{figure}
\begin{figure}[!htb] 
   \centering
\includegraphics[width=0.46\linewidth]{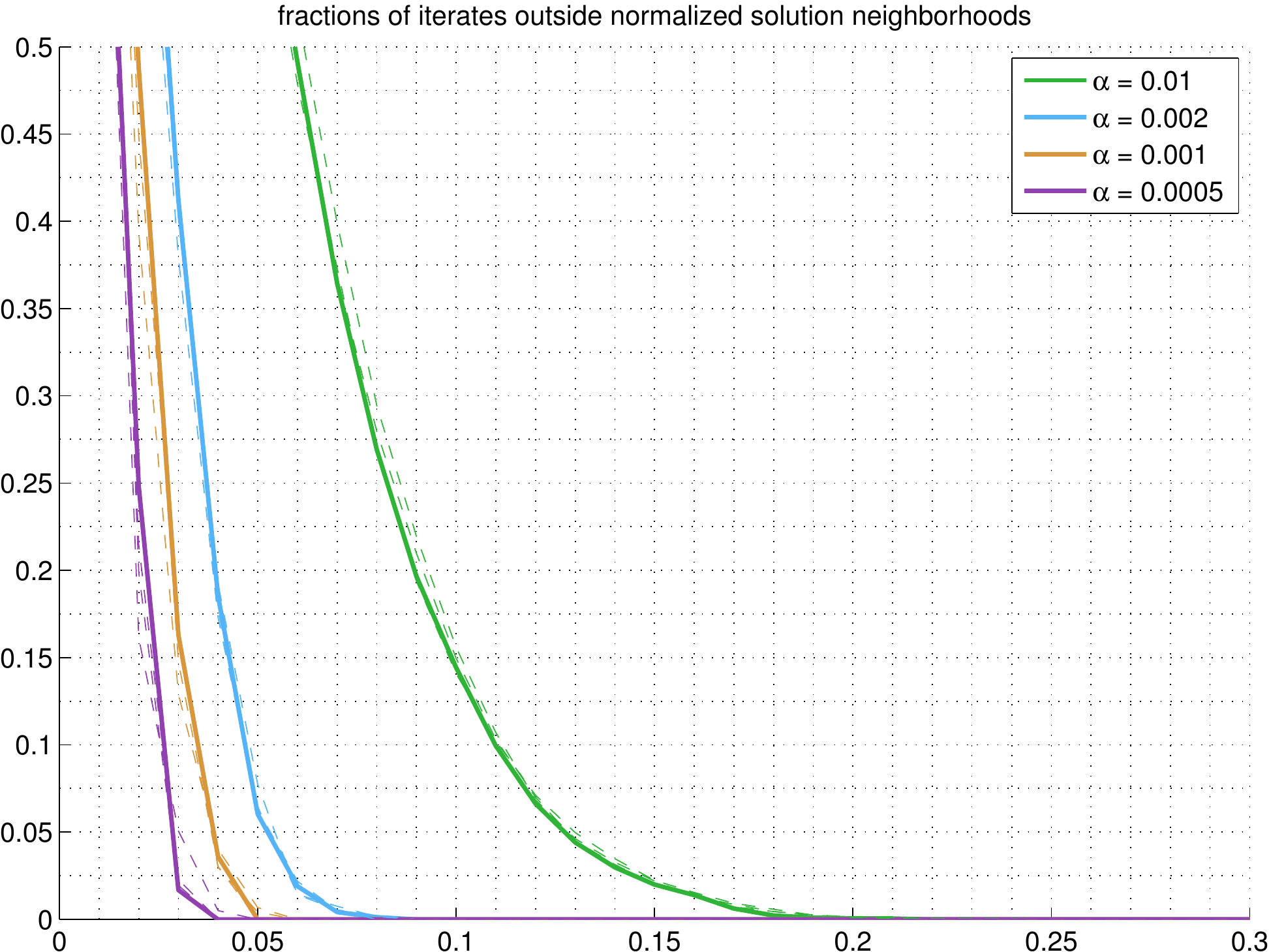} \hfill    
\includegraphics[width=0.46\linewidth]{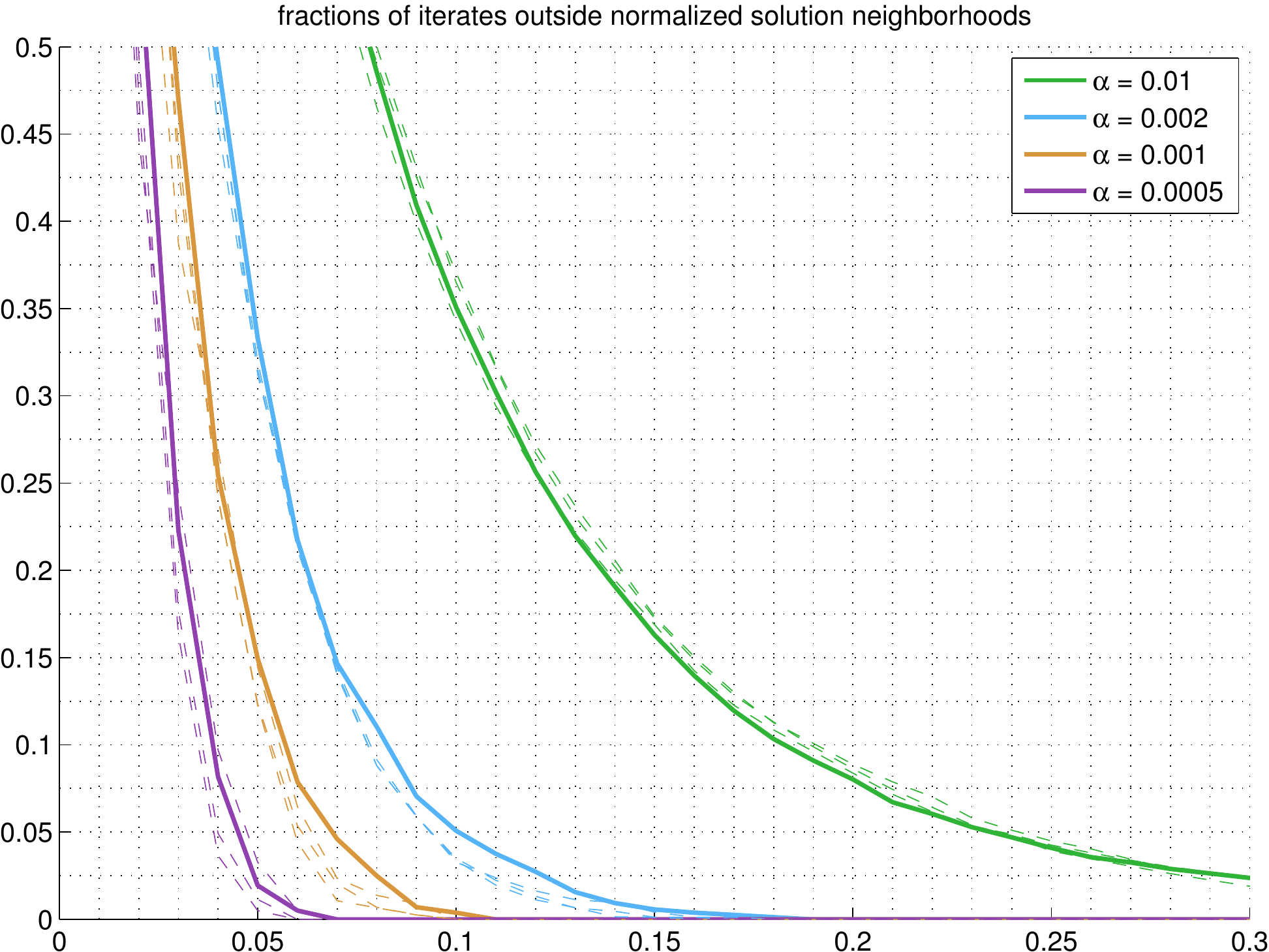}      
\caption{Variant I (left) and Variant II (right) with perturbation. The $x$-axis represents the $x |\theta^*|$-neighborhood of $\theta^*$. The $y$-component of a point $(x,y)$ represents the fraction of times (in a single run) that a segment of $\lfloor \tfrac{1}{\alpha} \rfloor$ consecutive iterates fails to lie entirely inside the $x|\theta^*|$-neighborhood of $\theta^*$. (See the text for more details.)}\label{fig-cnst-ex1e}
\end{figure}

\begin{figure}[!t] 
   \centering
\includegraphics[width=0.48\linewidth]{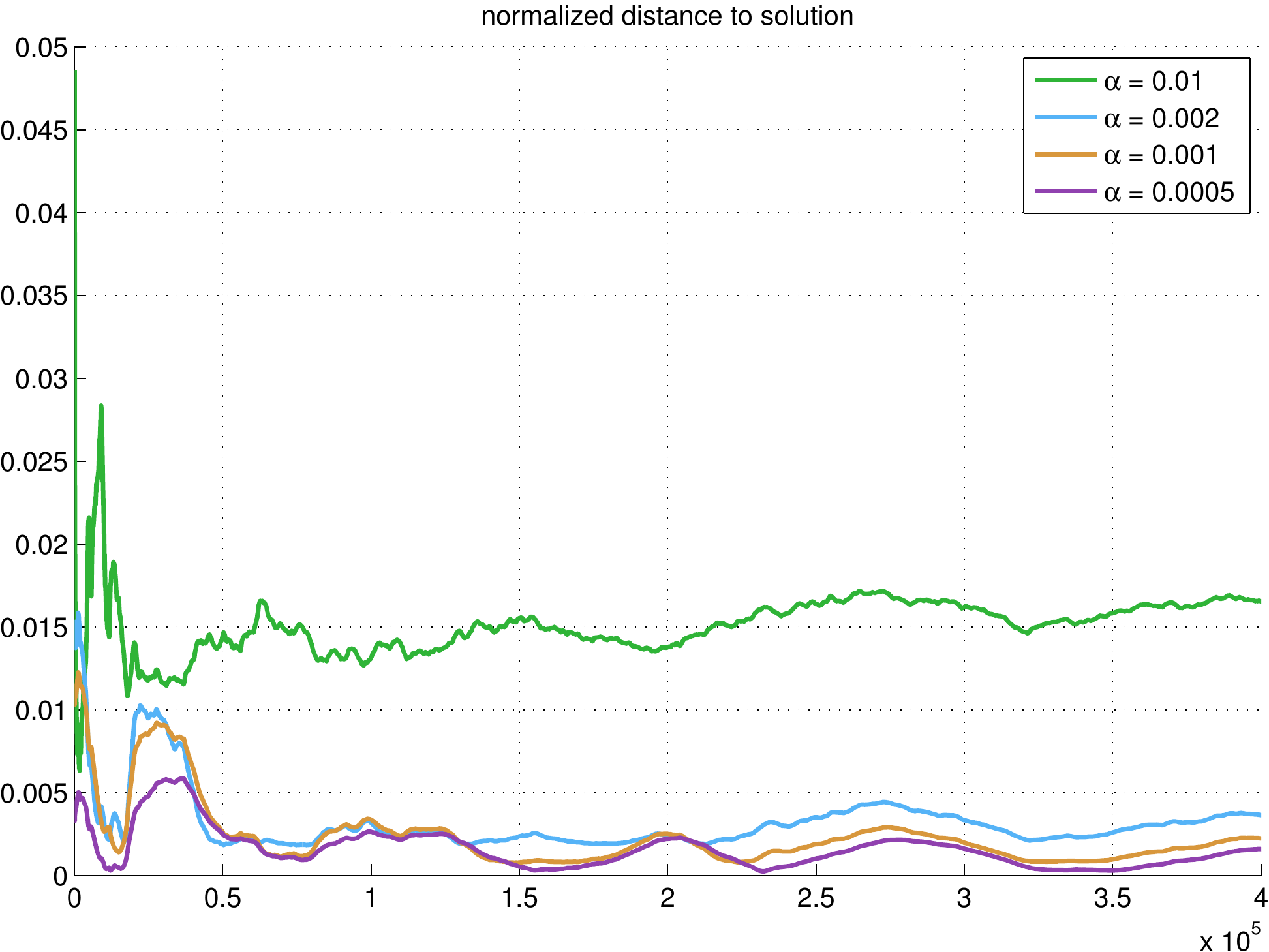} \hfill 
\includegraphics[width=0.48\linewidth]{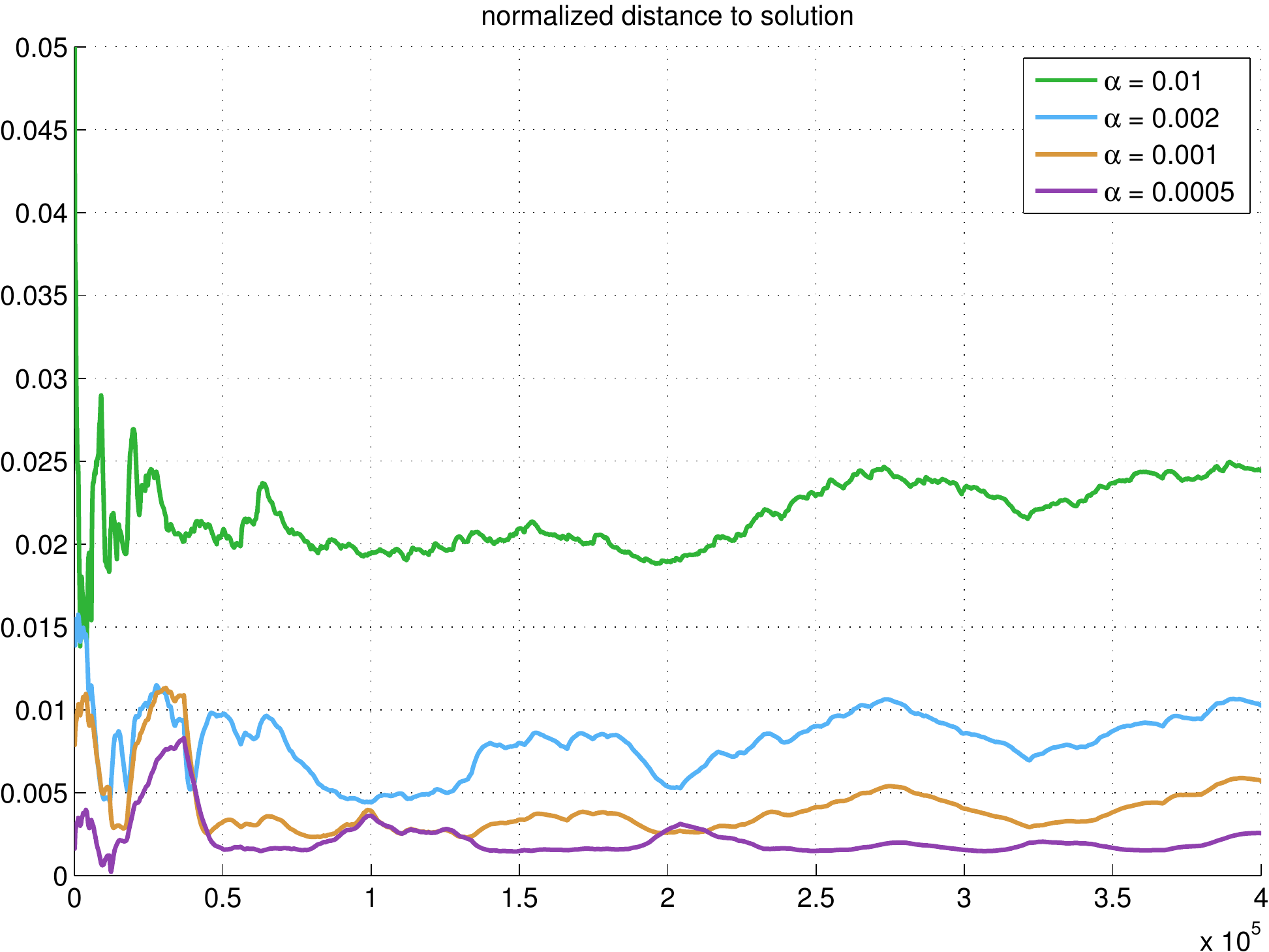}\\ 
\includegraphics[width=0.48\linewidth]{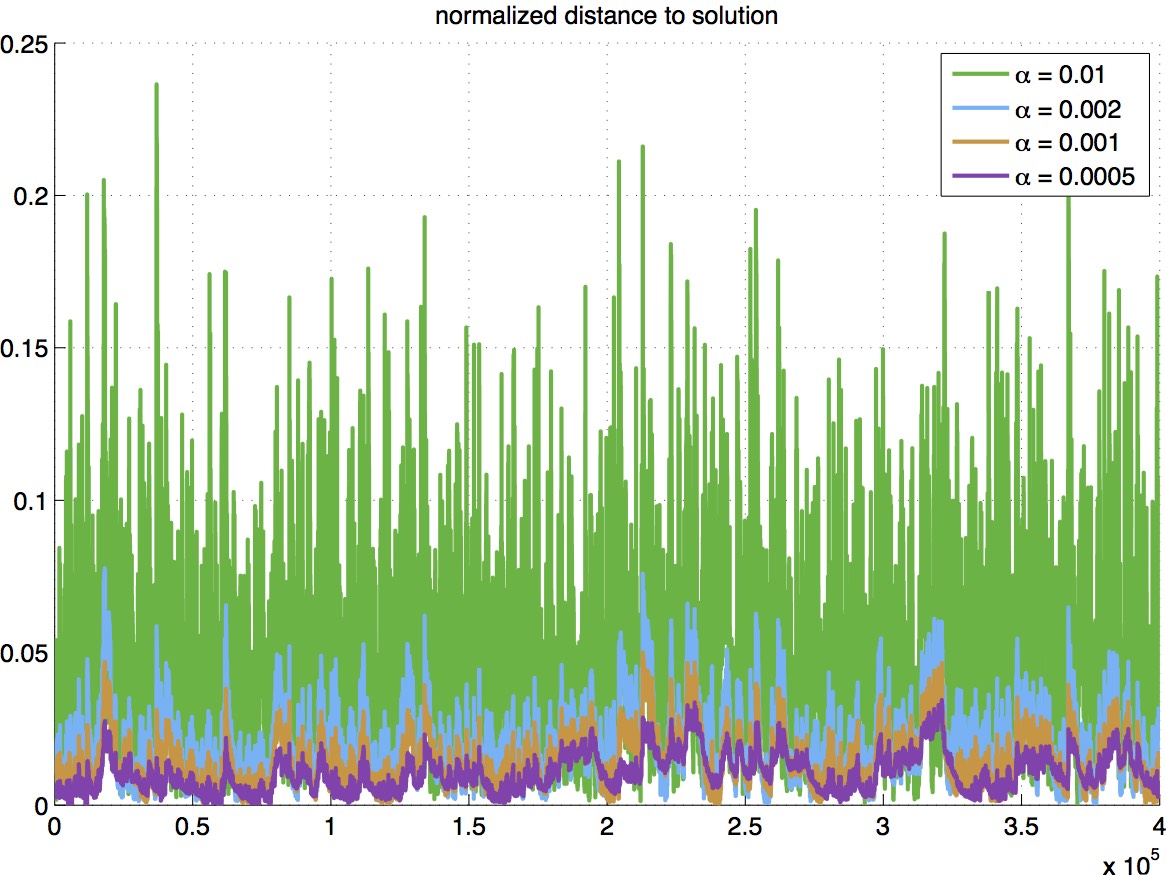} \hfill 
\includegraphics[width=0.48\linewidth]{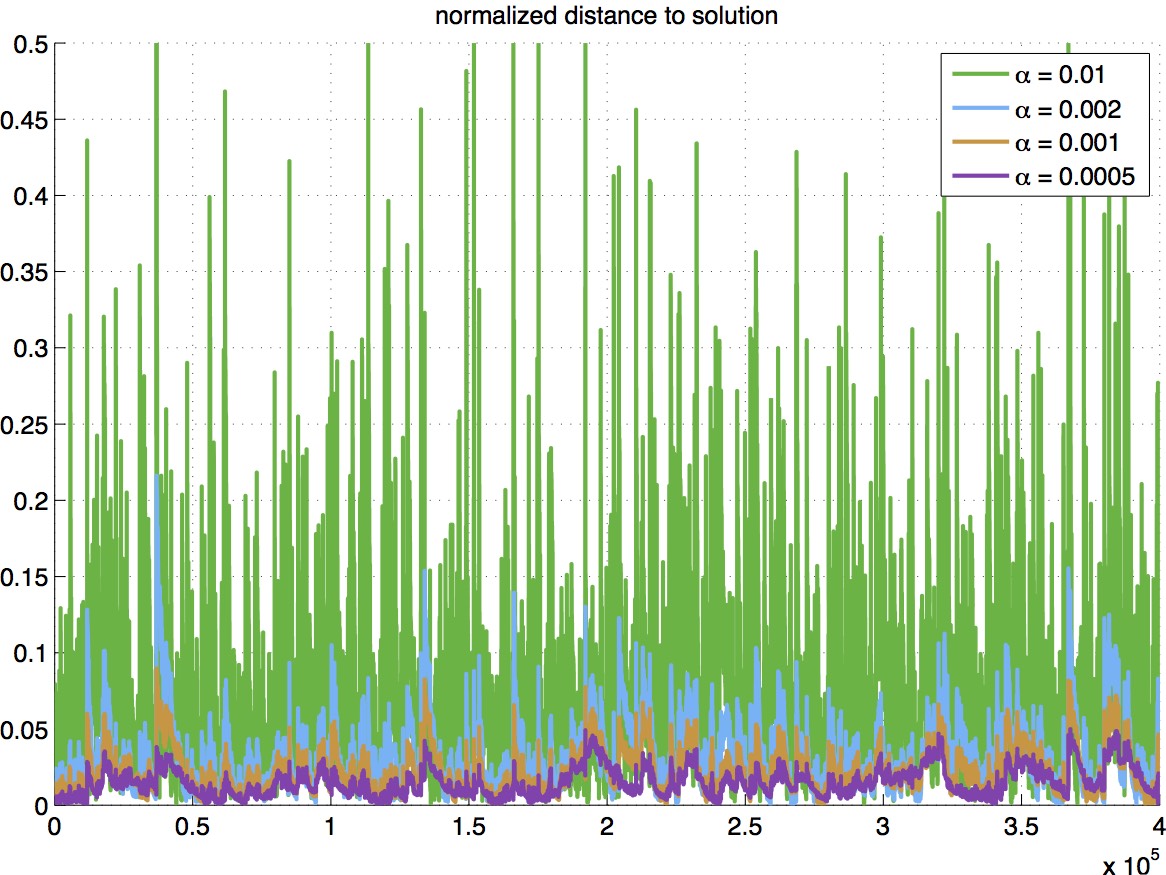} 
\caption{Variant I (left) and Variant II (right) without perturbation. Top: averaged iterates $\bar\theta_t^\alpha$; bottom: iterates $\theta_t^\alpha$. Data are from a single run.}\label{fig-cnst-ex1f}
\end{figure}

\begin{figure}[!thb] 
  \centering
\includegraphics[width=0.48\linewidth]{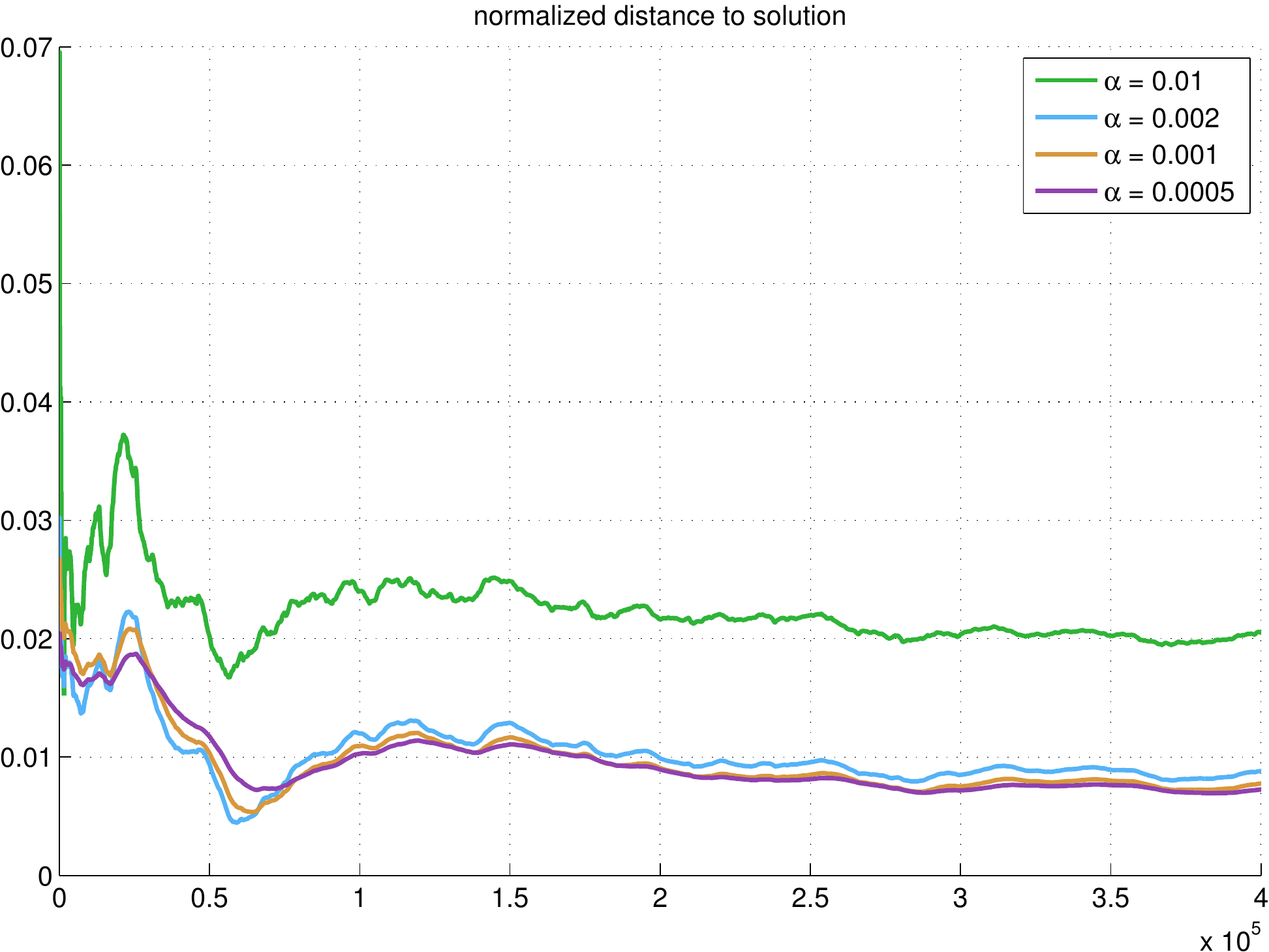} \hfill 
\includegraphics[width=0.48\linewidth]{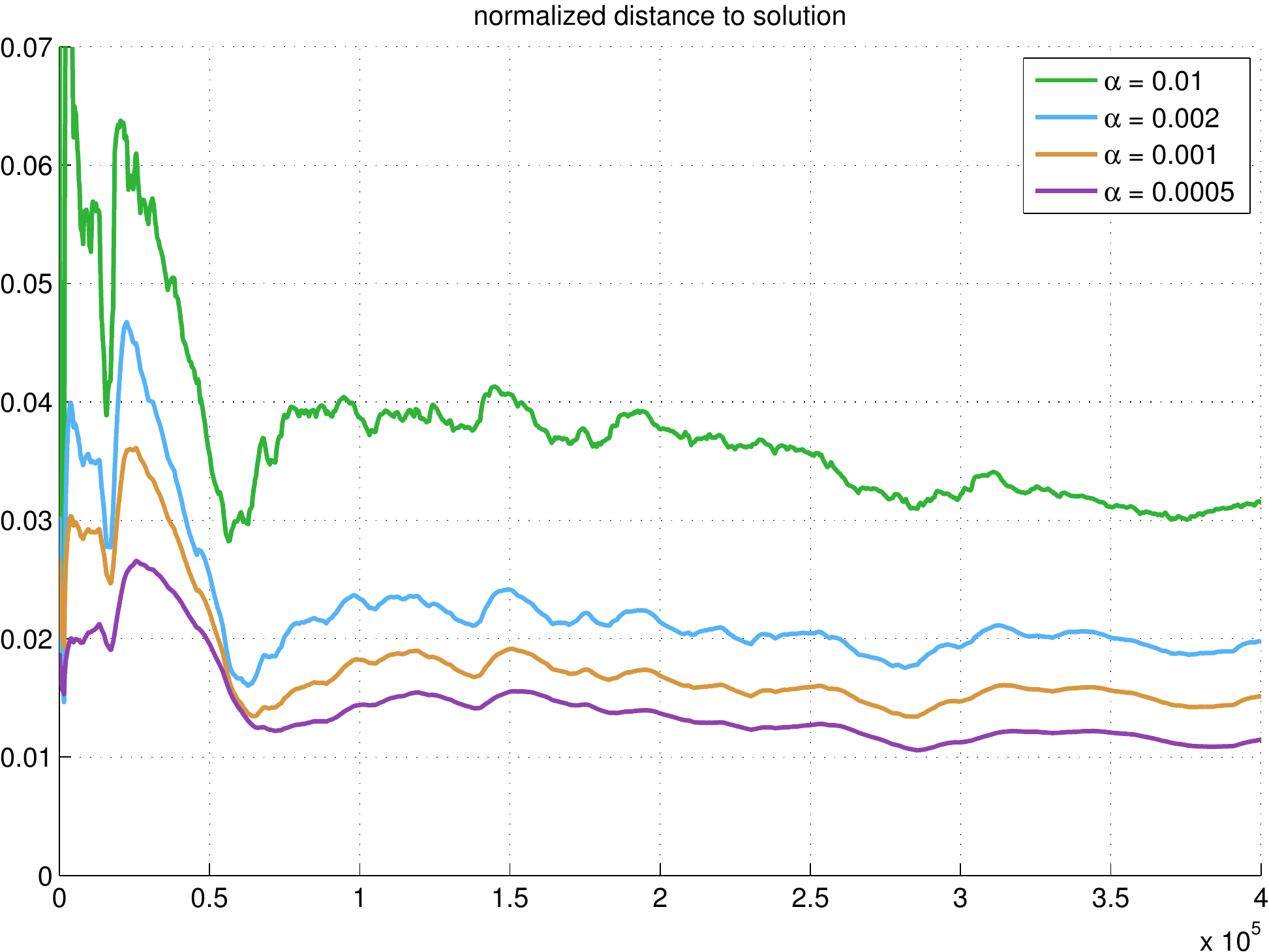}\\ 
\includegraphics[width=0.48\linewidth]{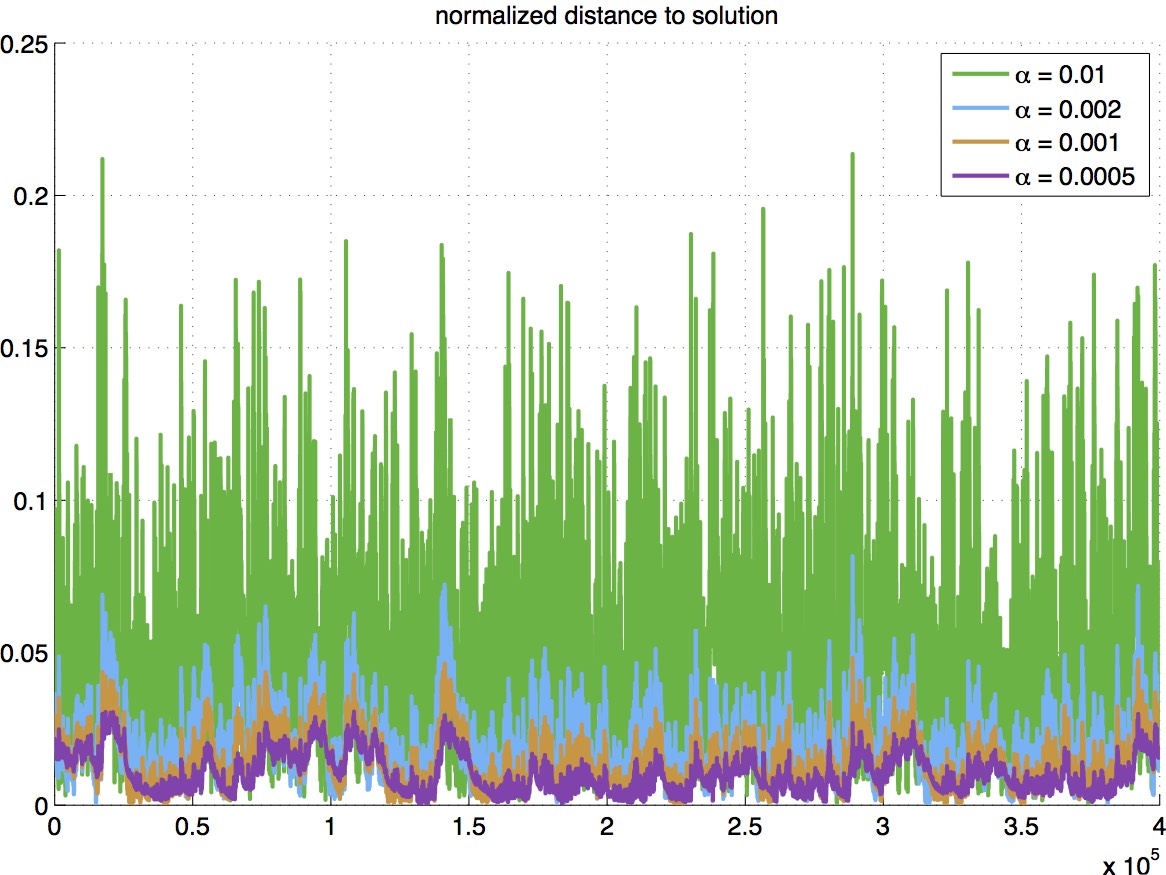} \hfill 
\includegraphics[width=0.48\linewidth]{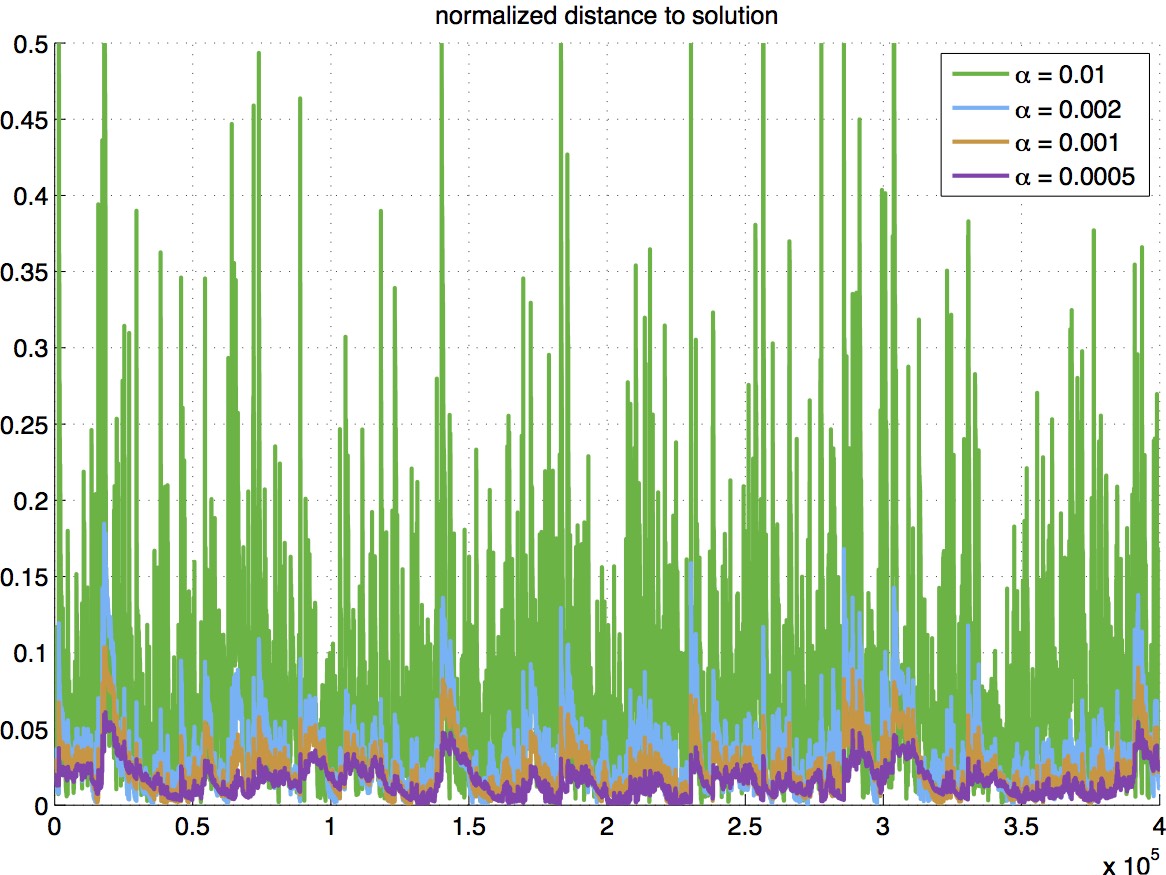} 
\caption{Variant I (left) and Variant II (right) with perturbation. Top: averaged iterates $\bar\theta_t^\alpha$; bottom: iterates $\theta_t^\alpha$. Data are from a single run.}\label{fig-cnst-ex1g}
\end{figure}

\begin{figure}[!htb] 
   \centering
\includegraphics[width=0.45\linewidth]{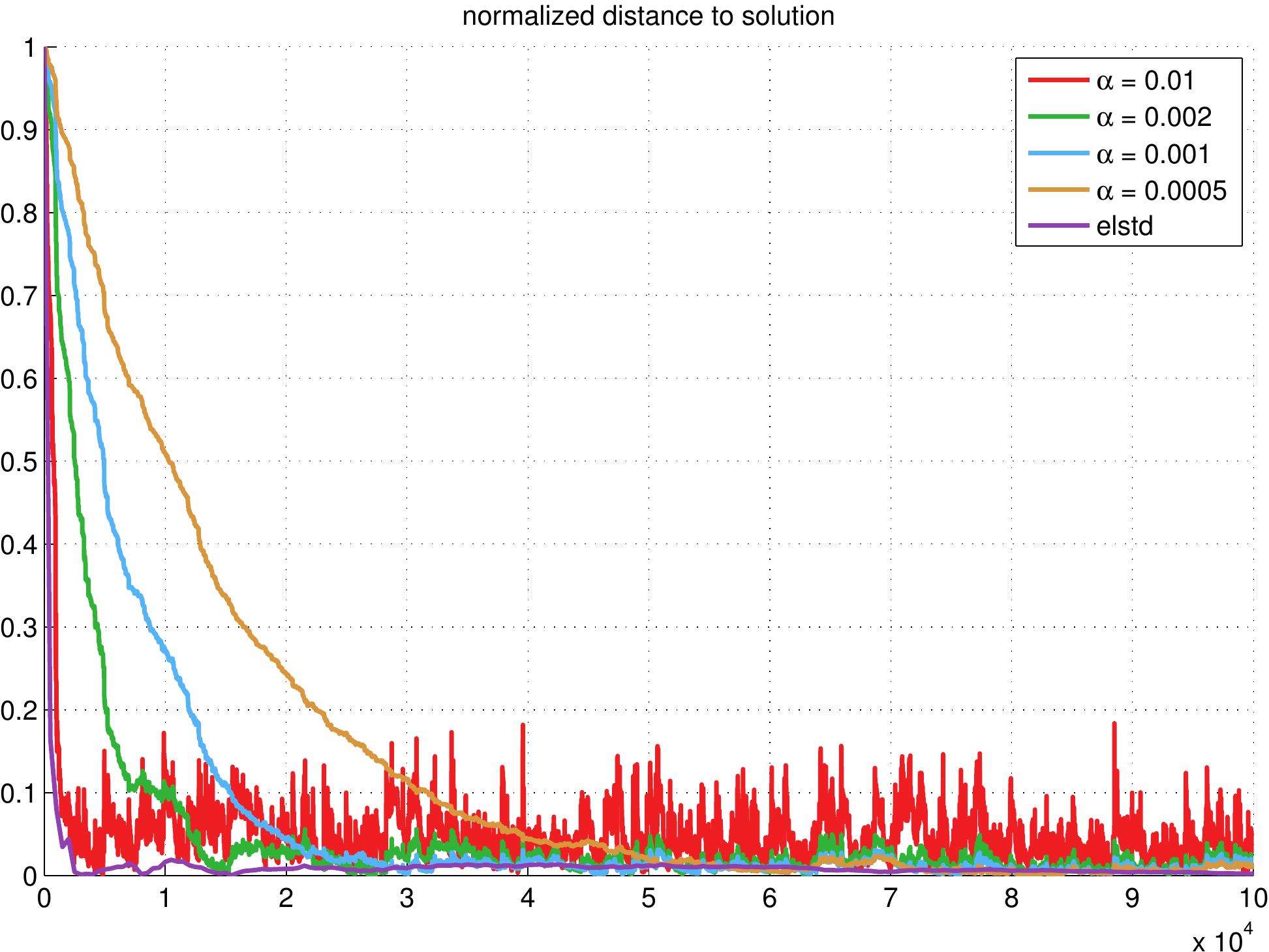} \hfill 
\includegraphics[width=0.45\linewidth]{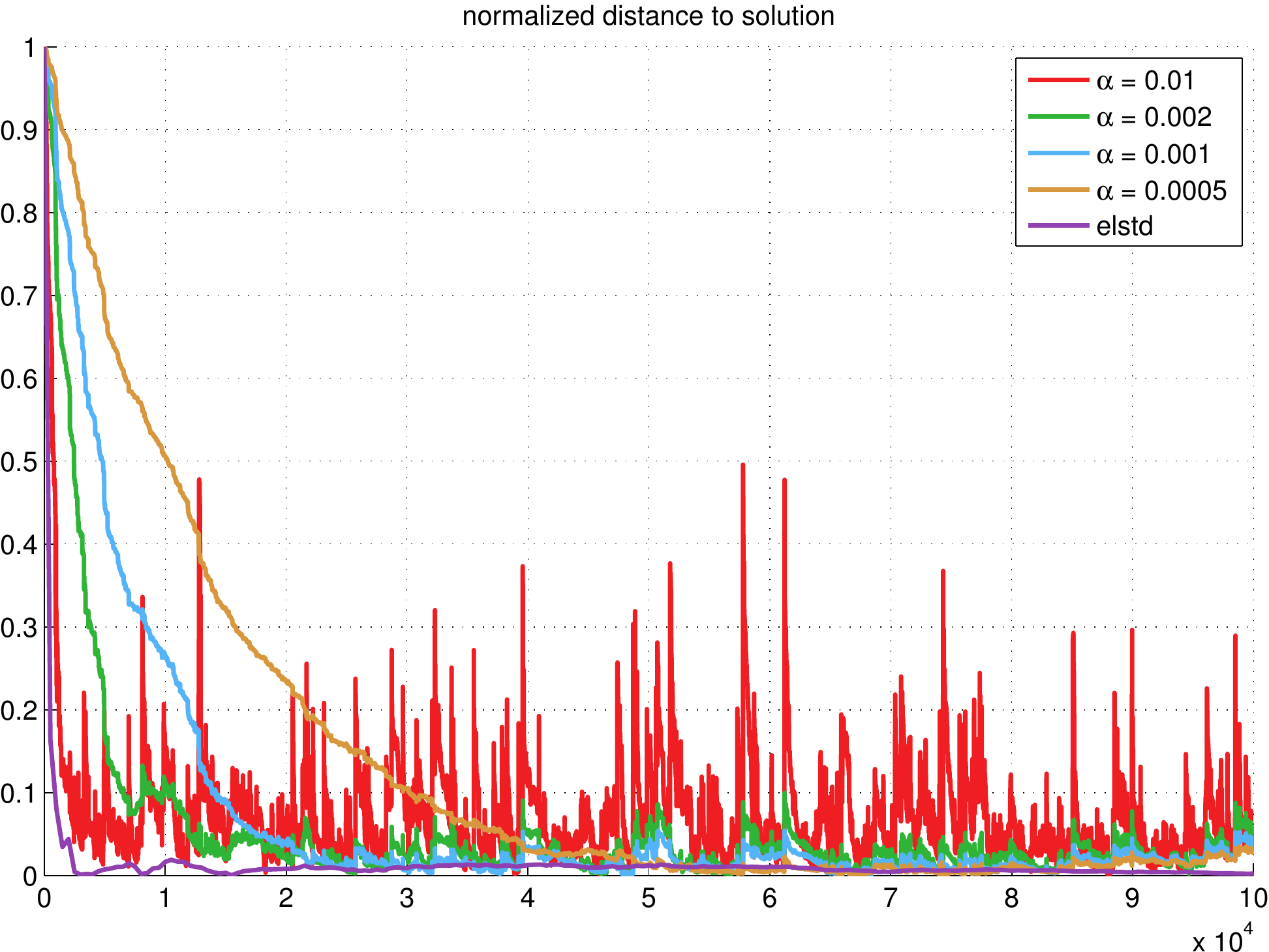} 
\caption{Variant I (left) and Variant II (right) without perturbation. Data are from a single run; ELSTD is also included for comparison.}\label{fig-cnst-ex1h}
\end{figure}

\begin{figure}[!htb] 
   \centering
\includegraphics[width=0.43\linewidth]{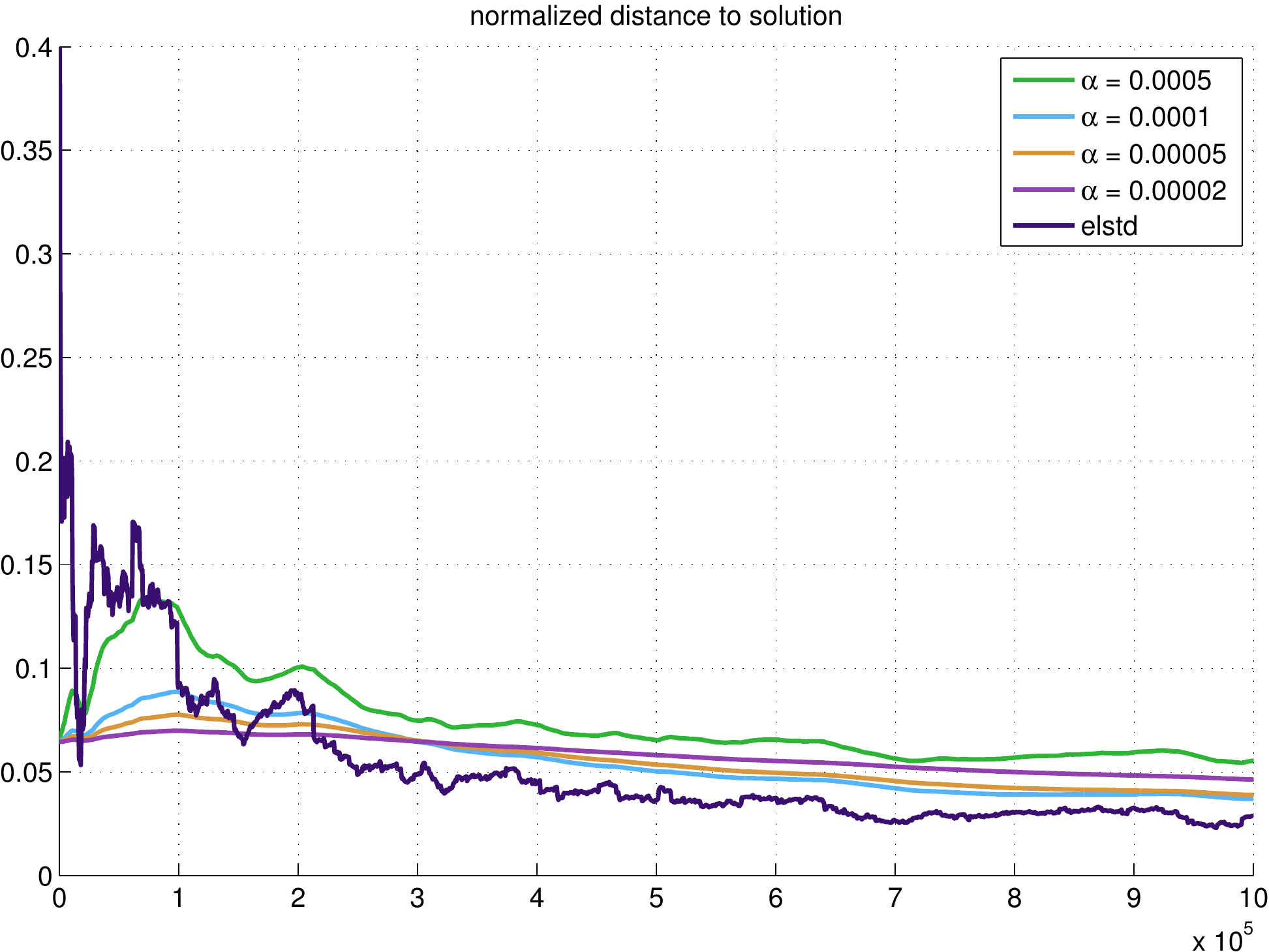} \qquad
\includegraphics[width=0.43\linewidth]{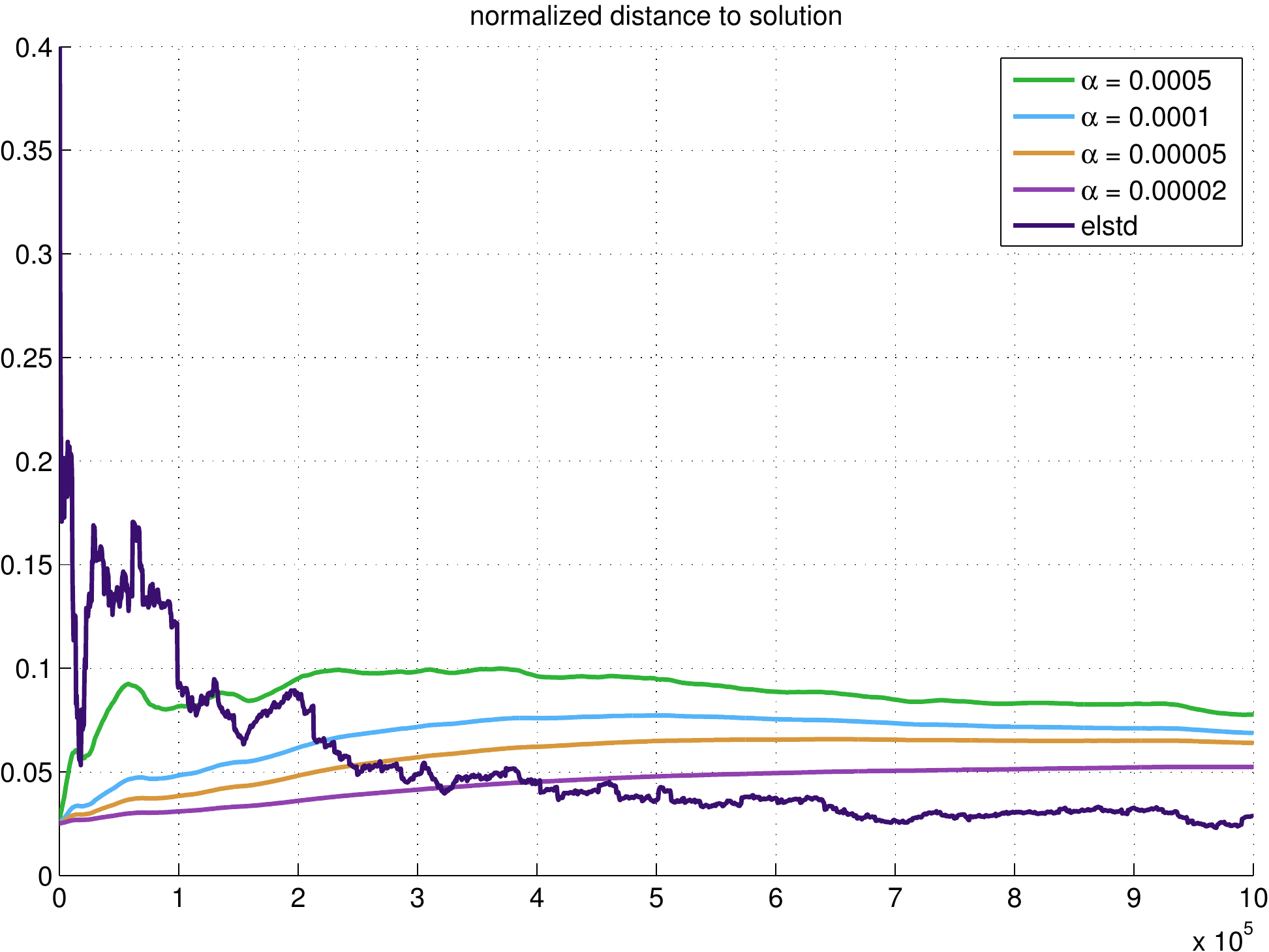}\\ 
\includegraphics[width=0.43\linewidth]{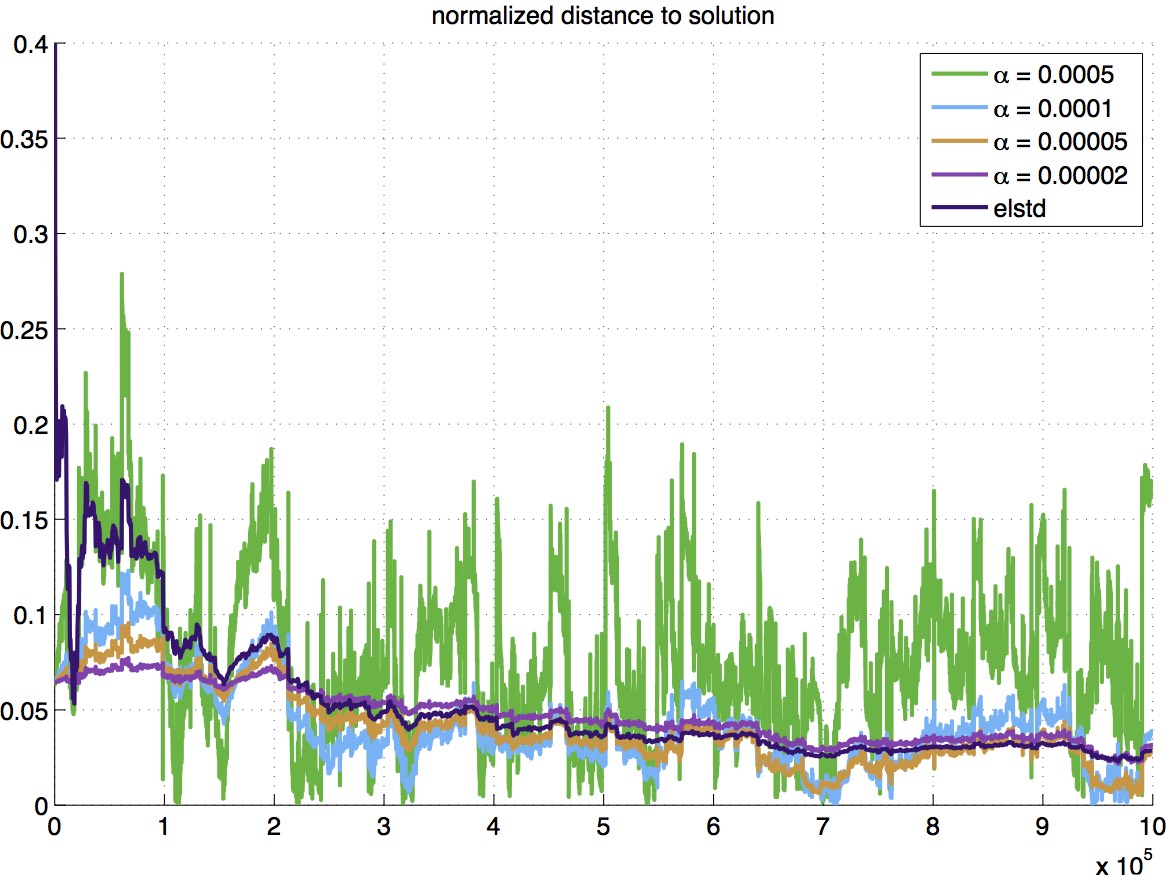} \qquad
\includegraphics[width=0.43\linewidth]{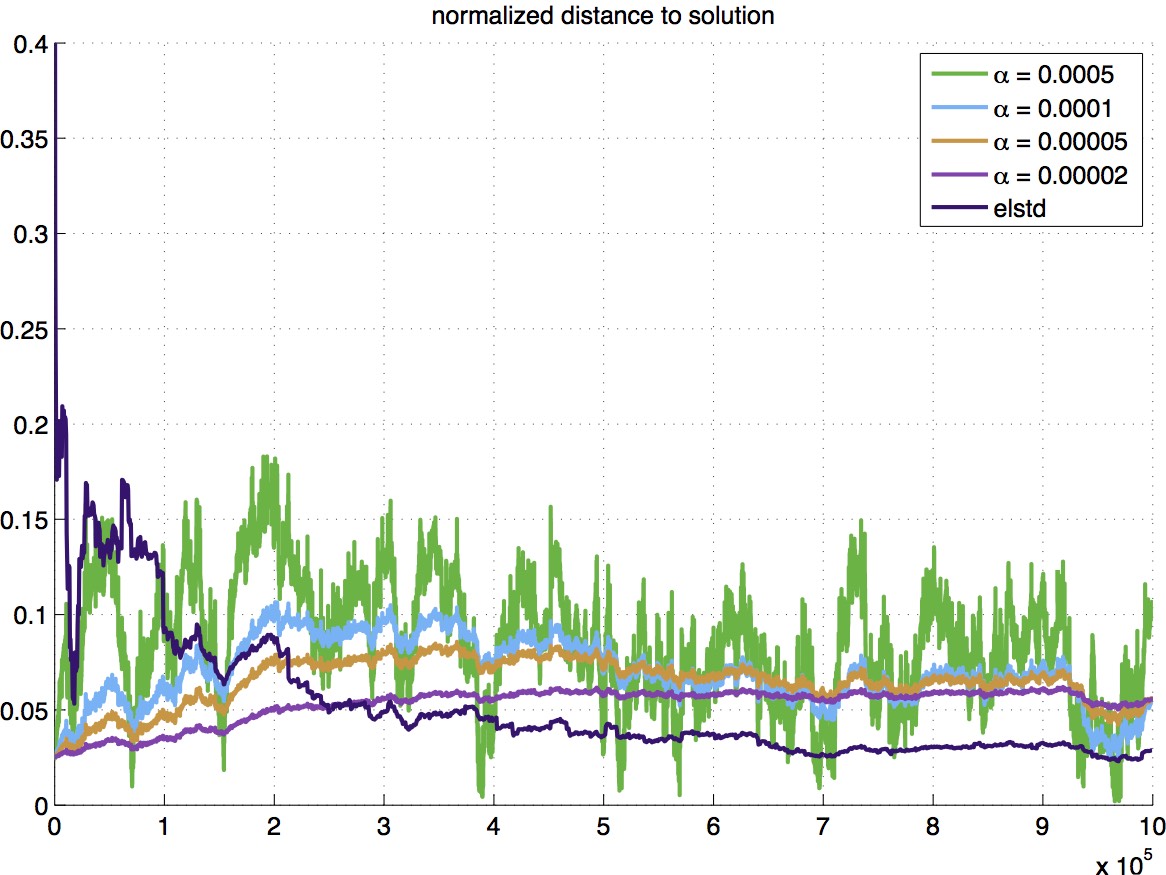} 
\caption{Variant I (left) and Variant II (right) without perturbation. Top: averaged iterates $\bar\theta_t^\alpha$; bottom: iterates $\theta_t^\alpha$. Data are from a single run; ELSTD is included in all the cases for comparison.}\label{fig-cnst-ex1i}
\end{figure}

In the rest of this subsection we show more trajectories of iterates from individual runs. The results are plotted in Figures~\ref{fig-cnst-ex1f}-\ref{fig-cnst-ex1i}, and the details of the experiments and our observations from them are as follows. 

\medskip
\noindent Figures \ref{fig-cnst-ex1f}-\ref{fig-cnst-ex1g}: In these plots we show the normalized distances (to $\theta^*$) of a trajectory of averaged iterates $\bar{\theta}_t^{\alpha}$ and original iterates $\theta_t^\alpha$, for each algorithm and each stepsize, using the data from one of the experimental runs that produced the previous four figures. Comparing the top rows with the bottom rows in Figures~\ref{fig-cnst-ex1f}-\ref{fig-cnst-ex1g}, the averaged iterates $\bar{\theta}_t^{\alpha}$ are better than $\theta_t^\alpha$ in terms of both the volatility of the iterates and the closeness to the desired solution, especially when the stepsize is relatively large. Comparing the right columns with the left ones in Figures~\ref{fig-cnst-ex1f}-\ref{fig-cnst-ex1g}, it can be seen that Variant II has a larger variance than Variant I (although we have also observed the opposite in other problems not reported in this note). Comparing Figure~\ref{fig-cnst-ex1f} with Figure~\ref{fig-cnst-ex1g}, it can be seen that for the same stepsize, the perturbed algorithms settled inside a larger neighborhood of $\theta^*$ than the unperturbed algorithms did. This suggests that the better asymptotic properties of the perturbed algorithms can be compromised by the noises brought by the perturbation (cf.\ Remark 3.2 at the end of Section 3.3 of \cite{etd-wkconv}), and the unperturbed algorithms may be adequate for practical purposes (cf.\ Remark 4.1 at the end of Section 4.3 in \cite{etd-wkconv}).

\medskip
\noindent Figure~\ref{fig-cnst-ex1h}: In this experiment we compare the transient behavior of the variant algorithms for the four stepsizes using a single run of $10^5$ iterations. All the algorithms start from the same initial condition, and no portion of the run is discarded. ELSTD is also included for comparison: the linear equations formed by ELSTD are solved every 500 iterations to produce the ELSTD curve shown in the figure. It can be seen that ELSTD converges rapidly. With a large stepsize $\alpha = 0.01$, Variants I and II also make quick initial progress, before they start to oscillate in a relatively large neighborhood of $\theta^*$.


\medskip
\noindent Figure~\ref{fig-cnst-ex1i}: This experiment serves as an example to show that the convergence behavior of the ETD algorithms are not affected when the matrix $C$ associated with ETD is negative semidefinite instead of negative definite (cf.\ \cite[Section 5.1]{etd-wkconv}). In this experiment we let $\i(s)=1$ for only two states, $s \in \{2,6\}$ (i.e., removing state $4$ from the original list of states of interest) and we set $\i(s)=0$ for the other states. Correspondingly, we set $\lambda(s)=0$ for $s \in \{2,6\}$ and $\lambda(s)=1$ otherwise. Then the $3 \times 3$ matrix $C$ has rank $2$ and becomes negative semidefinite. We ran the unperturbed Variant I and Variant II, initialized at zero, and we also ran ELSTD. The algorithms ran as in the previous cases and none of them had any issues (cf.\ the explanations given in Section 5.1 of \cite{etd-wkconv}). The iterates have higher variances in this case, so in order to obtain iterates that can approach the $0.1|\theta^*|$-neighborhood of $\theta^*$, we used a larger threshold $K=200$ in the function $\psi_K$, as well as smaller stepsizes in this experiment. (The higher variances in this case have nothing to do with the negative semidefiniteness of $C$. Instead it is a consequence of the following fact: here ETD is solving a generalized multistep Bellman equation for the two states $\{2, 6\}$ of interest, and these two states are far apart from each other in the directed transition graph. So compared with the original setting of Problem I, in this case it takes on average more steps to reach any of the states of interest again after visiting either one of them.) 

Plotted in Figure~\ref{fig-cnst-ex1i} are the normalized distances to $\theta^*$ of the averaged iterates $\bar \theta_t^\alpha$ and of the original iterates $\theta_t^\alpha$ generated in the later portion of a single run, for four choices of stepsizes. Specifically, to reduce transient effects and focus on the steady state behavior, we first ran all the algorithms for $3\times 10^5$ iterations with the stepsize $0.0005$, and we then continued the run for another $10^6$ iterations with the four different stepsizes indicated in the figure. The averaged iterates shown in the top row of the figure are generated from those later $10^6$ iterations of the run.
It can be seen that overall the behavior of the iterates is similar to what we observed in the previous experiments.

\subsection{Problem II}

\begin{figure}[!t] 
 \centering
\includegraphics[width=0.49\linewidth]{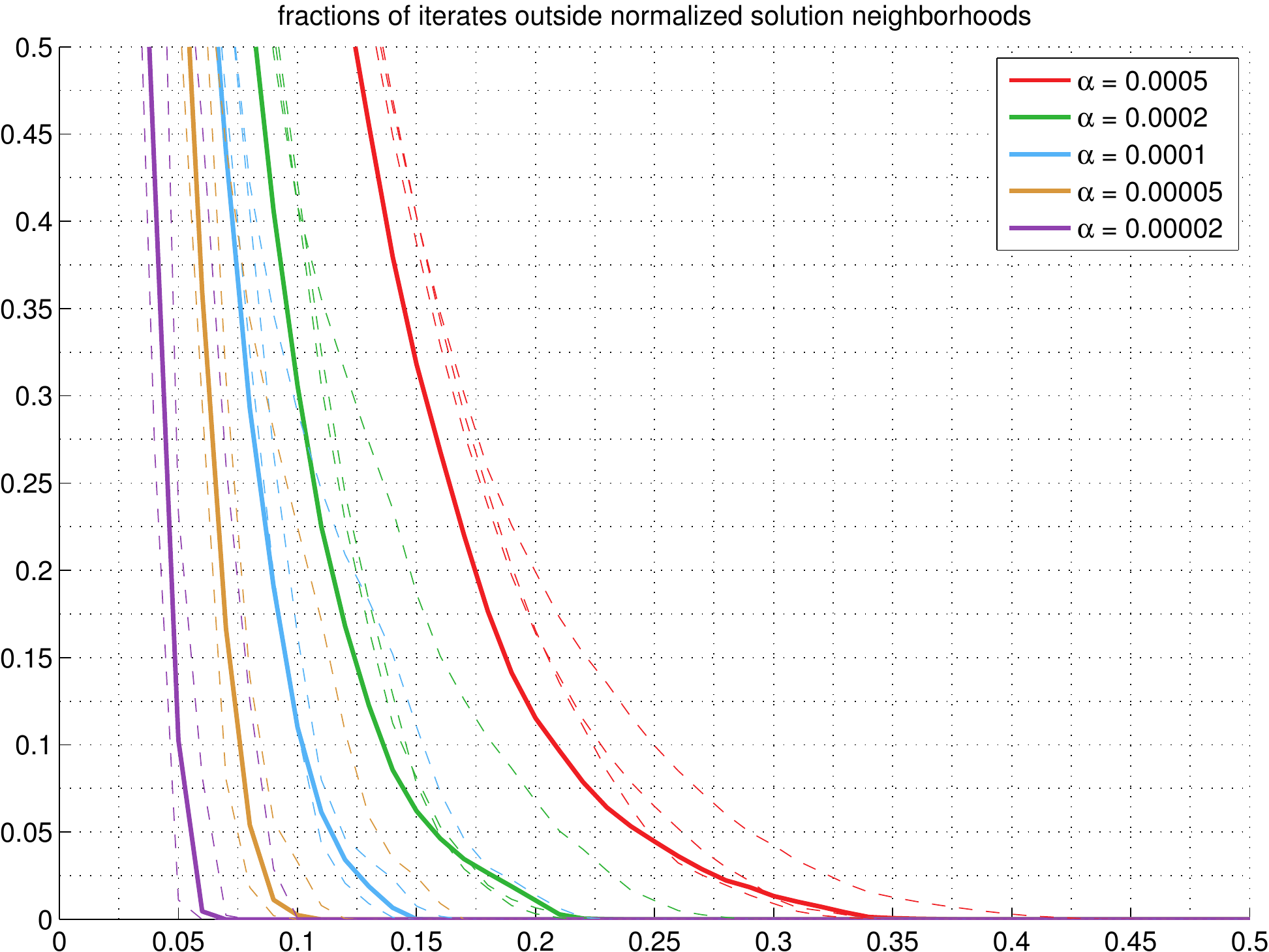} \hfill  
\includegraphics[width=0.49\linewidth]{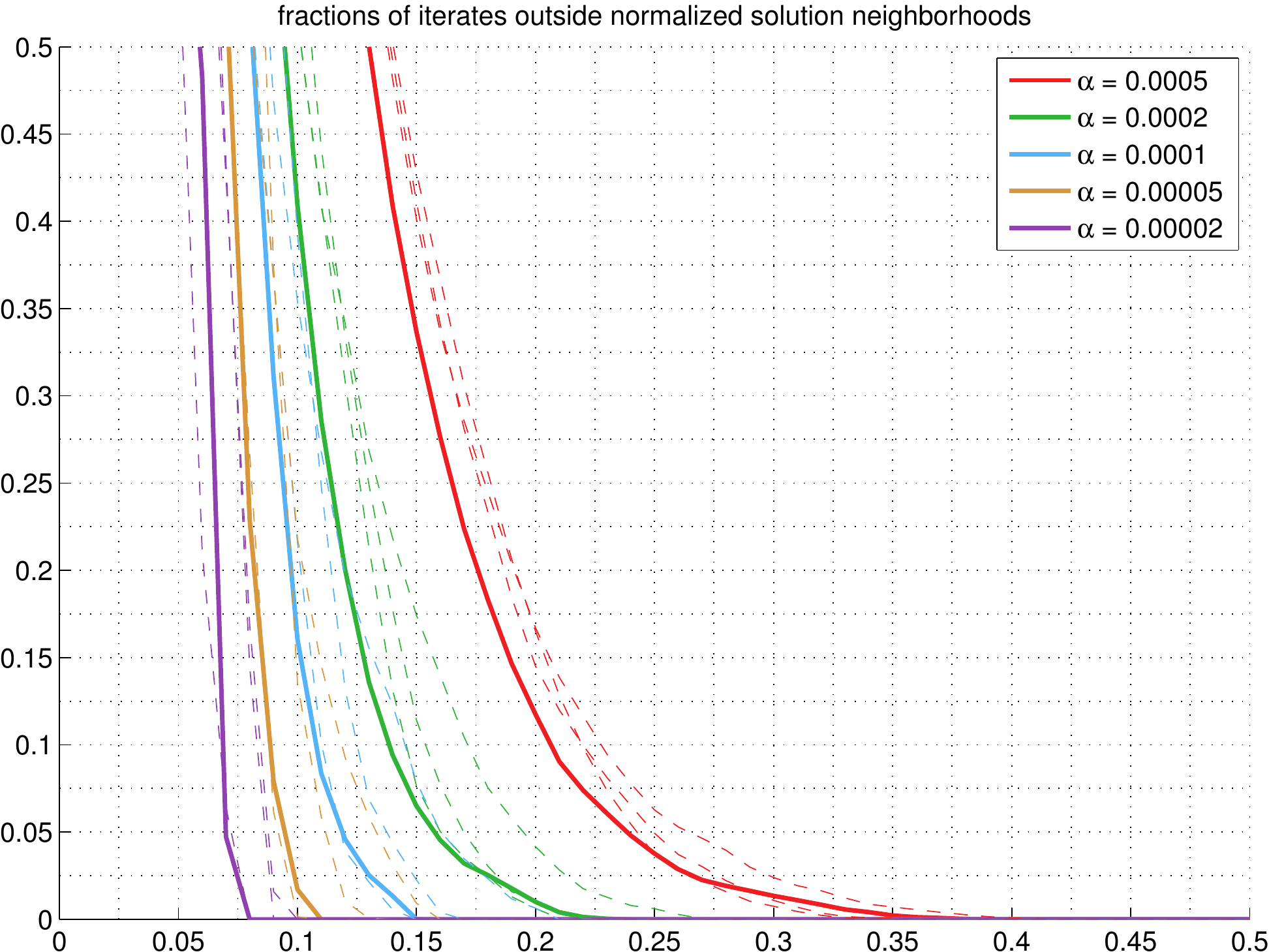}       
\caption{Variant I (left) and Variant II (right) without perturbation. The $x$-axis represents the $x |\theta^*|$-neighborhood of $\theta^*$. The $y$-component of a point $(x,y)$ represents the fraction of times (in a single run) that a segment of $100$ consecutive iterates fails to lie entirely inside the $x|\theta^*|$-neighborhood of $\theta^*$.}\label{fig-cnst-ex2b}
\end{figure}

\begin{figure}[!thb] 
\centering
\includegraphics[width=0.49\linewidth]{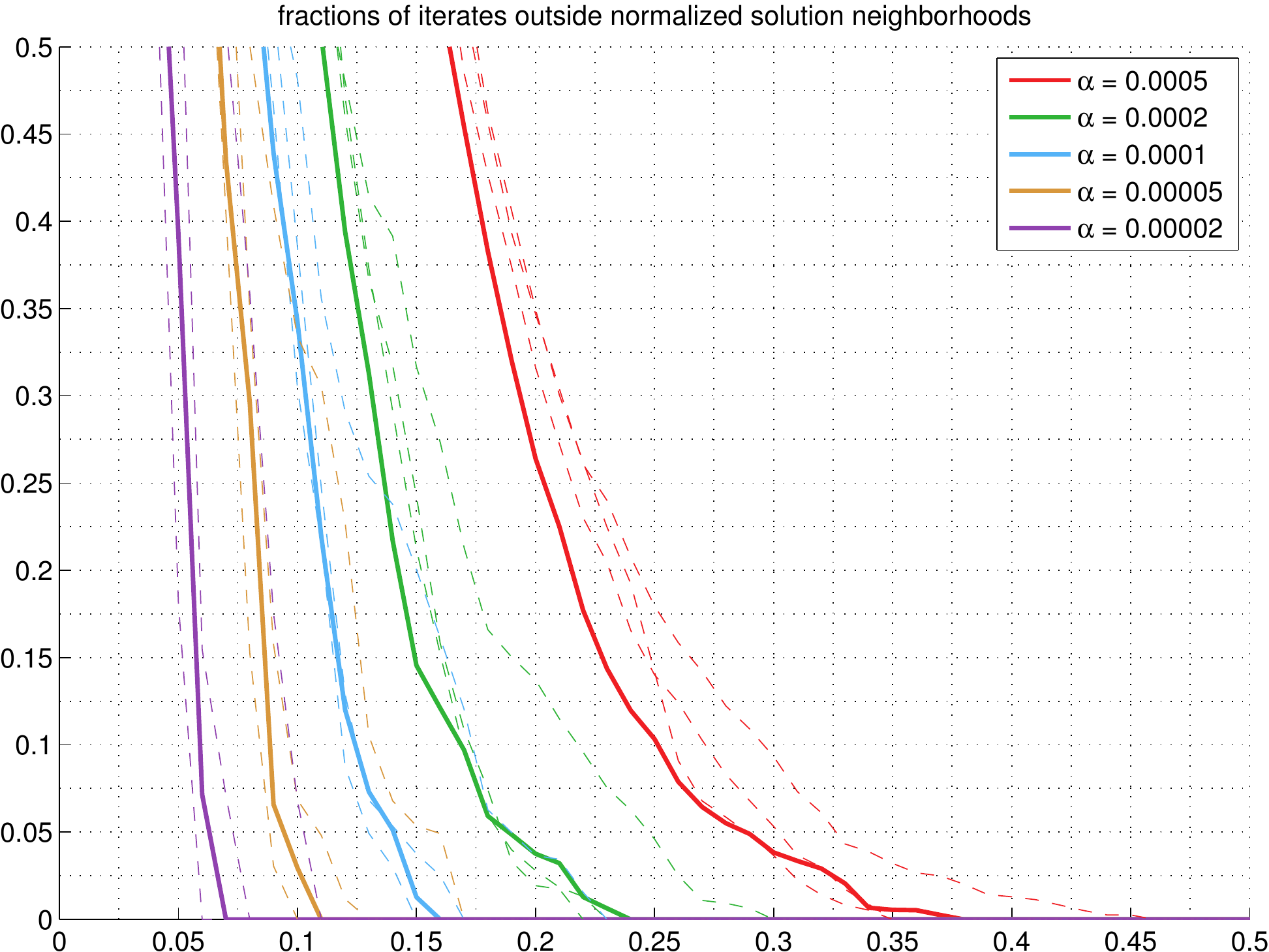} \hfill 
\includegraphics[width=0.49\linewidth]{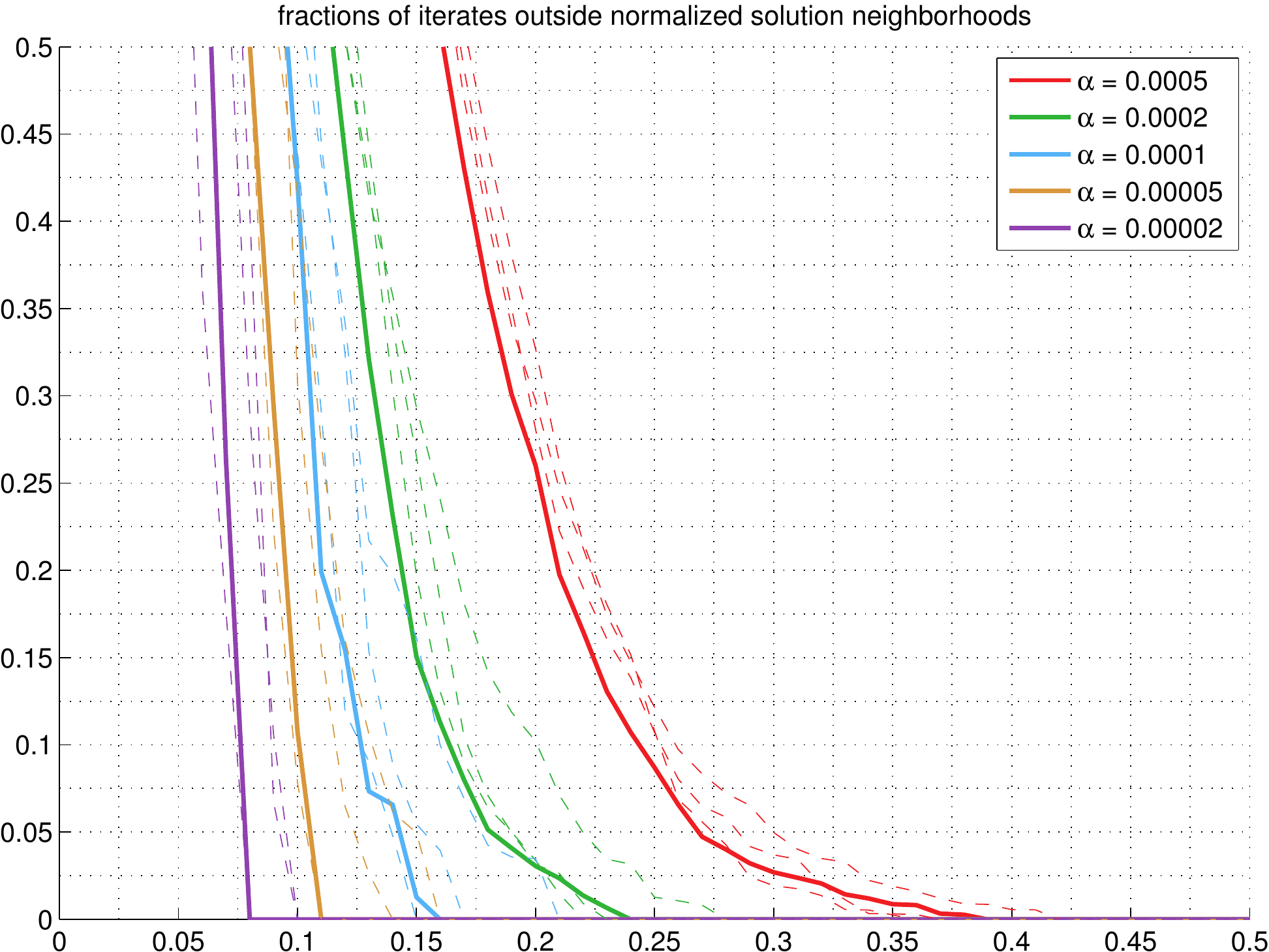}   
\caption{Variant I (left) and Variant II (right) without perturbation. The $x$-axis represents the $x |\theta^*|$-neighborhood of $\theta^*$. The $y$-component of a point $(x,y)$ represents the fraction of times (in a single run) that a segment of $\lfloor \tfrac{1}{\alpha} \rfloor$ consecutive iterates fails to lie entirely inside the $x|\theta^*|$-neighborhood of $\theta^*$.}\label{fig-cnst-ex2c}
\end{figure}

\begin{figure}[!tb] 
   \centering
\includegraphics[width=0.49\linewidth]{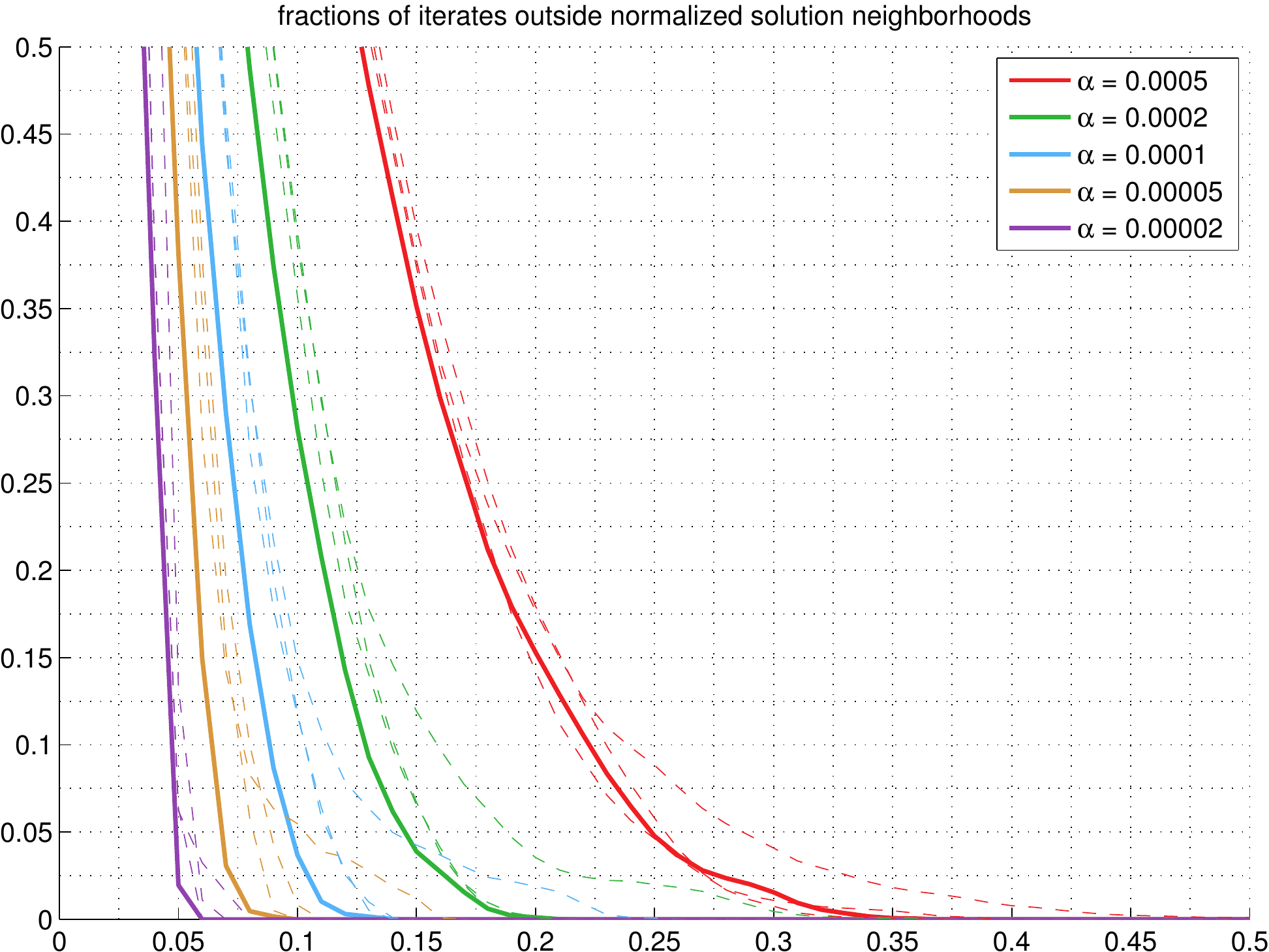} \hfill   
\includegraphics[width=0.49\linewidth]{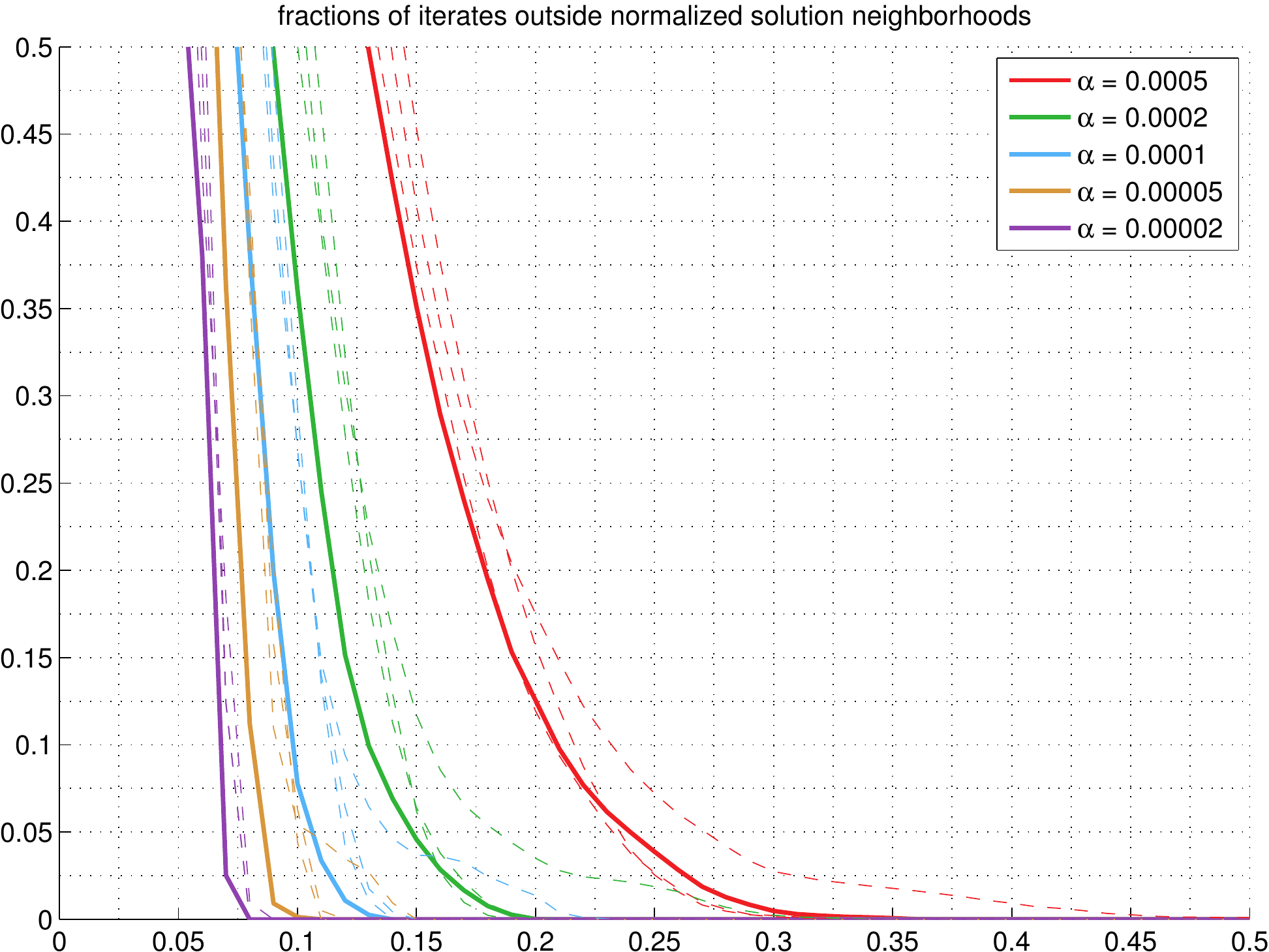}        
\caption{Variant I (left) and Variant II (right) with perturbation. The $x$-axis represents the $x |\theta^*|$-neighborhood of $\theta^*$. The $y$-component of a point $(x,y)$ represents the fraction of times (in a single run) that a segment of $100$ consecutive iterates fails to lie entirely inside the $x|\theta^*|$-neighborhood of $\theta^*$.} \label{fig-cnst-ex2d}
\end{figure}

\begin{figure}[!htb] 
   \centering
\includegraphics[width=0.49\linewidth]{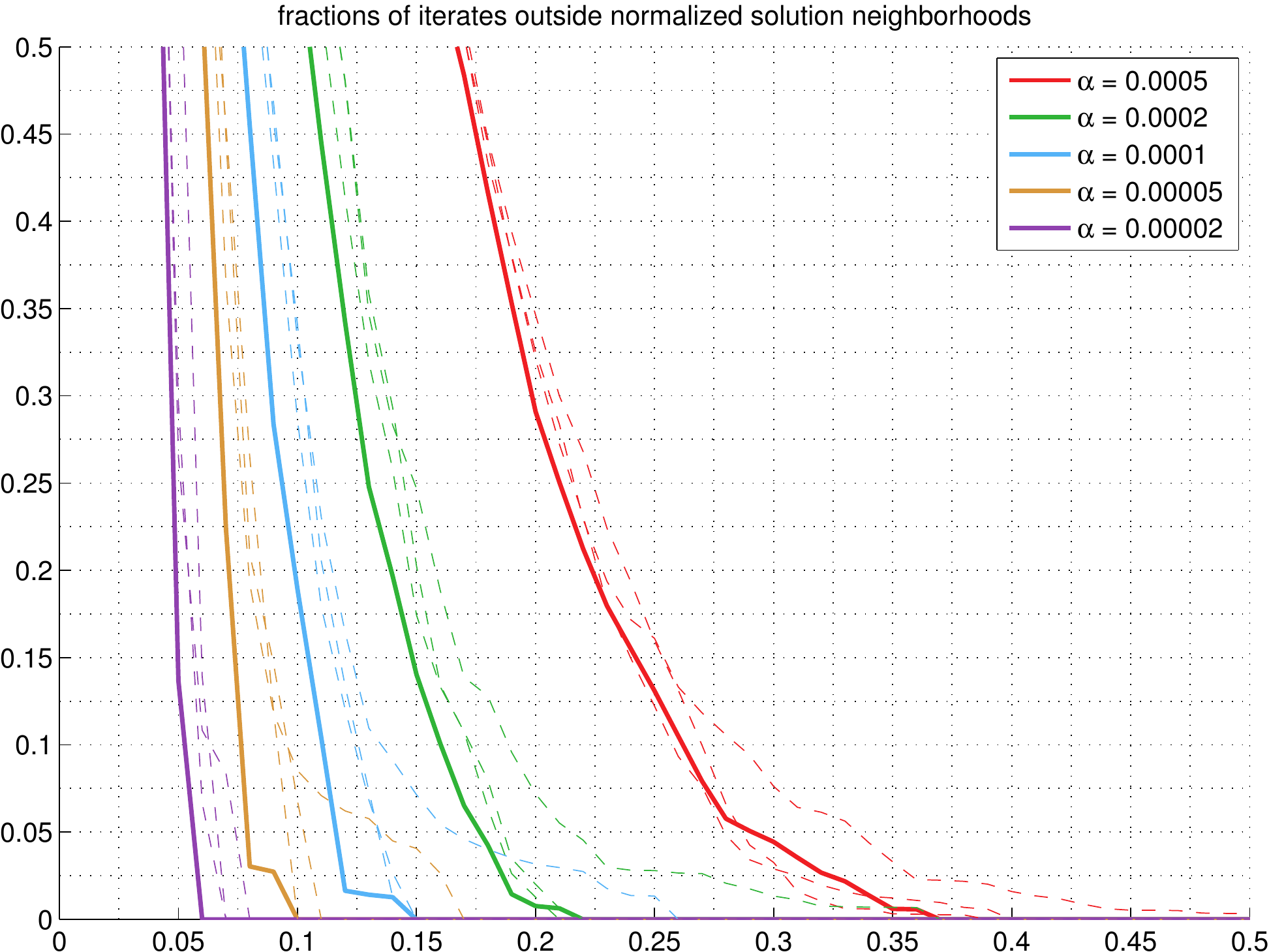} \hfill
\includegraphics[width=0.49\linewidth]{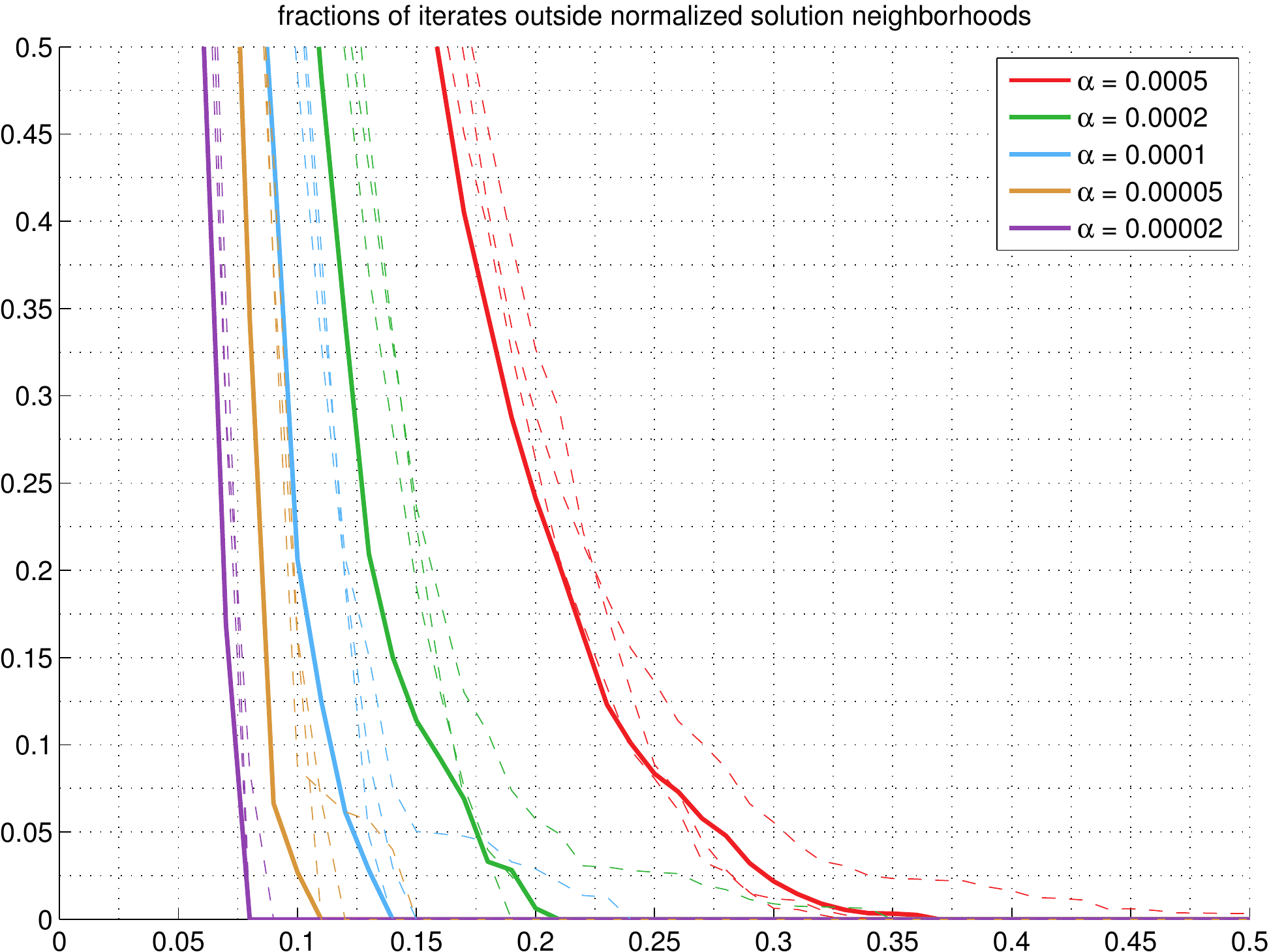}
\caption{Variant I (left) and Variant II (right) with perturbation. The $x$-axis represents the $x |\theta^*|$-neighborhood of $\theta^*$. The $y$-component of a point $(x,y)$ represents the fraction of times (in a single run) that a segment of $\lfloor \tfrac{1}{\alpha} \rfloor$ consecutive iterates fails to lie entirely inside the $x|\theta^*|$-neighborhood of $\theta^*$.}\label{fig-cnst-ex2e}
\end{figure}

We repeated for Problem II the same experiments we did for Problem I. 
All the algorithms are tested for five stepsizes: $\alpha = 0.0005, 0.0002, 0.0001, 0.00005, 0.00002$. 
First, we did $4$ independent runs of Variants I and II and their perturbed versions. 
Each run has $11 \times 10^5$ iterations, and the last $8 \times 10^5$ iterations are used to obtain the statistics of multiple consecutive iterates shown in Figures~\ref{fig-cnst-ex2b}-\ref{fig-cnst-ex2e}, in order to show the stead state behavior of the algorithms. More details are as follows.

\begin{figure}[!t] 
\centering
\includegraphics[width=0.4\linewidth]{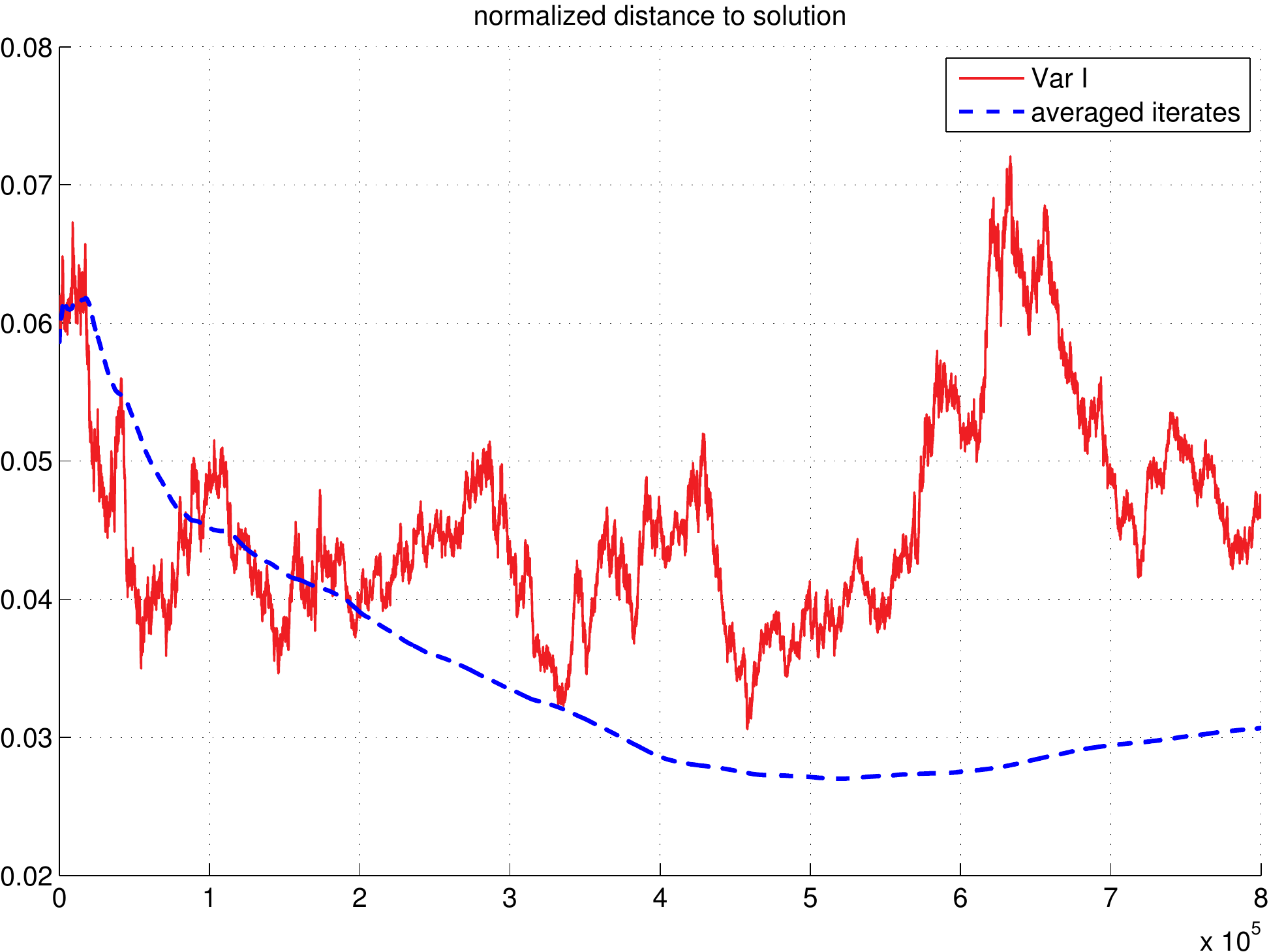} \qquad 
\includegraphics[width=0.4\linewidth]{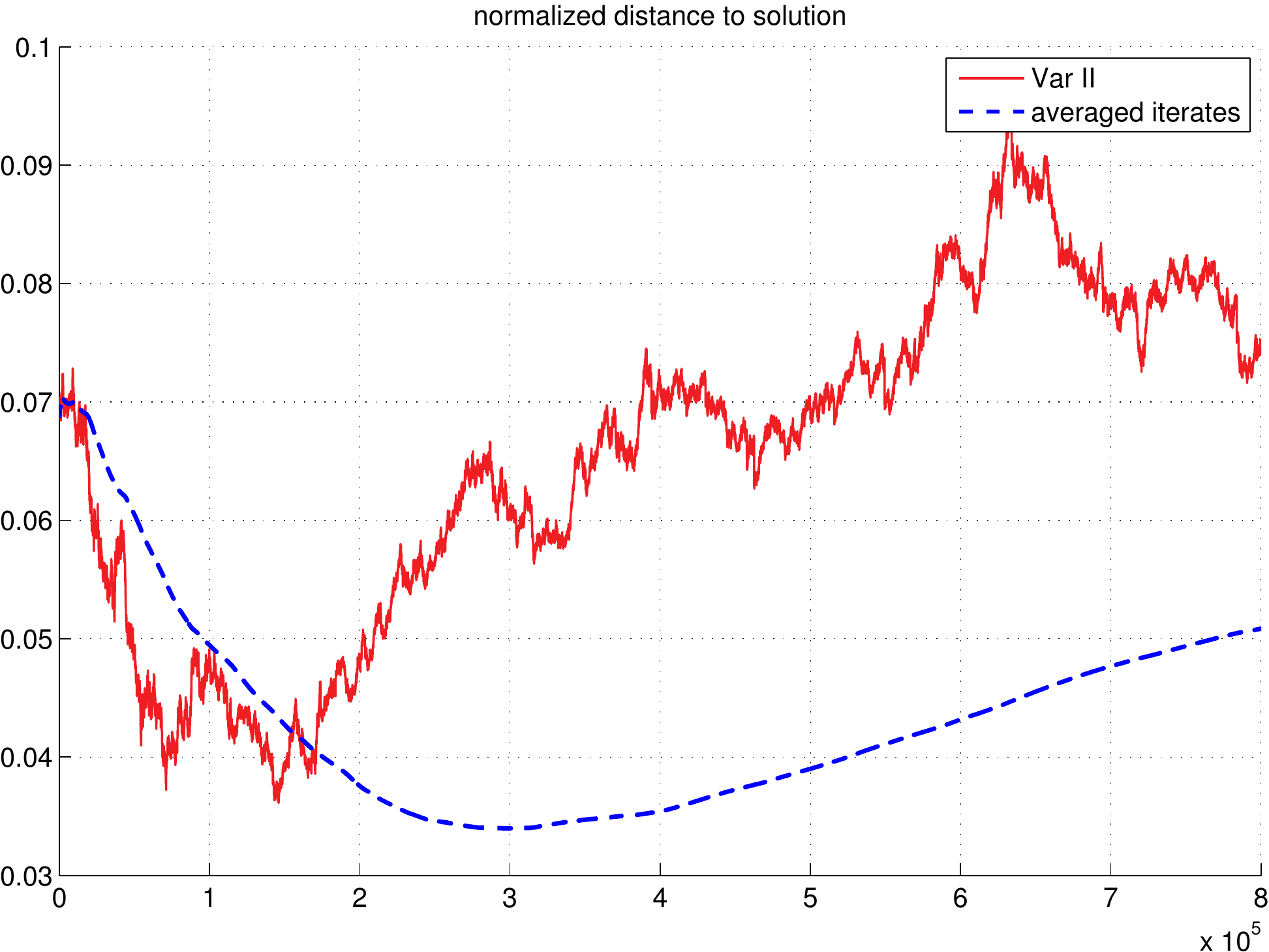}\\*[0.1cm] 
\includegraphics[width=0.4\linewidth]{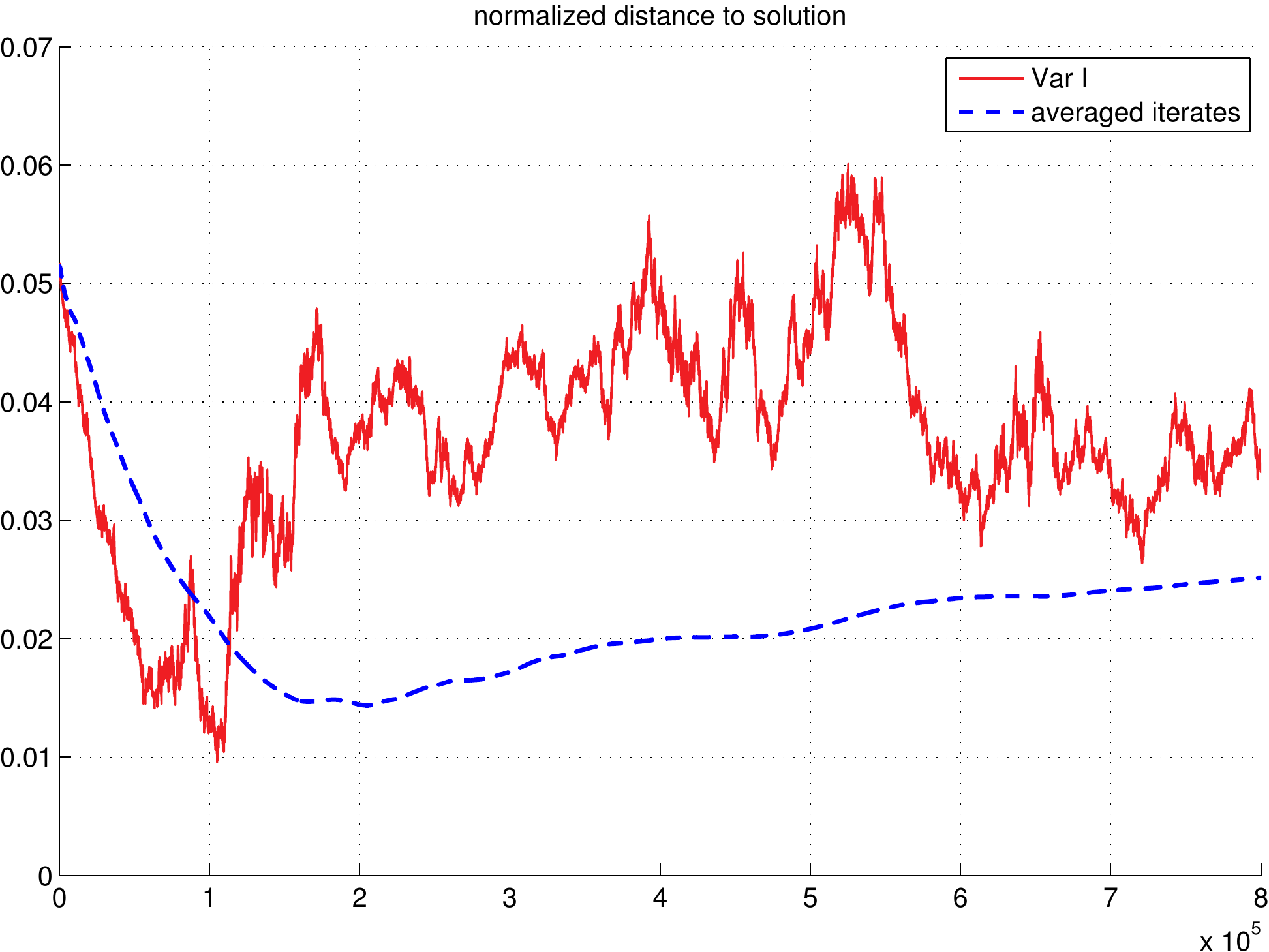} \qquad 
\includegraphics[width=0.4\linewidth]{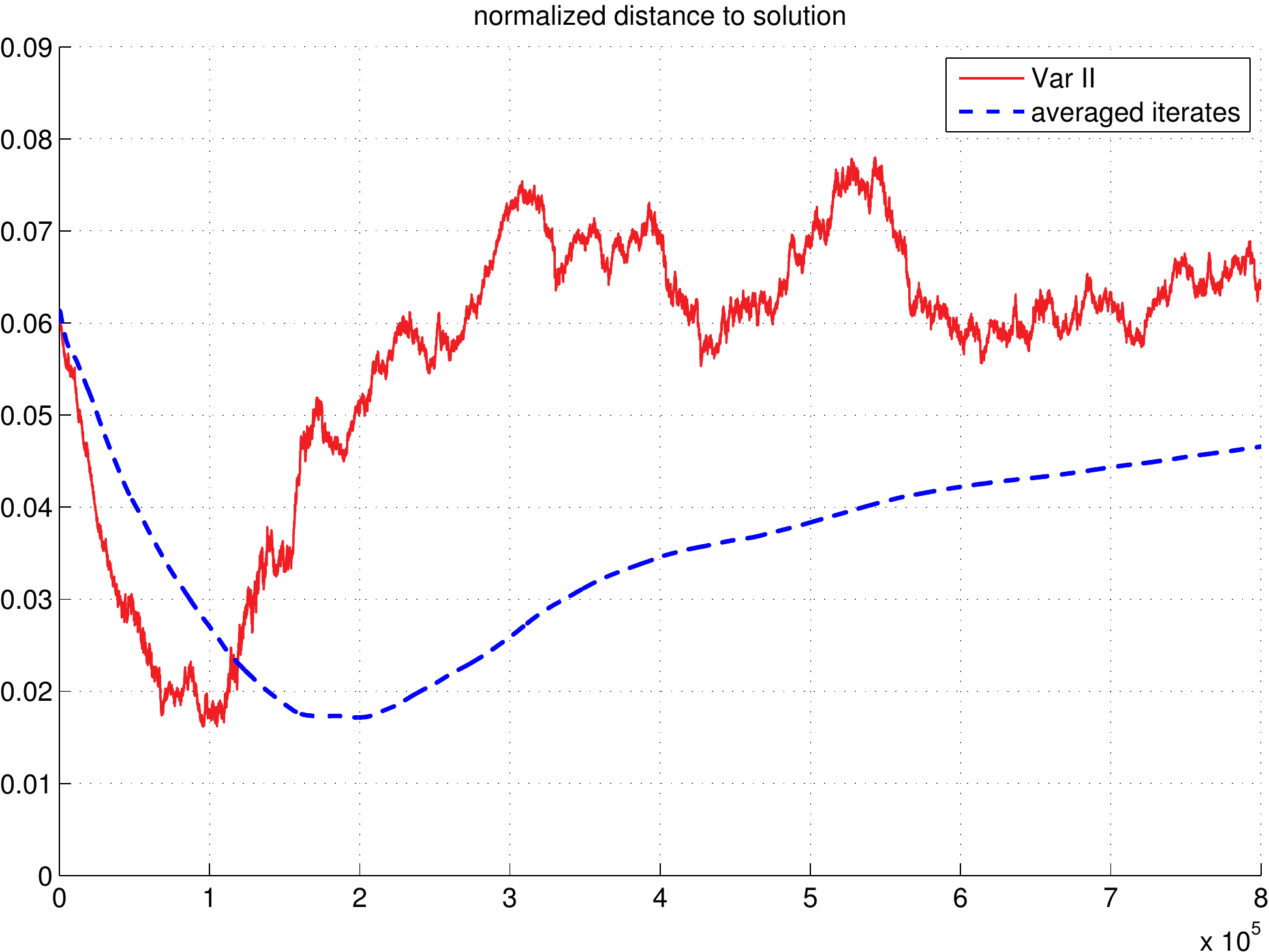}  
\caption{Variants I and II without (top) and with (bottom) perturbation ($\alpha = 0.00002$).} \label{fig-cnst-ex2a}
\end{figure}

\medskip
\noindent Figure~\ref{fig-cnst-ex2a}: This figure shows an example trajectory from a single run. 
Plotted in the figure are the normalized distances to $\theta^*$ of the iterates $\theta_t^\alpha$ generated by the four algorithms for the smallest stepsize $\alpha = 0.00002$.
The dashed lines correspond to the averaged iterates $\bar \theta_t^\alpha$, which, like in the case of Problem I, can be seen to behave better than the original iterates $\theta^\alpha_t$. We ran ELSTD to get an estimate of the degree of bias of Variant I. Averaged over $8$ independent runs of $8 \times 10^5$ iterations each, the mean normalized distance of the ELSTD final solution to $\theta^*$ was $0.043$ with standard deviation $0.003$. So based on Figure~\ref{fig-cnst-ex2a}, it seems that the iterates generated by Variant I with the stepsize $\alpha = 0.00002$ are not far from the smallest neighborhood that Variant I can reach.

\medskip
\noindent Figures~\ref{fig-cnst-ex2b}-\ref{fig-cnst-ex2e}: These figures show the behavior of multiple consecutive iterates for the four algorithms with different stepsizes.
For Variants I and II, plotted in Figure~\ref{fig-cnst-ex2b} are the fractions of times (during a single run) that a segment of length $100$, $(\theta_{t}^\alpha, \ldots, \theta_{t+99}^\alpha)$, fails to lie entirely inside the $x |\theta^*|$-neighborhood, and plotted in Figure~\ref{fig-cnst-ex2c} are the fractions of times (during a single run) that a segment of length $\lfloor \tfrac{1}{\alpha} \rfloor$, $\big(\theta_{t}^\alpha, \ldots, \theta^\alpha_{t+\lfloor 1/\alpha \rfloor-1} \big)$, fails to lie entirely inside the $x |\theta^*|$-neighborhood.
For the perturbed version of Variant I and Variant II, the same plots are shown in Figure~\ref{fig-cnst-ex2d} and Figure~\ref{fig-cnst-ex2e}, respectively.
In all these figures, for each color and each algorithm, the solid lines correspond to the results from one of the four runs, and the dashed lines the other three runs. 
The behavior exhibited here is similar to what we observed in the case of Problem I:  as the stepsize becomes smaller, the iterates spread out less and the trajectory spends more time inside a smaller neighborhood of $\theta^*$. This can be compared with the assertions in Theorem~3.4(ii) and Theorem~3.6(i) of \cite{etd-wkconv} for Variants I and II, and with the assertions in Theorem~3.8 of \cite{etd-wkconv} for the perturbed versions of these two variants.

\clearpage

\clearpage

\clearpage

In the rest of this subsection we show more trajectories of iterates from individual runs. The details of the experiments are as follows. 

\begin{figure}[!htb] 
   \centering
\includegraphics[width=0.49\linewidth]{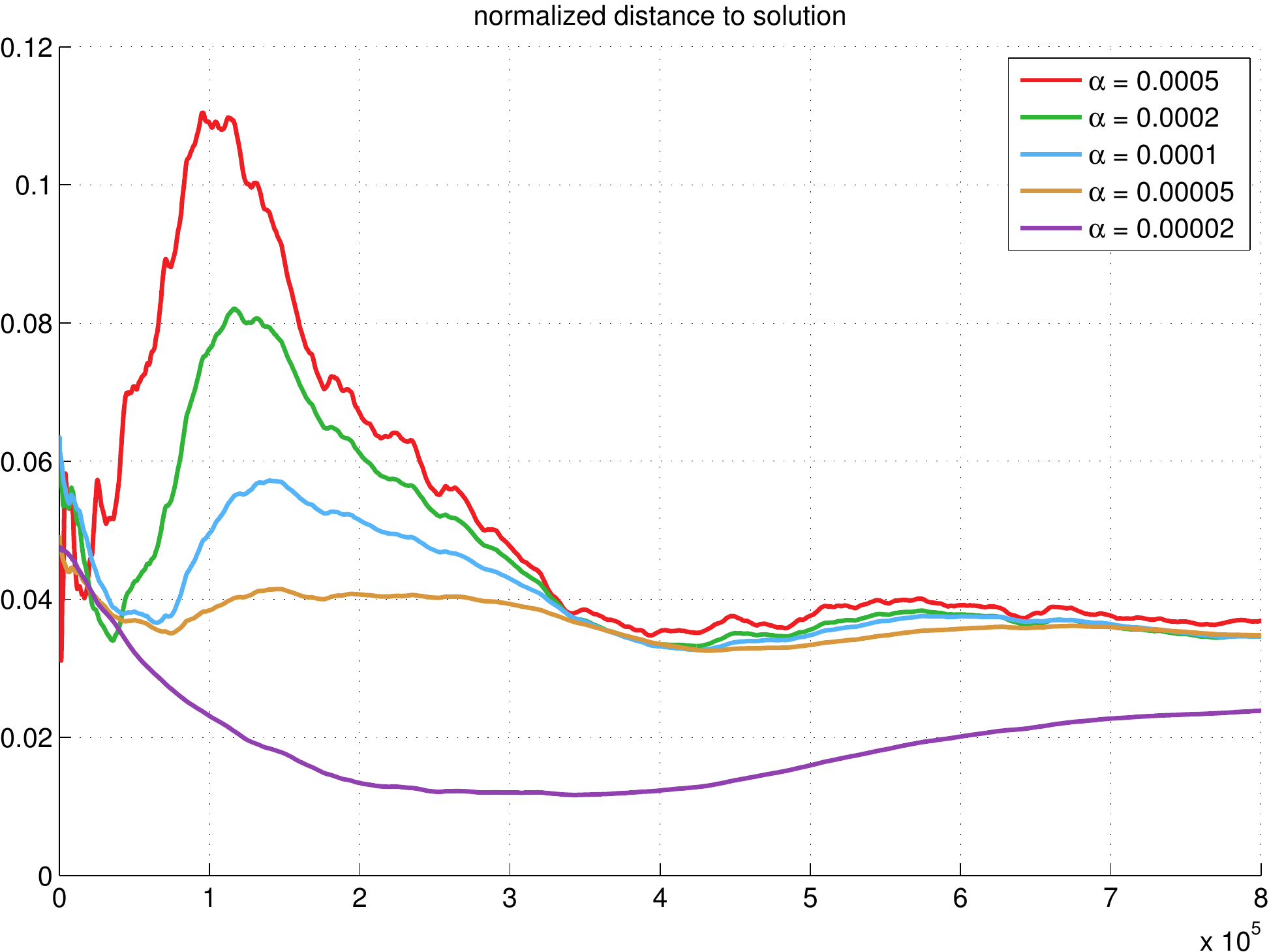} \hfill  
\includegraphics[width=0.49\linewidth]{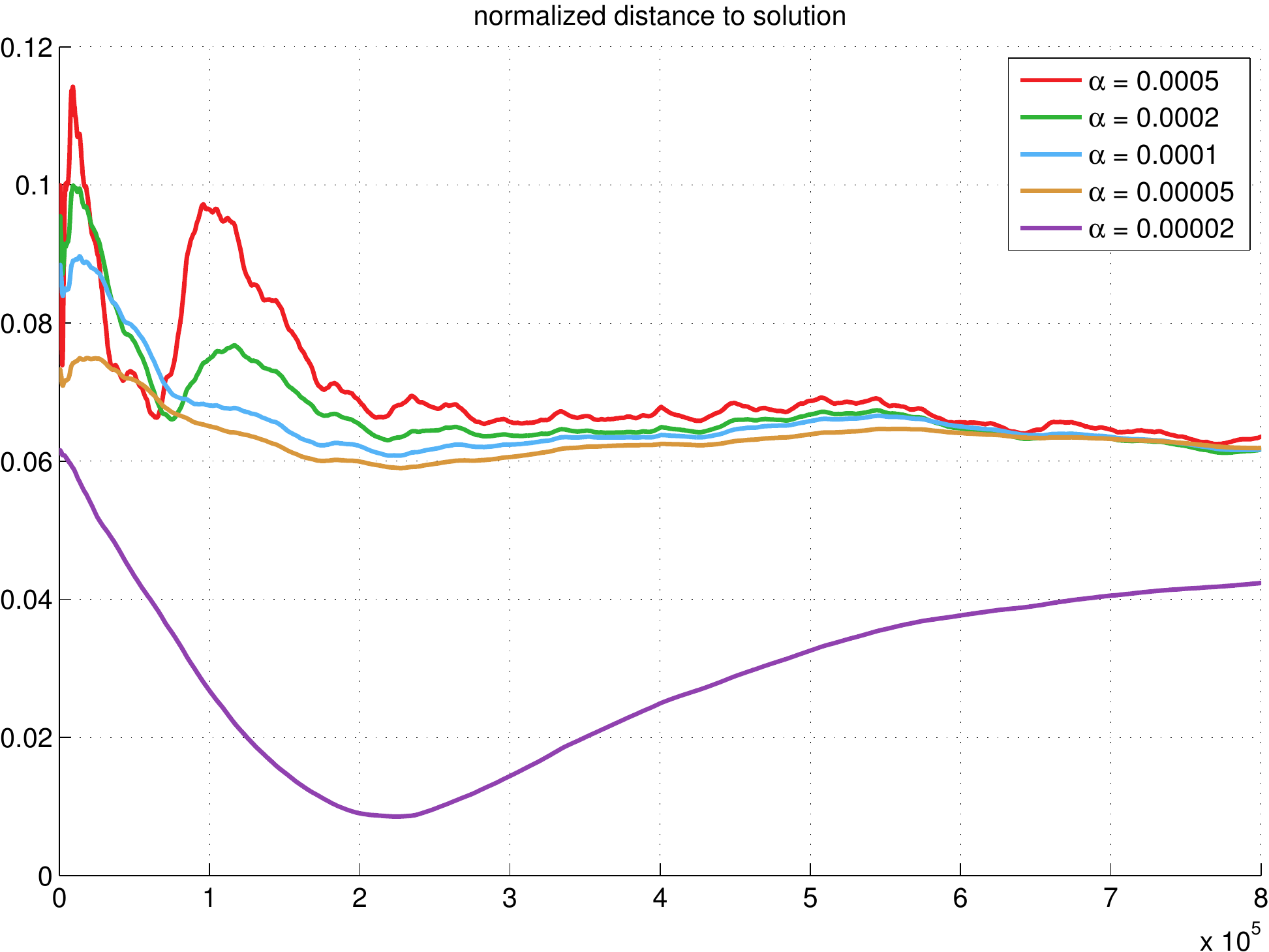}\\ 
\includegraphics[width=0.49\linewidth]{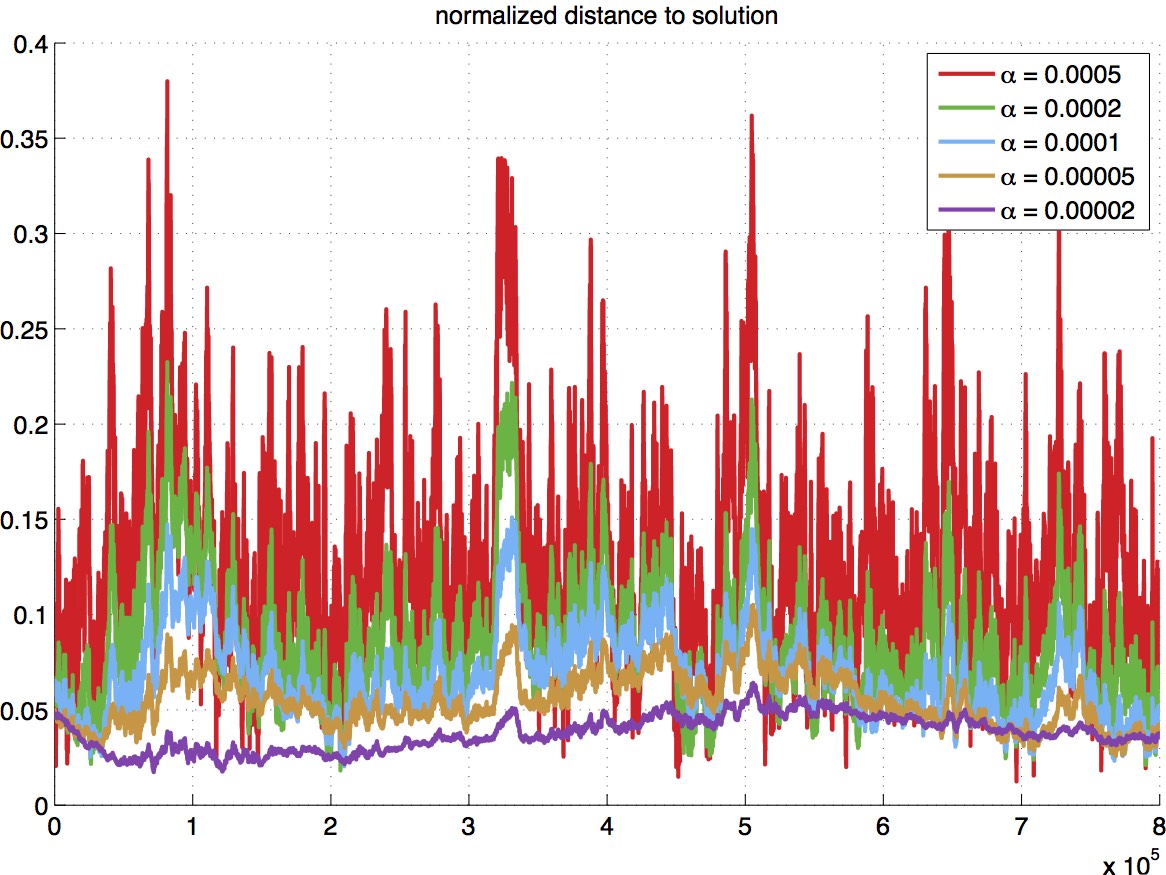} \hfill 
\includegraphics[width=0.49\linewidth]{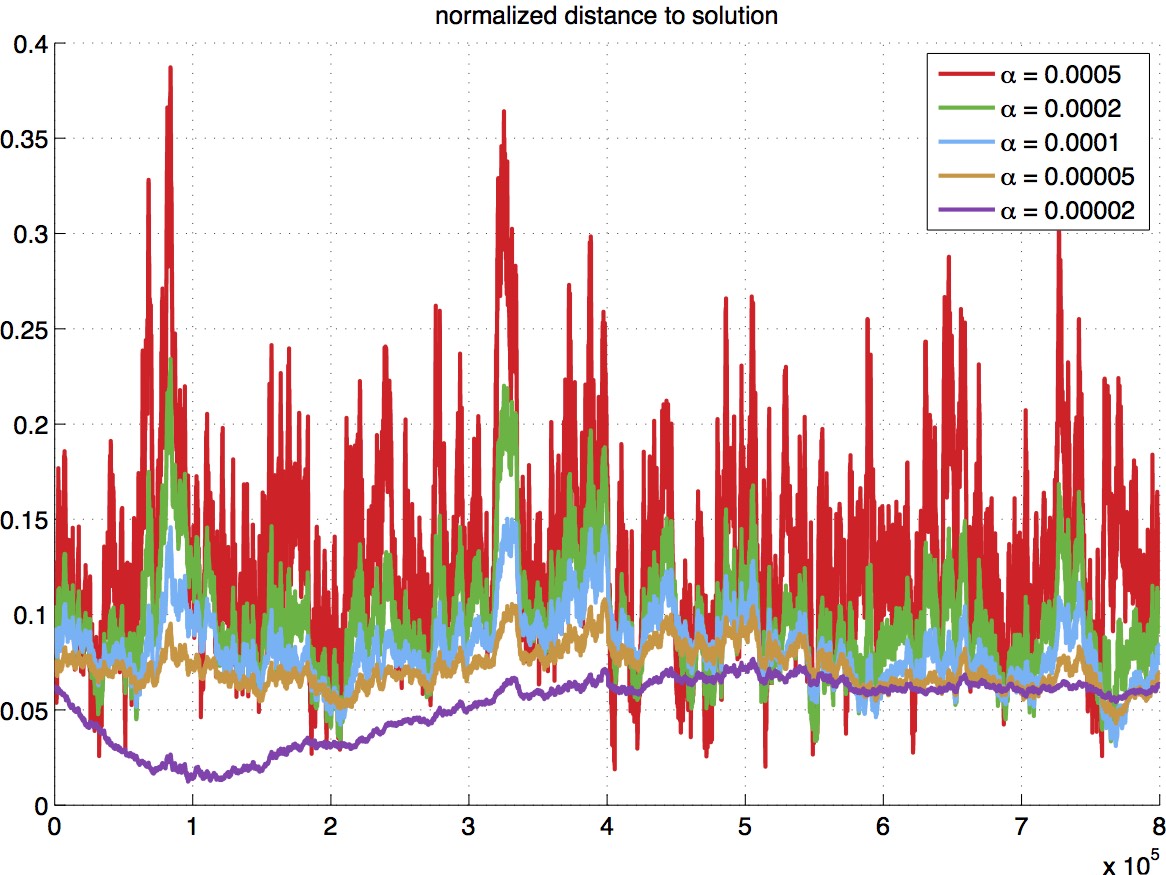} 
\caption{Variant I (left) and Variant II (right) without perturbation. Top: averaged iterates $\bar\theta^\alpha_t$; bottom: iterates $\theta_t^\alpha$. Data are from a single run.}\label{fig-cnst-ex2f}
\end{figure}

\begin{figure}[!t] 
   \centering
\includegraphics[width=0.49\linewidth]{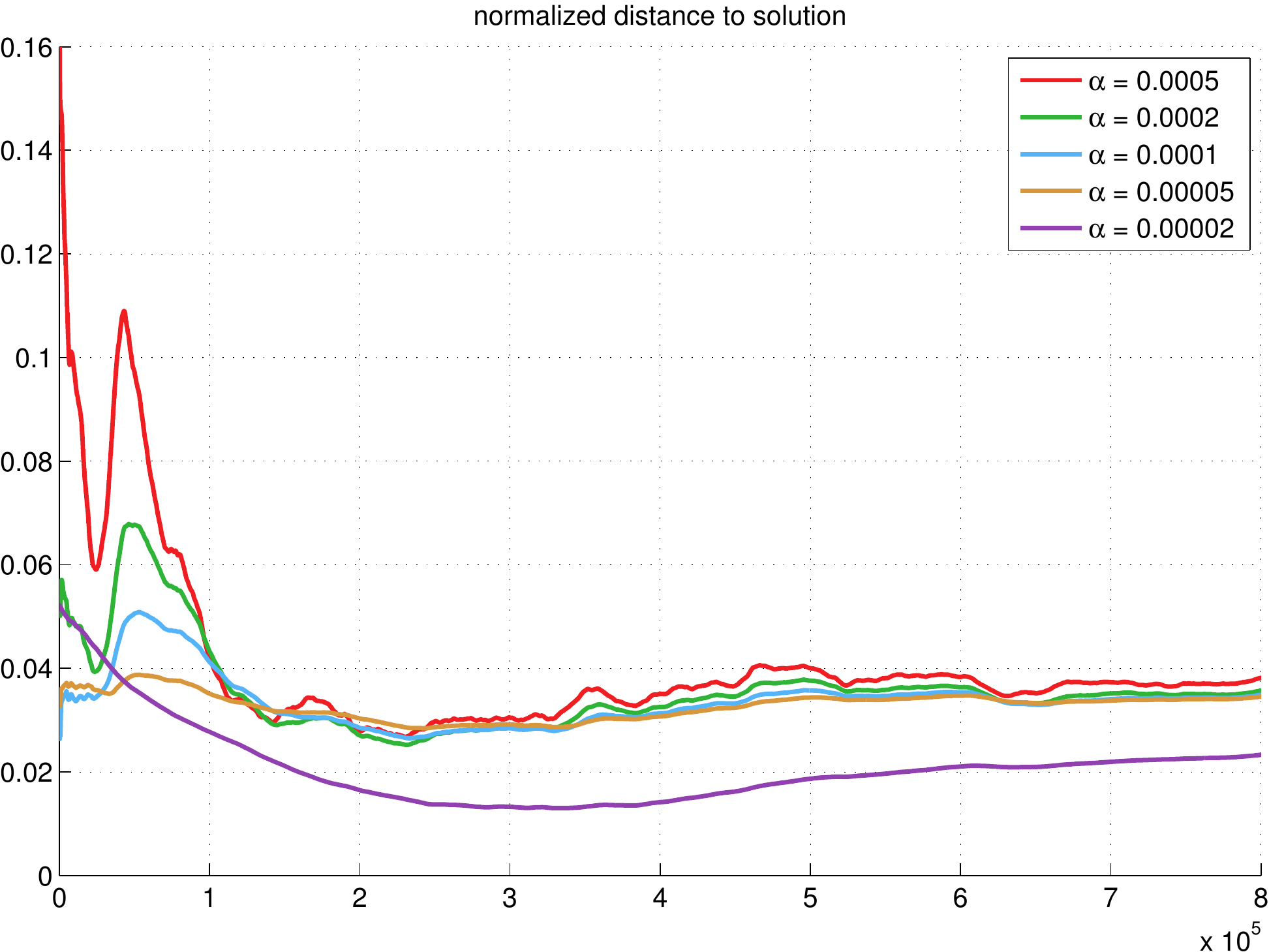} \hfill  
\includegraphics[width=0.49\linewidth]{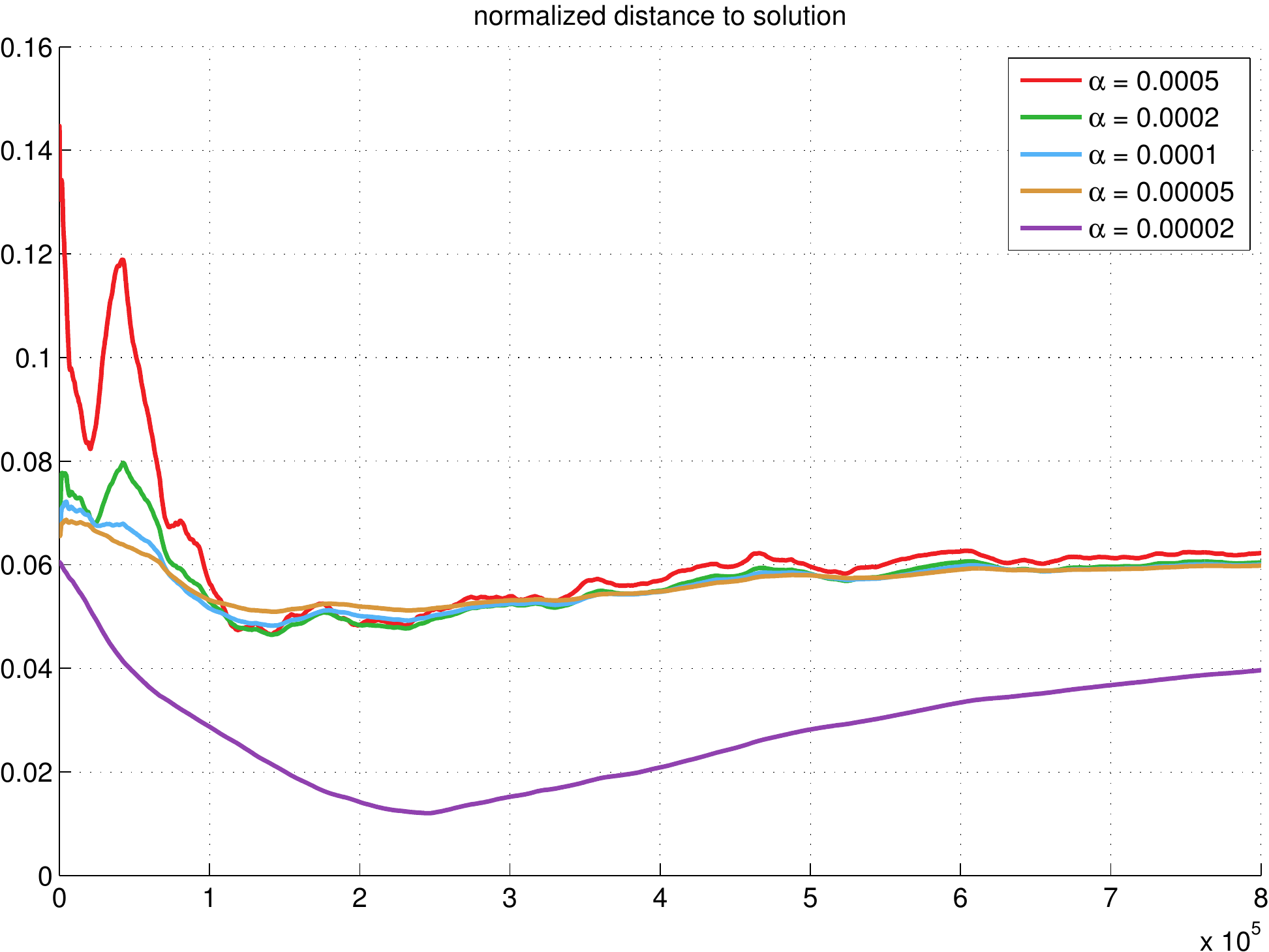}\\ 
\includegraphics[width=0.49\linewidth]{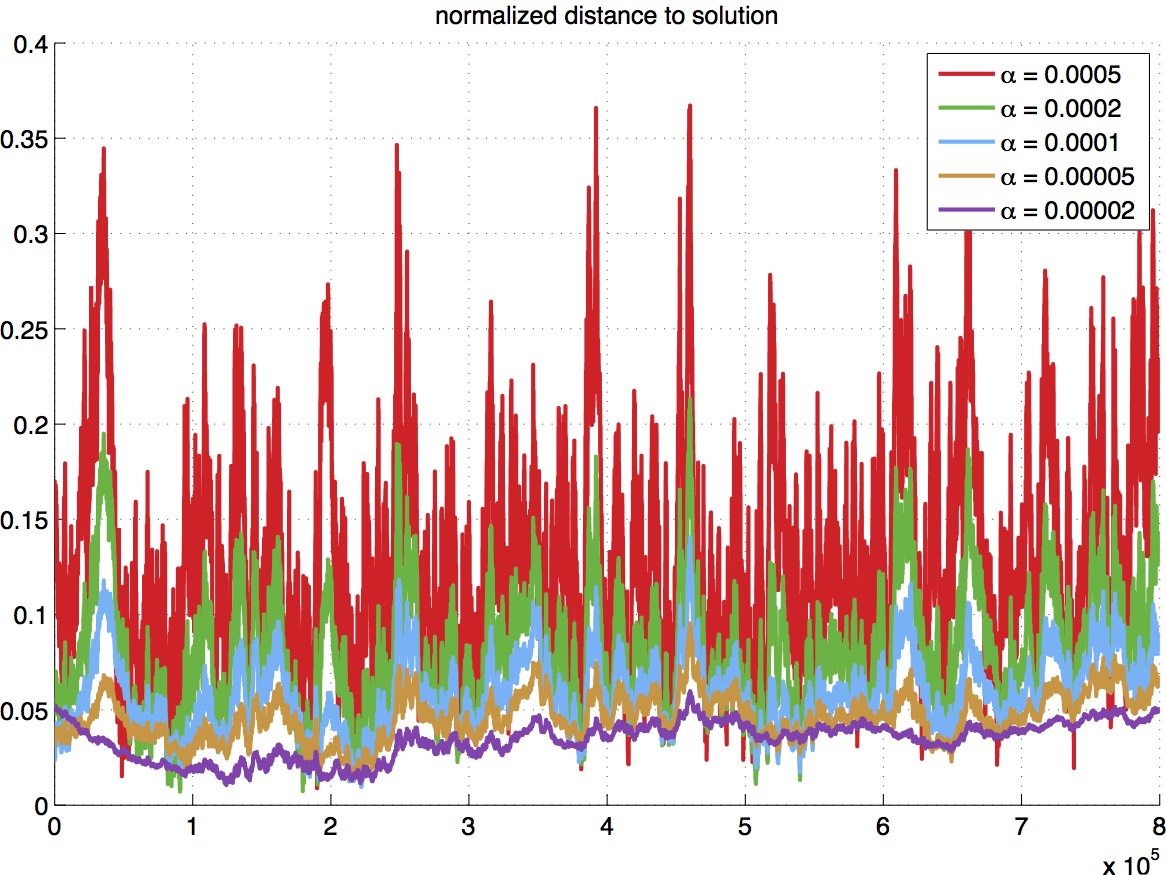} \hfill 
\includegraphics[width=0.49\linewidth]{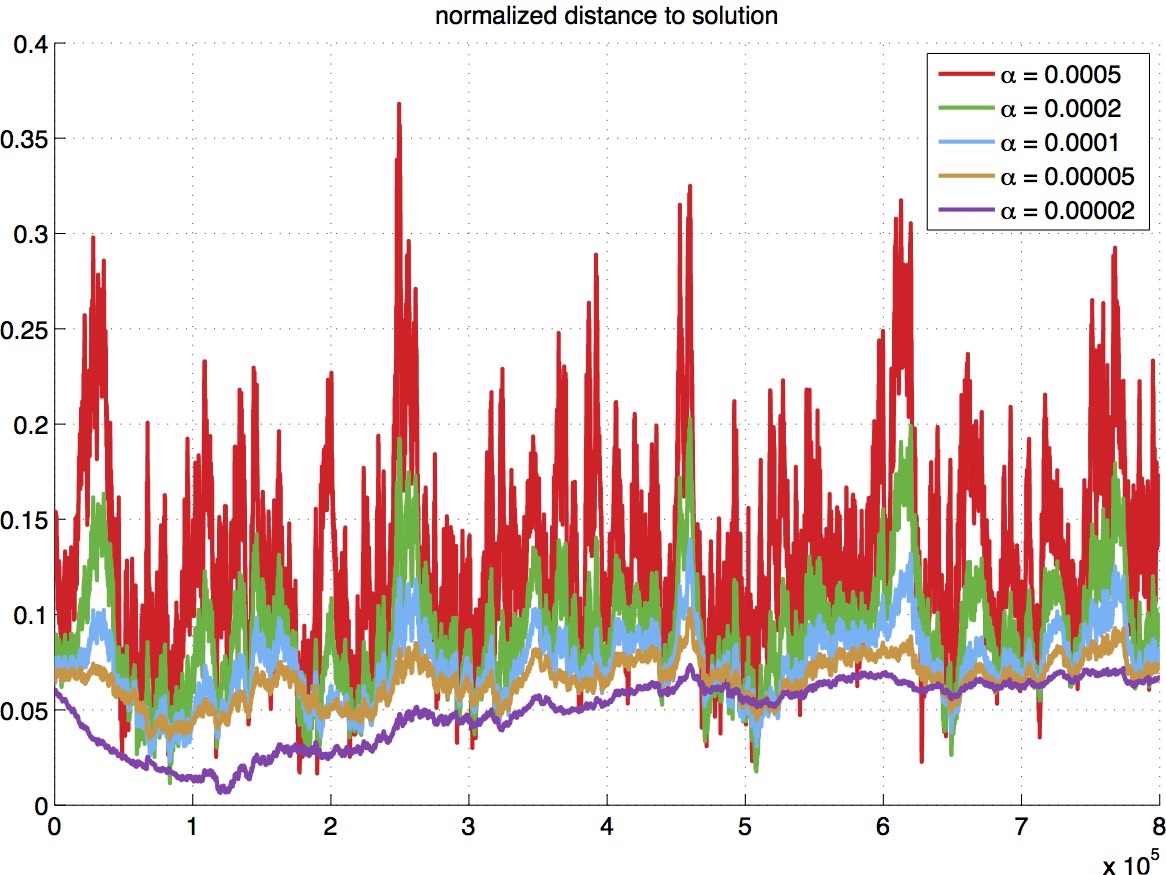} 
\caption{Variant I (left) and Variant II (right) with perturbation. Top: averaged iterates $\bar\theta^\alpha_t$; bottom: iterates $\theta_t^\alpha$. Data are from a single run.}\label{fig-cnst-ex2g}
\end{figure}

\medskip
\noindent Figures \ref{fig-cnst-ex2f}-\ref{fig-cnst-ex2g}: In these two figures we plotted the normalized distances to $\theta^*$ of a trajectory of averaged iterates $\bar{\theta}_t^{\alpha}$ and original iterates $\theta_t^\alpha$, for each algorithm and each stepsize, using the data from one of the runs of the algorithms that produced the previous four figures. 
Our observations from these results are the same as those from Figures~\ref{fig-cnst-ex1f}-\ref{fig-cnst-ex1g} in the case of Problem I:
(i) the averaged iterates $\bar{\theta}_t^{\alpha}$ perform better than $\theta_t^\alpha$ in that they vary less and can approach a smaller neighborhood of $\theta^*$; 
(ii) the unperturbed algorithms do not seem to have any disadvantages compared with the perturbed algorithms for the same stepsize.

\begin{figure}[!htb] 
   \centering
\includegraphics[width=0.49\linewidth]{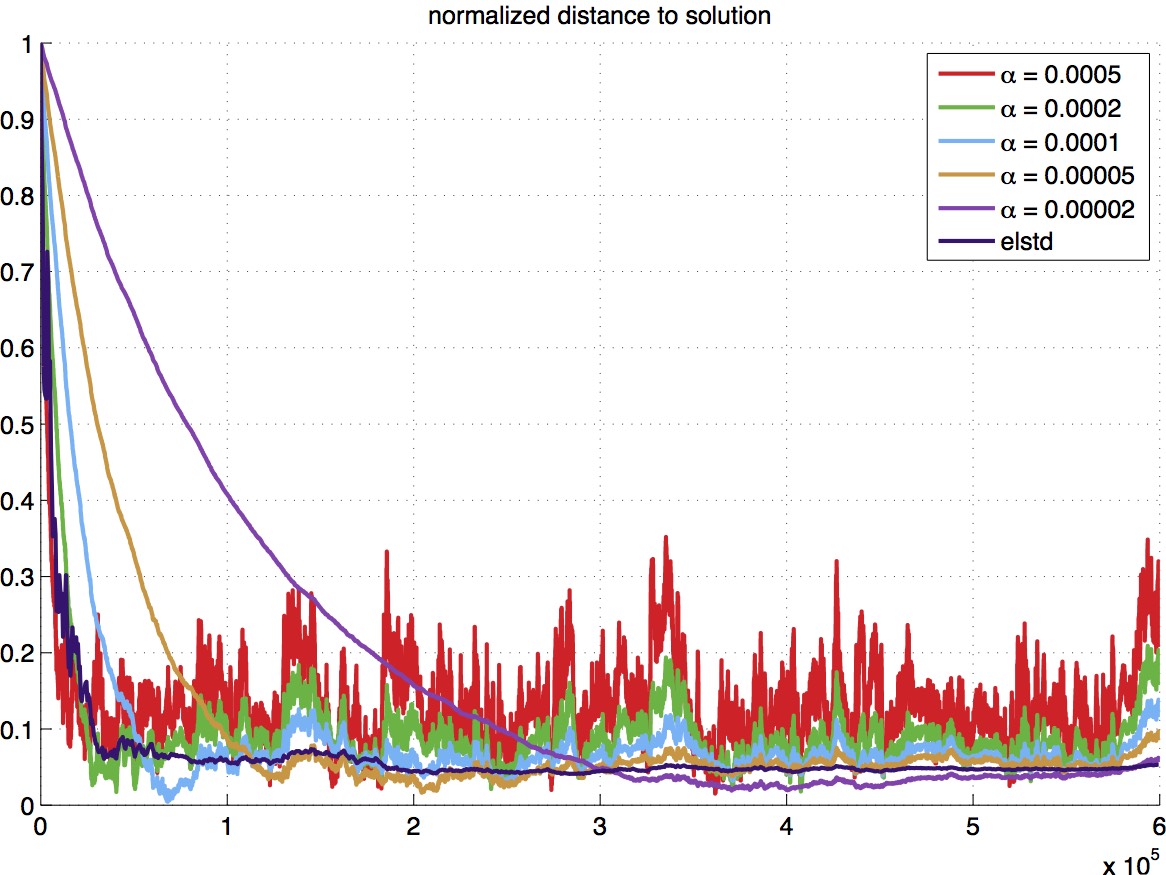} \hfill 
\includegraphics[width=0.49\linewidth]{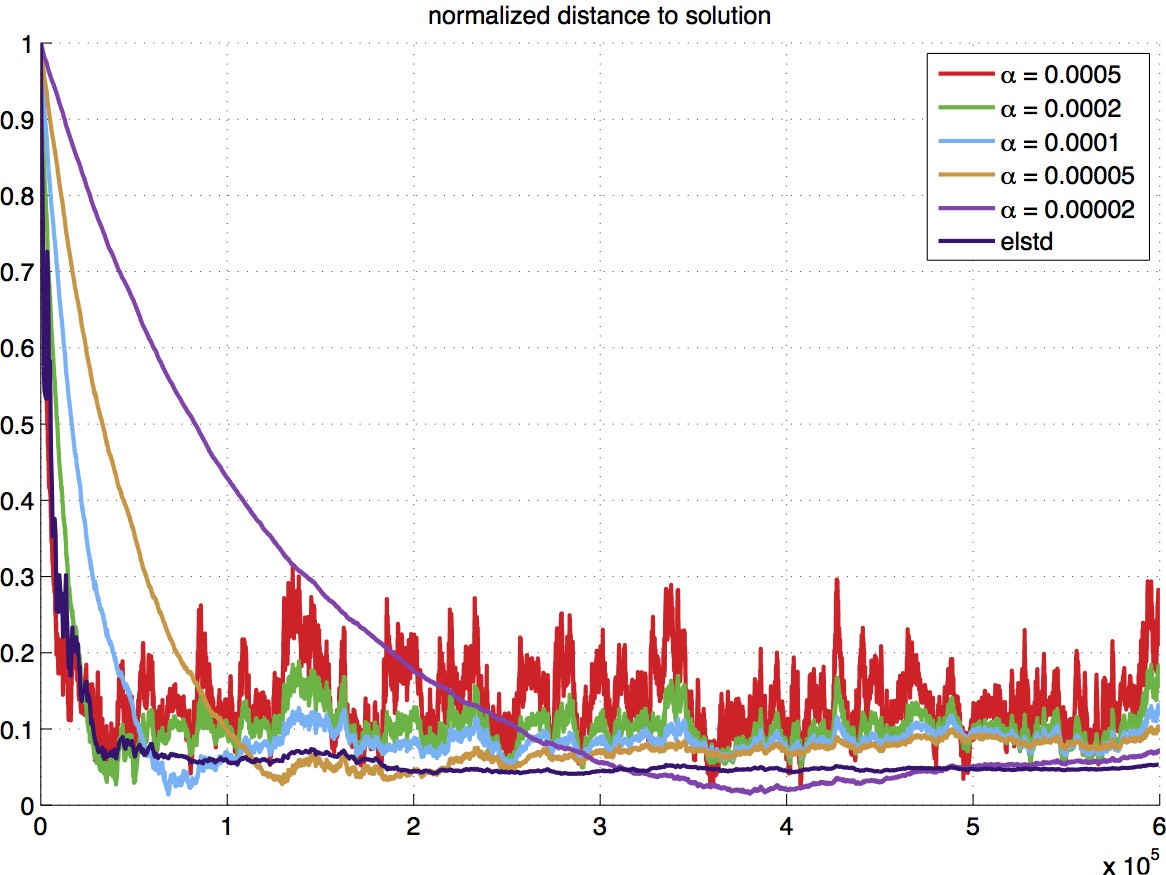} 
\caption{Variant I (left) and Variant II (right) without perturbation. Data are from a single run; ELSTD is also included for comparison.}\label{fig-cnst-ex2h}
\end{figure}

\medskip
\noindent Figure~\ref{fig-cnst-ex2h}: This experiment compares the transient behavior of the variant algorithms using a single run of $6 \times 10^5$ iterations. All the algorithms start from the same initial condition, no portion of the run is discarded, and ELSTD is also included for comparison. The linear equations formed by ELSTD are solved every 500 iterations to produce the ELSTD curve shown in the figure. It can be seen that ELSTD converges rapidly. The variant algorithms can make fast initial progress too with the largest stepsize $\alpha = 0.0005$, although because of the big stepsize, they quickly start to oscillate in a relatively large neighborhood of $\theta^*$.


\clearpage

\section{Simulation Results for the Diminishing-stepsize Case} \label{sec-dimstp}

In this section we illustrate the behavior of Variant I and Variant II with diminishing stepsize for the two test problems.
As in the previous case, we set the radius parameter $r_\H=100$ and use the componentwise truncation function $\psi_K$ with $K=50$ in the two variant algorithms.
For visualizing the behavior of the $\theta$-iterates as well as the behavior of multiple consecutive iterates, we will plot their normalized distances to the desired ETD solution $\theta^*$ as before. Given the close connection between the constant-stepsize case and the diminishing-stepsize case, and given also what we already observed in the former case, the results from the present part of the experiments, to be reported below, turn out to be as expected.

\subsection{Problem I}

\begin{figure}[!th] 
   \centering
\includegraphics[width=0.45\linewidth]{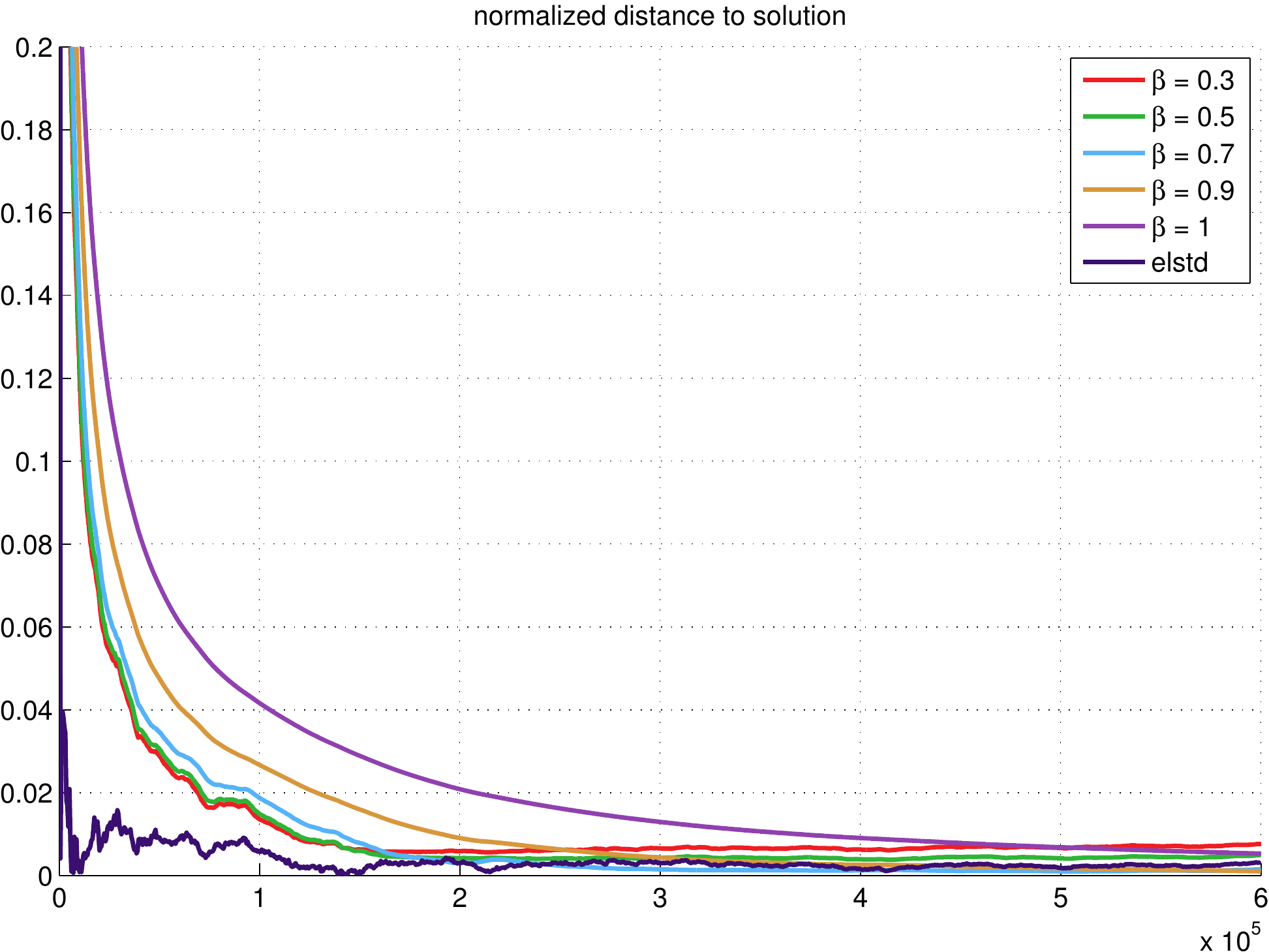} \hfill  
\includegraphics[width=0.45\linewidth]{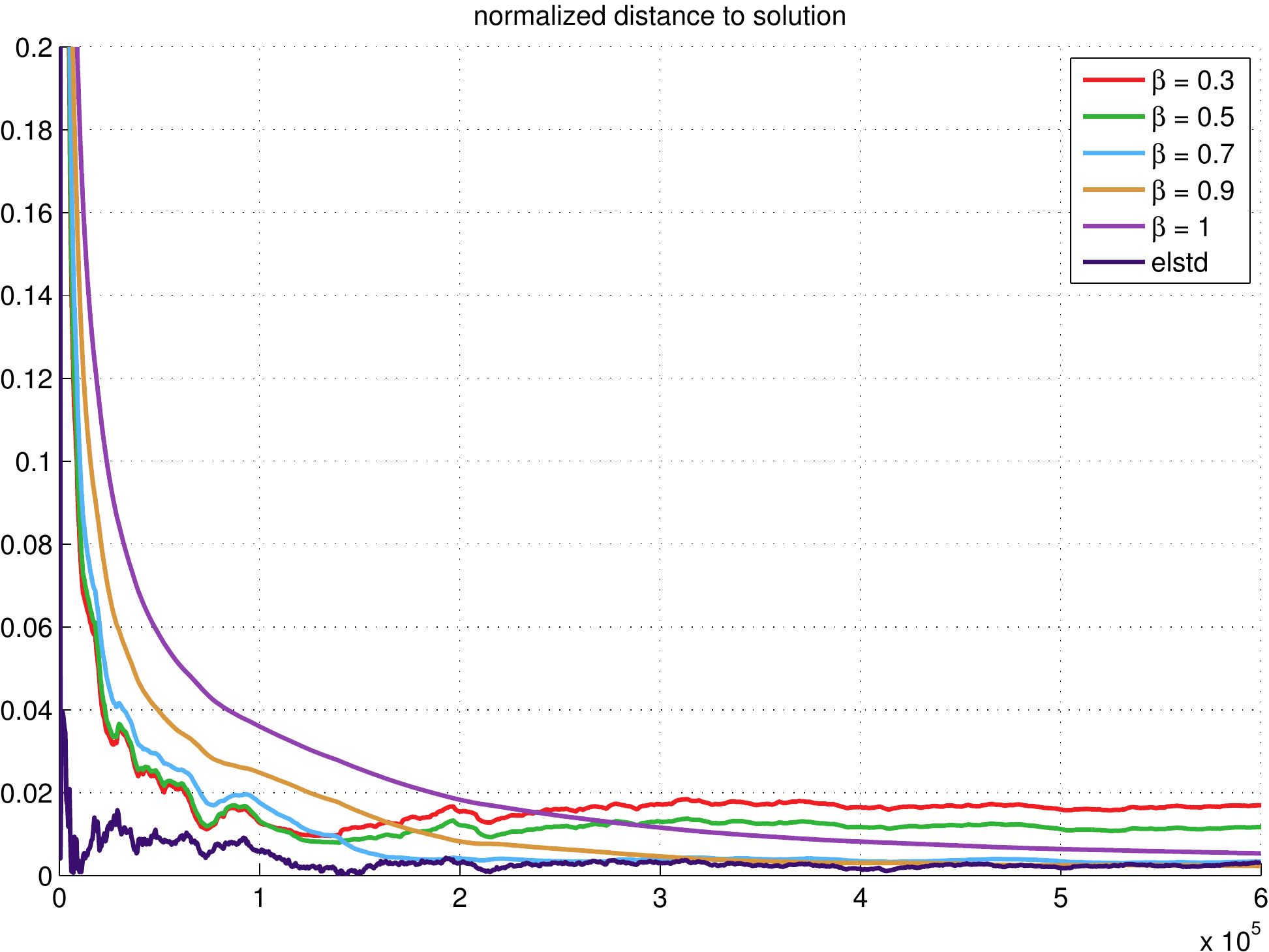}\\ 
\includegraphics[width=0.45\linewidth]{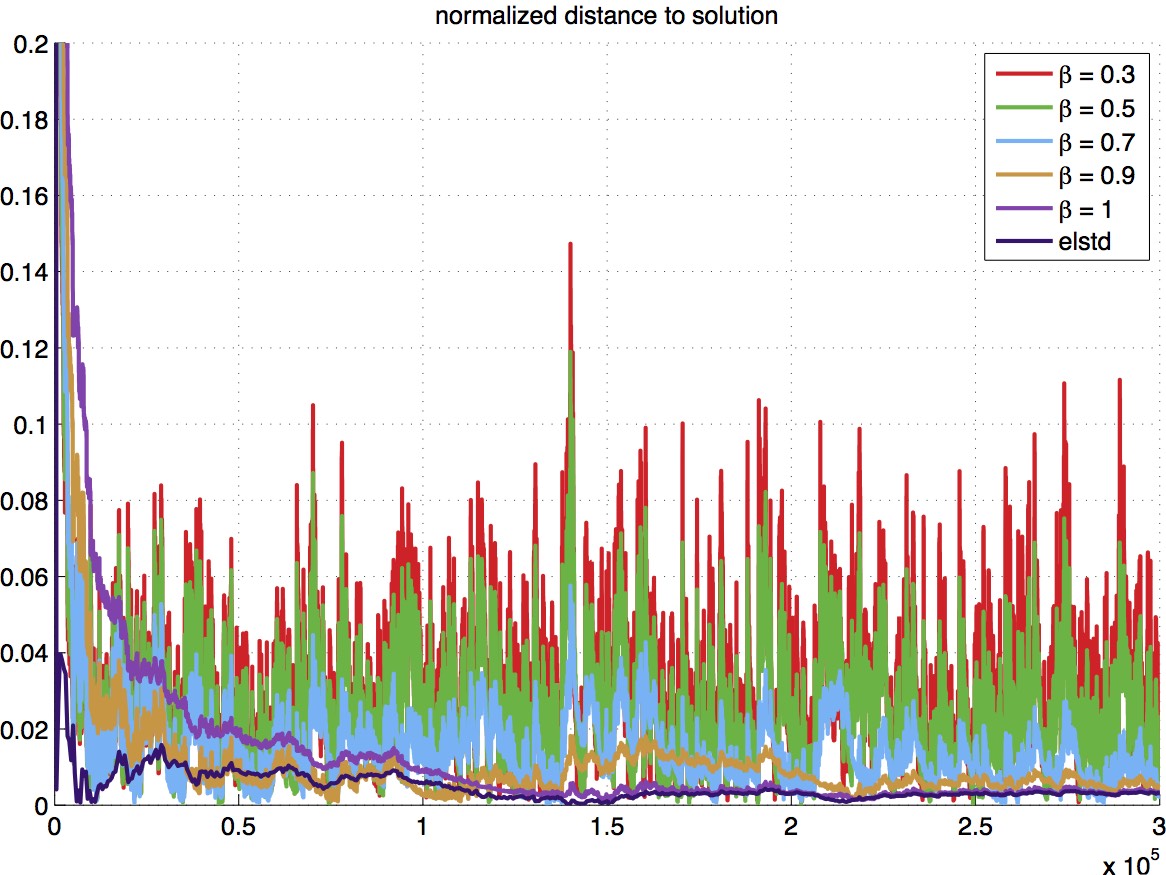} \hfill 
\includegraphics[width=0.45\linewidth]{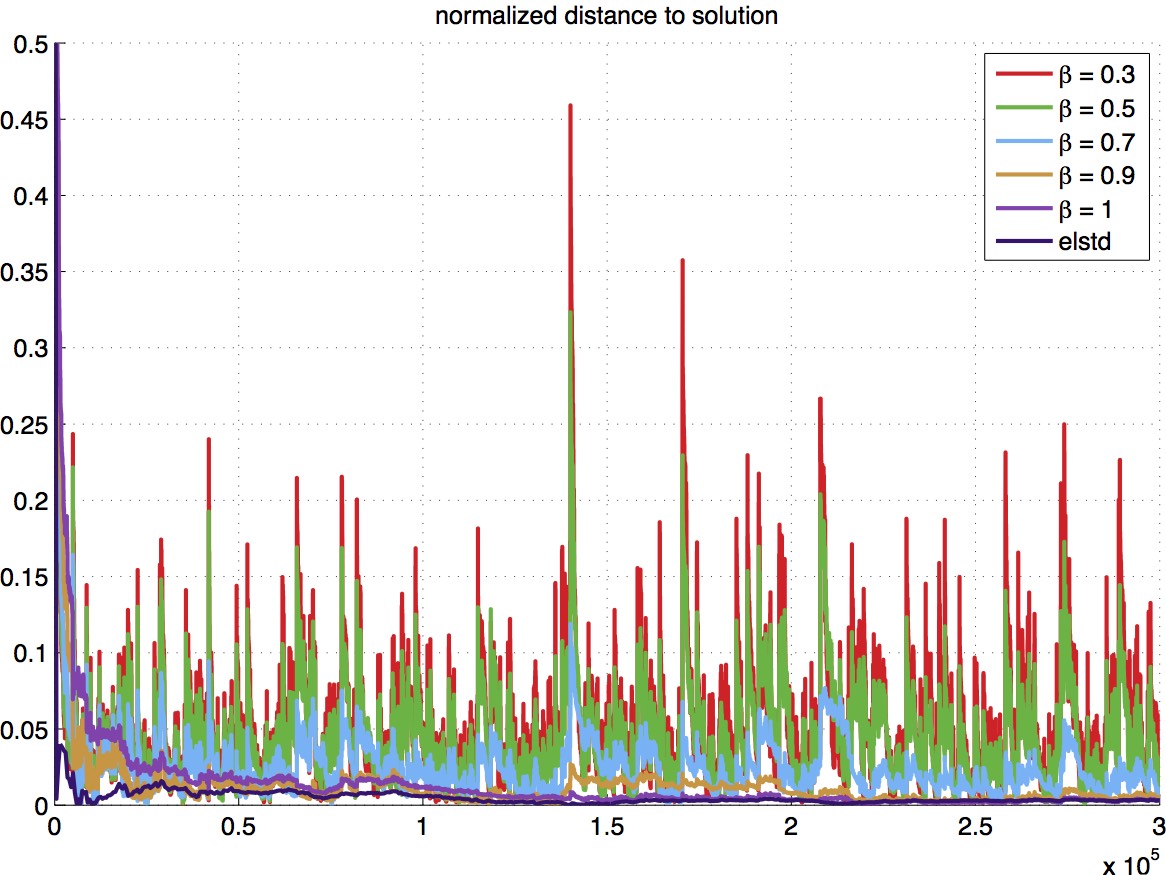} 
\caption{Variant I (left) and Variant II (right). Top: averaged iterates $\bar\theta_t$ for the entire run; bottom: iterates $\theta_t$ for the first half of the run. ELSTD is also included for comparison. See the text for details.} \label{fig-dim-ex1a}
\end{figure}

In the first experiment, we used five stepsize sequences that decrease at different rates $\beta$,
$$ \alpha_t = \frac{1}{200 + (0.1 \, t)^\beta} \qquad \text{for} \ \ \  \beta \in \{0.3, 0.5, 0.7, 0.9, 1\}.$$
We ran the two algorithms with these five stepsize rules simultaneously for $6 \times 10^5$ iterations, using a common state trajectory.  
The results are plotted in Figure~\ref{fig-dim-ex1a}.
ELSTD (modified as in Section~\ref{sec-conststp}) is also included for comparison: the linear equations formed by ELSTD are solved every $500$ iterations to produce the ELSTD curve in the figure.

The top row of Figure~\ref{fig-dim-ex1a} shows the normalized distances of the averaged iterates $\bar\theta_t$ for the entire run, and the bottom row shows the normalized distances of the iterates $\theta_t$ for only the first half of the run, in order to have a close-up view of the transient behavior. Comparing the top row with the bottom row, the advantages of the averaged iterates for large stepsizes, especially $\beta=0.3, 0.5$, can be seen. (The use of averaging for $\beta < 1$ is known as Polyak-averaging). We can also see that the iterates $\theta_t$ for $\beta = 0.3$ or $0.5$ did not settle in a small neighborhood of $\theta^*$ like the iterates generated with smaller stepsizes. This can be explained as follows: Even after $t=6 \times 10^5$ iterations, we have $\alpha_t \approx 0.004$ for $\beta=0.3$ and $\alpha_t \approx 0.002$ for $\beta=0.5$, so we can expect the iterates for $\beta = 0.3$ or $0.5$ to behave at best like the iterates with constant stepsize $0.002$ (cf.\ the bottom row of Figure~\ref{fig-cnst-ex1f}).

\begin{figure}[htb] 
   \centering
\includegraphics[width=0.42\linewidth]{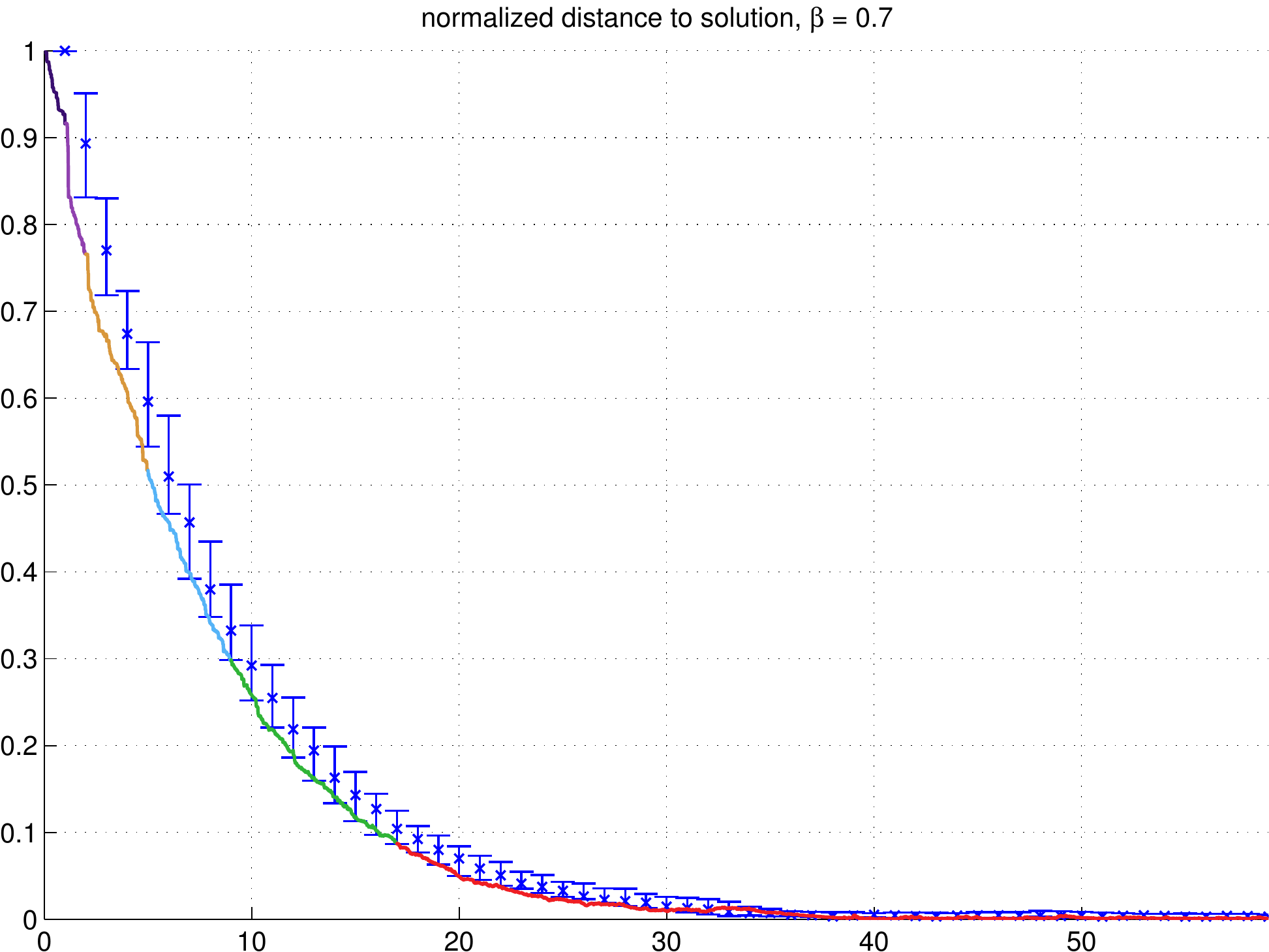} \qquad
\includegraphics[width=0.42\linewidth]{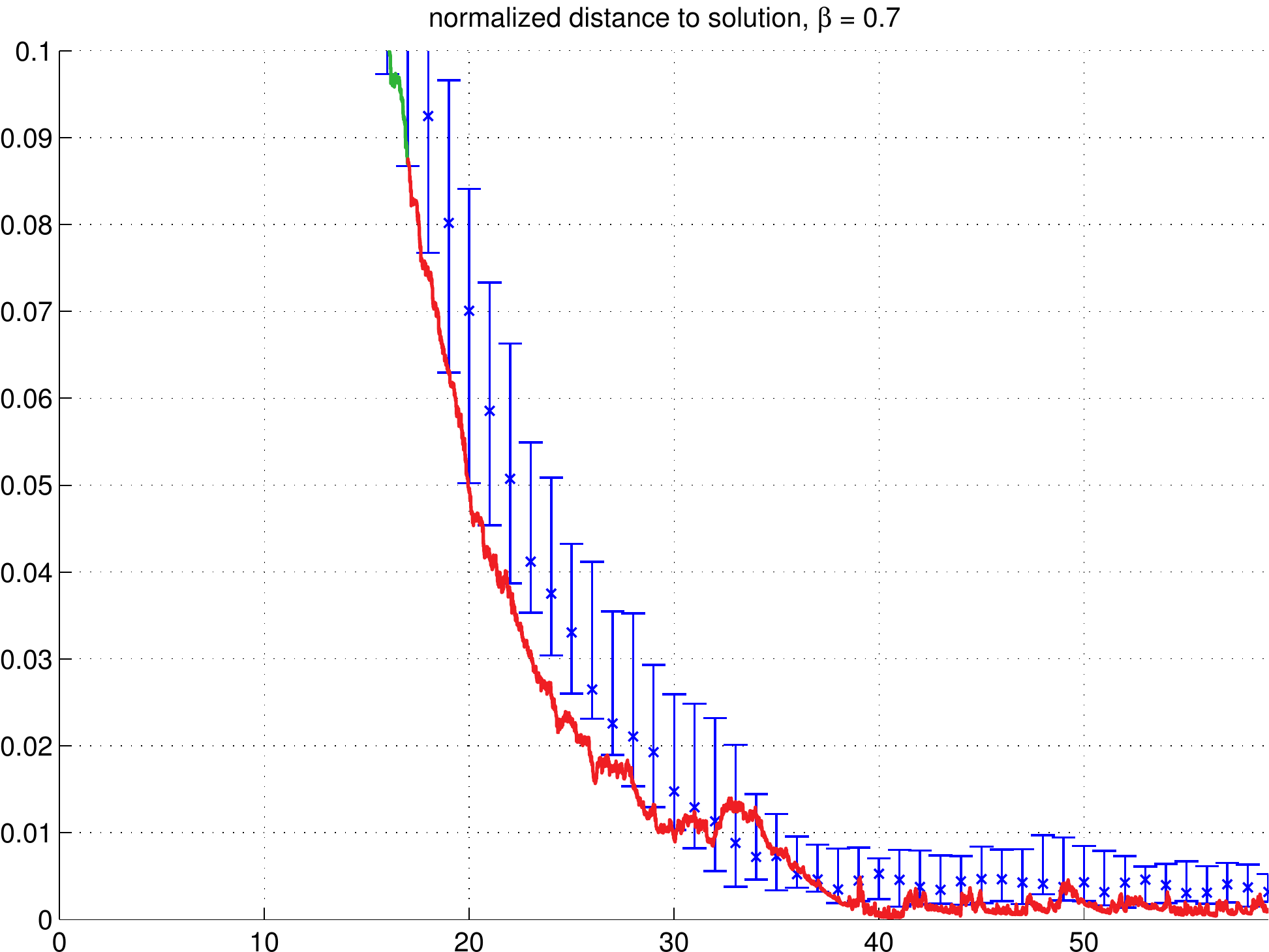}\\*[0.1cm] 
\includegraphics[width=0.42\linewidth]{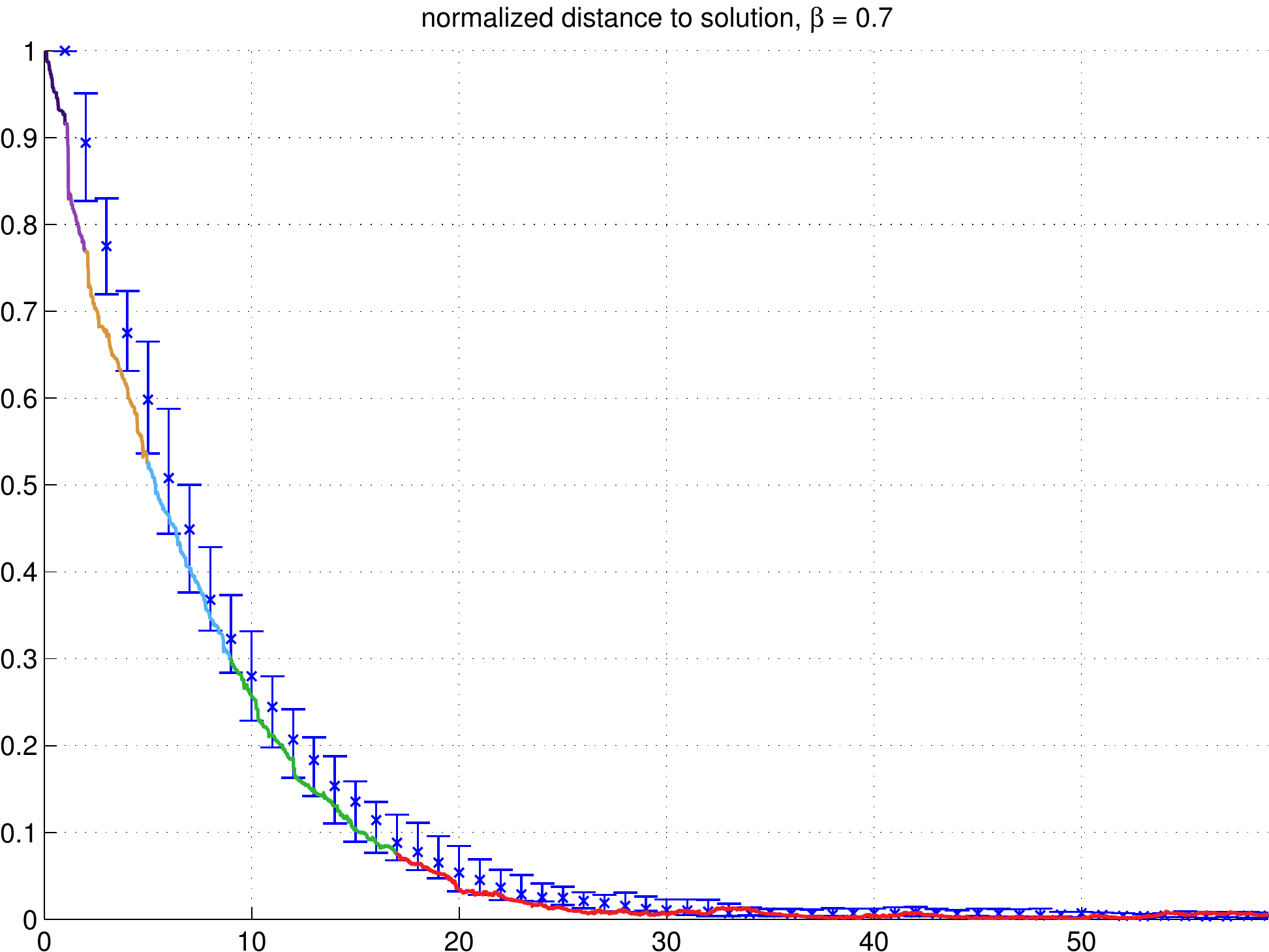} \qquad
\includegraphics[width=0.42\linewidth]{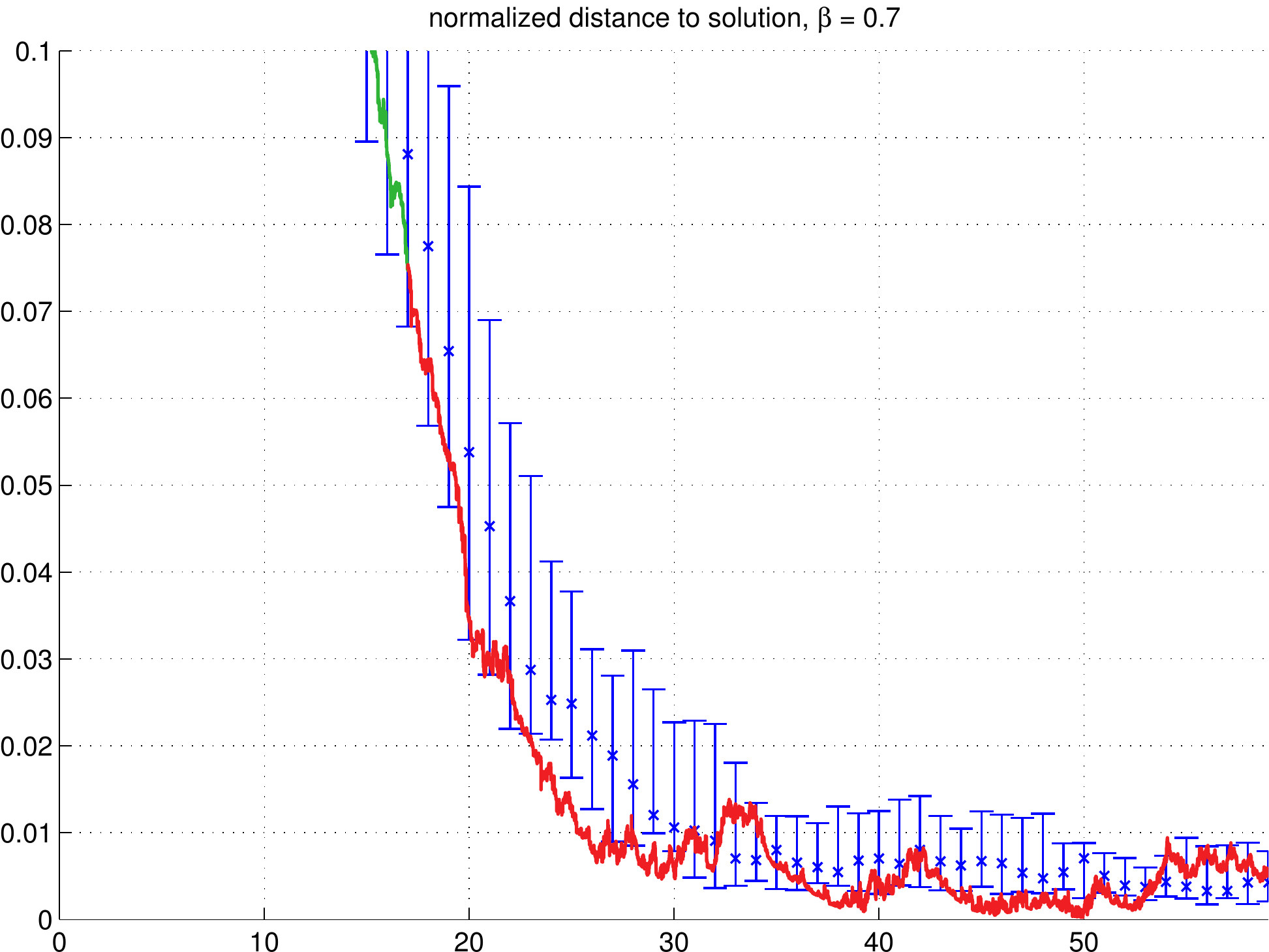} 
\caption{Variant I (top) and Variant II (bottom) with $\beta = 0.7$. The bottom portion of each plot on the left is enlarged and shown on the right. The solid curve corresponds to one run, with the $x$-component being $x=\sum_{k=0}^t \alpha_k$ and the $y$-component being the normalized distance of $\theta_t$ to $\theta^*$. (See the text for more details.)}\label{fig-dim-ex1b}
\end{figure}

\begin{figure}[!h] 
   \centering
\includegraphics[width=0.42\linewidth]{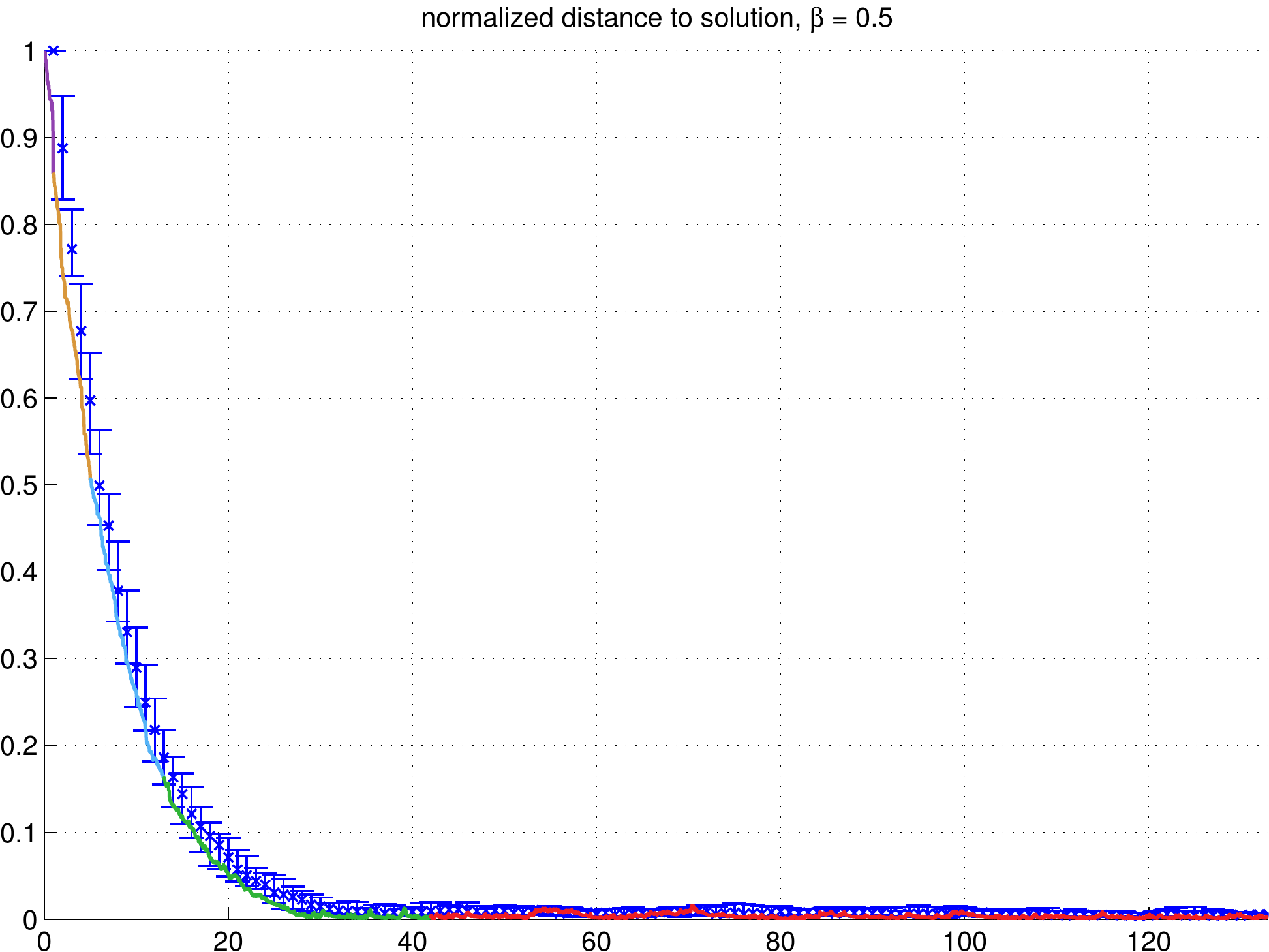} \qquad  
\includegraphics[width=0.42\linewidth]{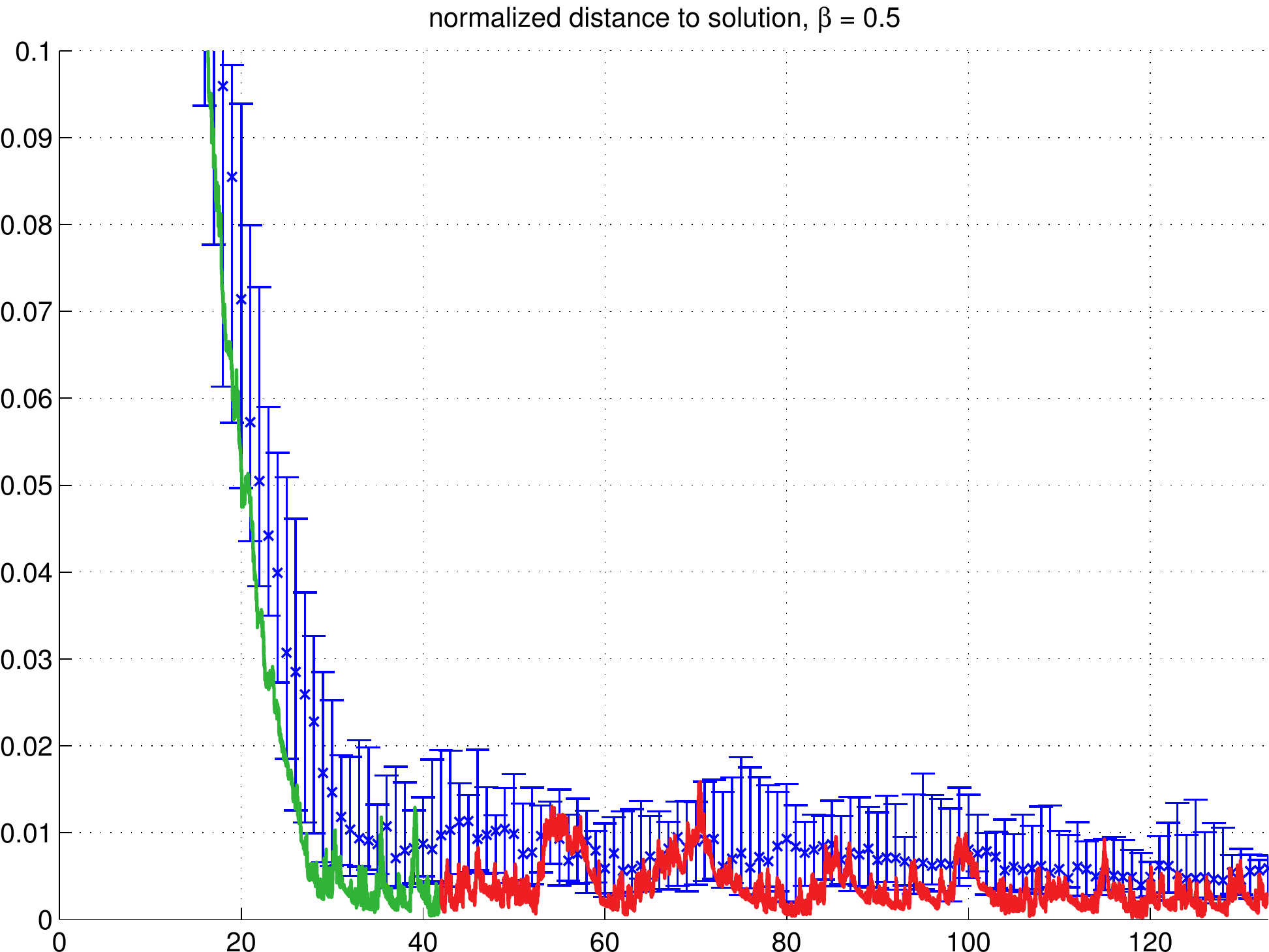}\\*[0.1cm] 
\includegraphics[width=0.42\linewidth]{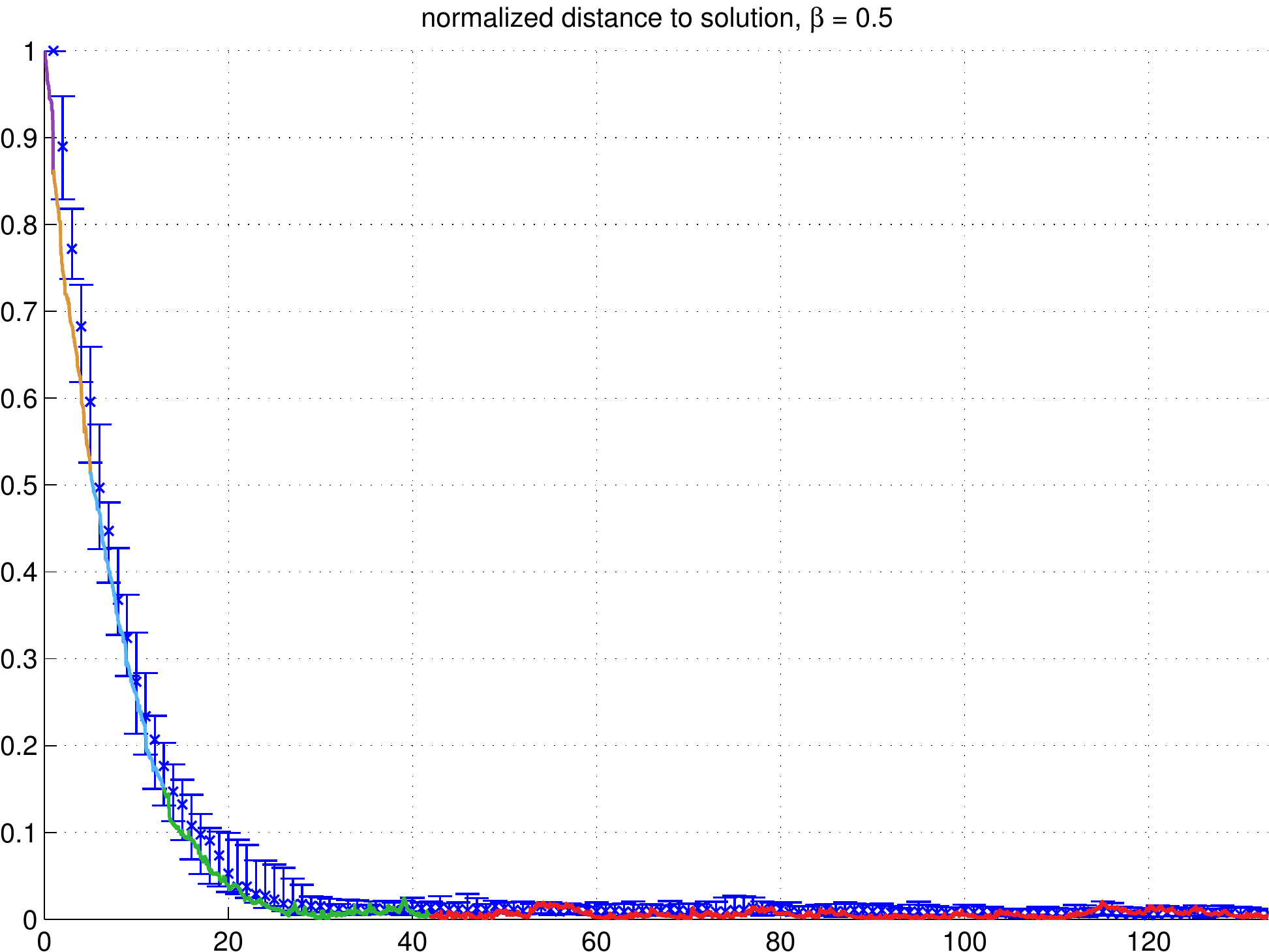} \qquad
\includegraphics[width=0.42\linewidth]{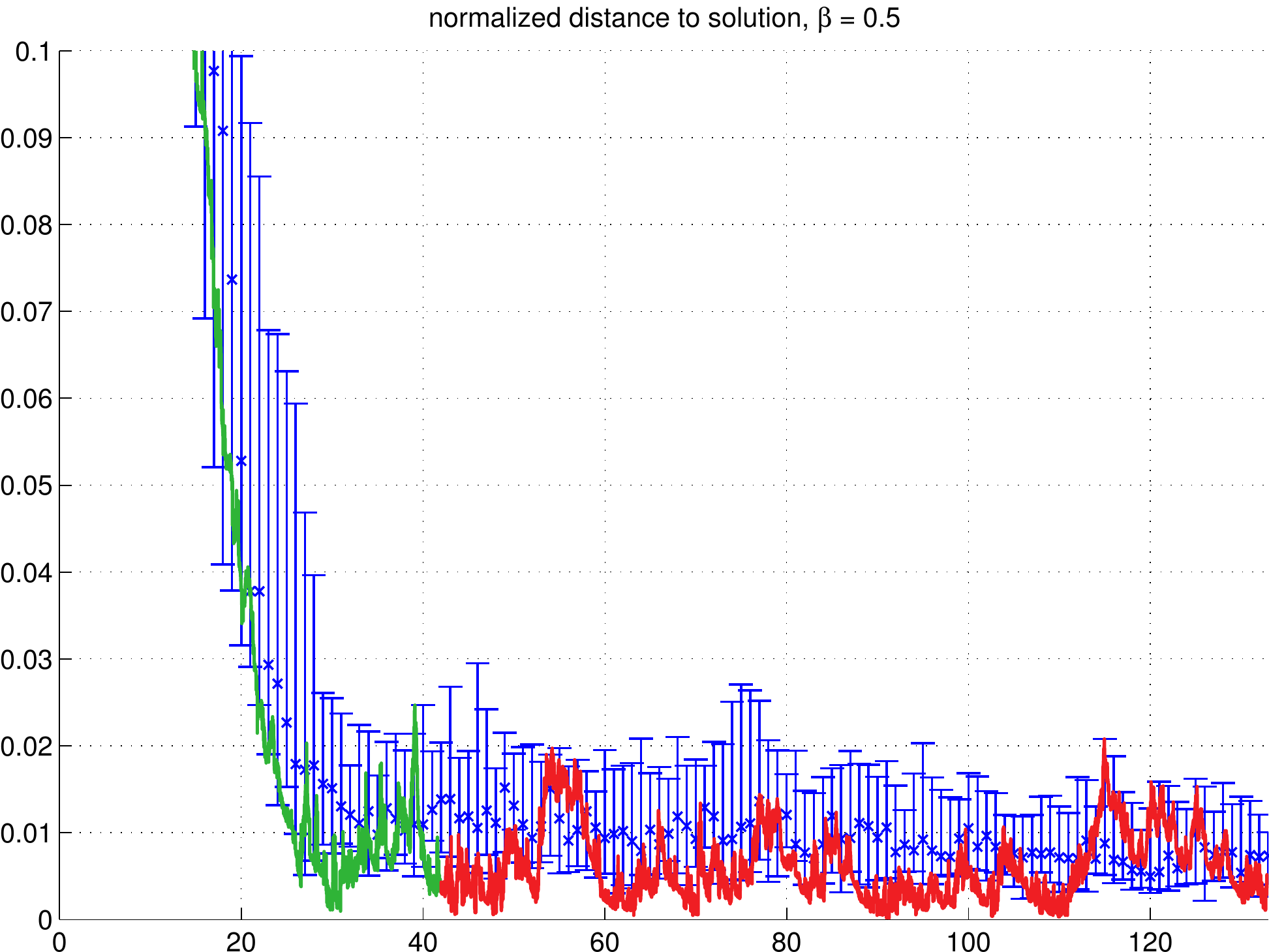} 
\caption{Variant I (top) and Variant II (bottom) with $\beta = 0.5$. The bottom portion of each plot on the left is enlarged and shown on the right. The solid curve corresponds to one run, with the $x$-component being $x=\sum_{k=0}^t \alpha_k$ and the $y$-component being the normalized distance of $\theta_t$ to $\theta^*$. (See the text for more details.)}\label{fig-dim-ex1c}
\end{figure}

The next experiment was designed in accordance with the observation of the relation between the diminishing-stepsize case and the constant-stepsize case just mentioned.  
We did $10$ independent runs of $10^6$ iterations each, for the two variant algorithms using two stepsize rules:
$$ \alpha_t = \frac{1}{200 + (5  t)^\beta} \ \ \  \text{for} \ \beta = 0.7, \qquad \text{and} \quad \alpha_t = \frac{1}{200 + (200 \, t)^\beta} \ \ \ \text{for} \ \beta = 0.5. $$
Since we want to test the convergence behavior of the algorithms, in defining the preceding stepsize rules, we have made sure that the stepsize becomes small enough later in the run (at $t=10^6$, $\alpha_t$ is of the order $10^{-5}$ in both cases of $\beta$).
The simulation results are plotted in Figure~\ref{fig-dim-ex1b} and Figure~\ref{fig-dim-ex1c} for $\beta=0.7$ and $\beta=0.5$ respectively. The details are as follows.

In Figures~\ref{fig-dim-ex1b}-\ref{fig-dim-ex1c}, plotted in solid lines are the normalized distances (to $\theta^*$) of the iterates from one of the $10$ runs.
Specifically, each solid curve is made up of points $\big(\sum_{k=0}^t \alpha_k, |\theta_t - \theta^*|/|\theta^*| \big)$, $t \geq 0$, from a single run of an algorithm.
In words, the $x$-axis represents a continuous timeline (cf.\ \cite[Section 3.1]{etd-wkconv}), and the $x$-component of a solid curve corresponds to the sum of stepsizes up to an iteration, whereas the $y$-component of the curve corresponds to the normalized distance of that iterate. 
The whole curve is plotted on the left side of each figure, with a close-up view of its bottom portion shown on the right side.

To give a rough indication of the values of the decreasing stepsizes themselves, 
we colored segments of the solid curves in different colors according to the range of stepsizes in each segment as follows:
$\alpha_t \geq 0.003$ (black), $\alpha_t \in (0.003, 0.002]$ (purple), $\alpha_t \in (0.002, 0.001]$ (brown), 
$\alpha_t \in (0.001, 0.0005]$ (blue), $\alpha_t \in (0.0005, 0.0002]$ (green),
$\alpha_t < 0.0002$ (red).

The blue error bars in Figures~\ref{fig-dim-ex1b}-\ref{fig-dim-ex1c} give statistics about the maximal deviation from $\theta^*$ for multiple consecutive iterates from the $10$ experimental runs. 
The horizontal positions of these error bars equal positive integers $x$, and for each $x$, the $x$-th error bar is generated as follows.
For each run, the iterates $\theta_t$ are grouped into segments such that the $x$-th segment consists of those $\theta_t$ with $\sum_{k=0}^t \alpha_k \in [x-1, x)$. 
So as $x$ increases, the $x$-th segment consists of more and more iterates, but measured with respect to the continuous timeline, all the segments are of length $1$ approximately. We then calculate for each $x \geq 1$ the maximal normalized distance for the $\theta$-iterates in the $x$-th segment of each run, $\max_{\text{$x$-th segment}} |\theta_t - \theta^*|/|\theta^*|$. This gives us $10$ numbers, one for each run. We take the median, min and max of these numbers to form the $x$-th error bar. 
In particular, the point with an `$\times$' mark inside the bar is the median, and the lower and upper ends of the bar correspond to the minimum and maximum of the $10$ numbers, respectively.

The simulation results shown in Figures~\ref{fig-dim-ex1b}-\ref{fig-dim-ex1c} can be compared with the assertion in Theorem 3.3 of \cite{etd-wkconv} for Variants I and II with diminishing stepsizes.

\subsection{Problem II}

\begin{figure}[!ht] 
   \centering
\includegraphics[width=0.45\linewidth]{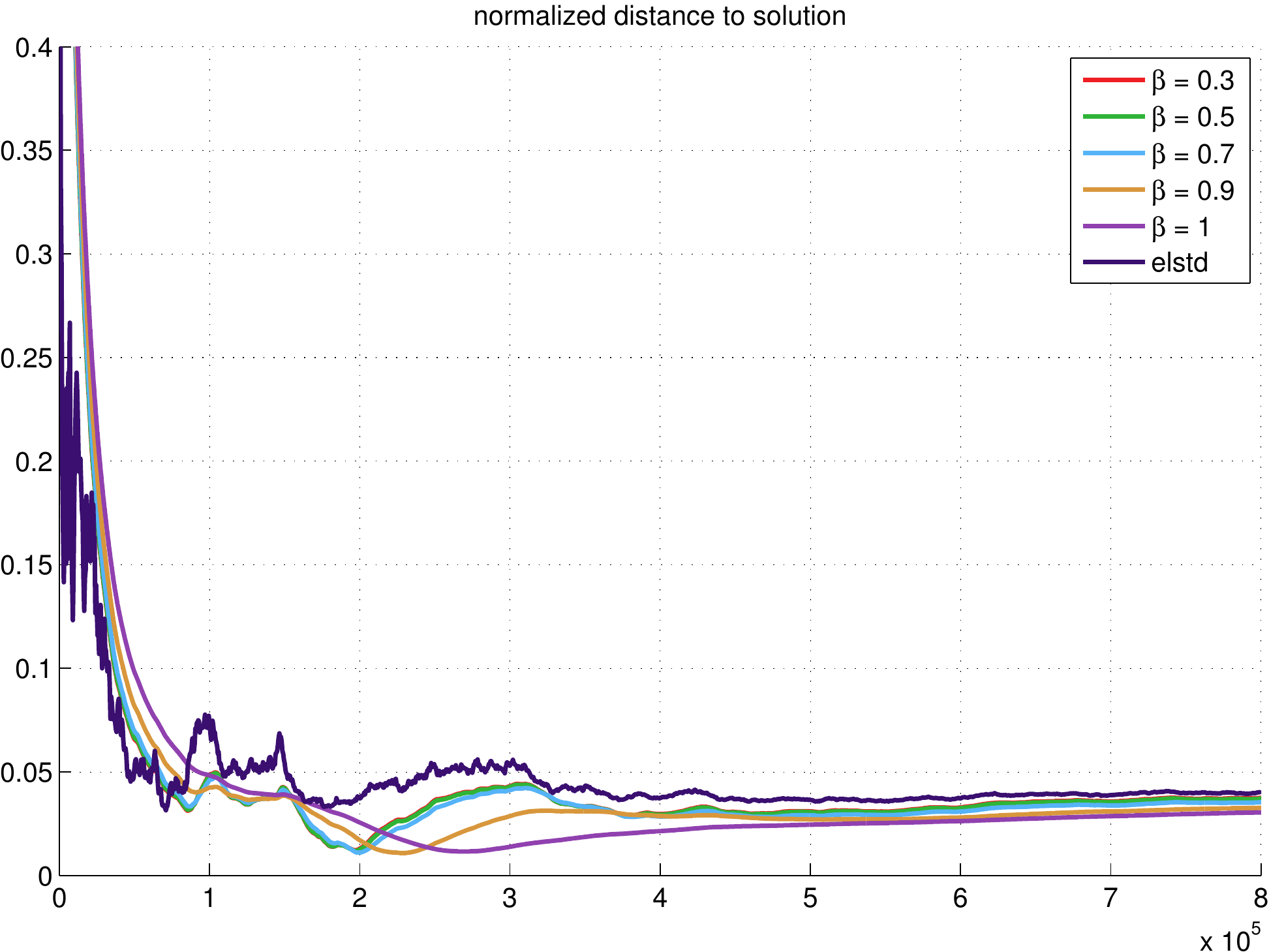} \hfill  
\includegraphics[width=0.45\linewidth]{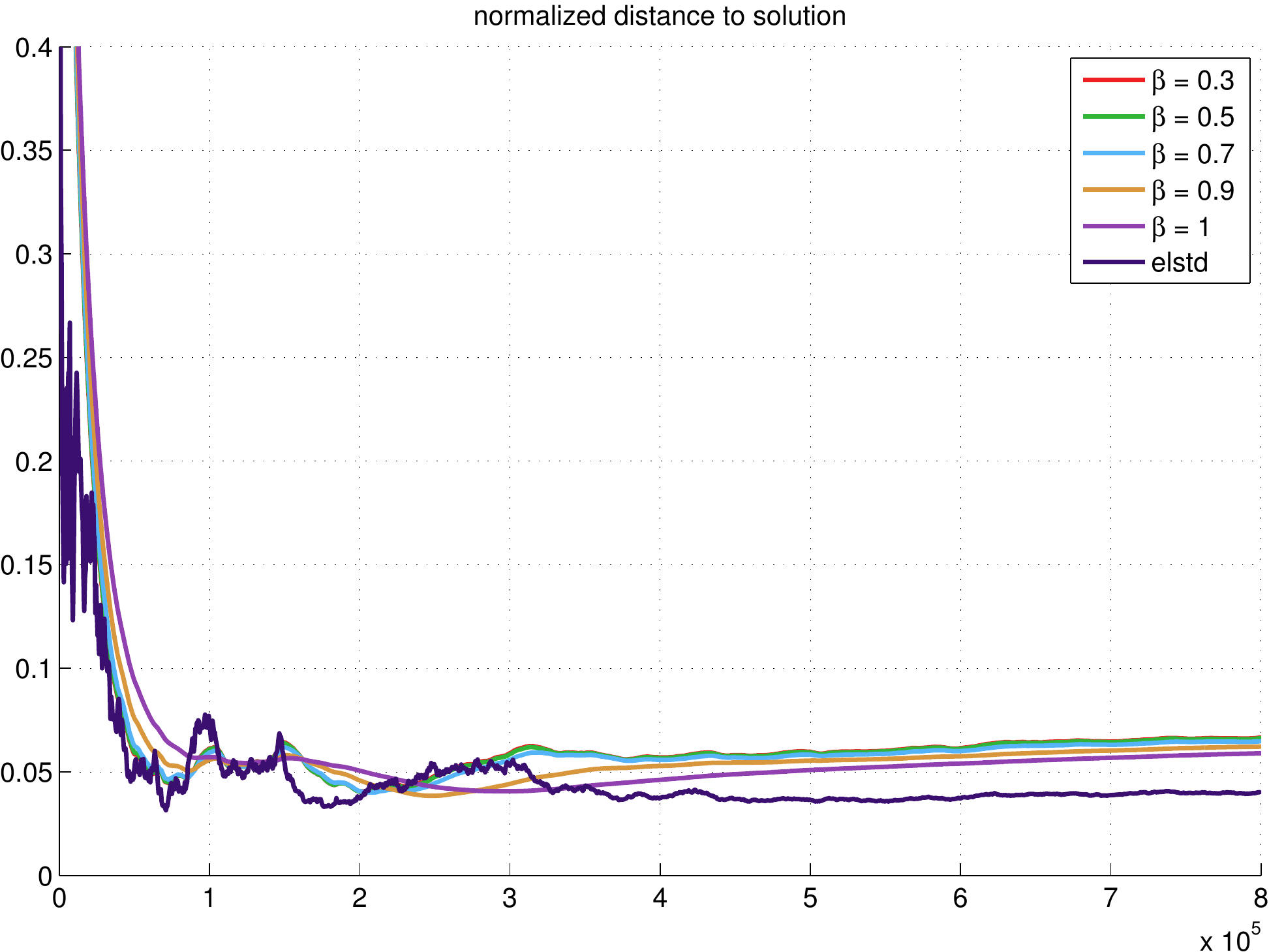}\\ 
\includegraphics[width=0.45\linewidth]{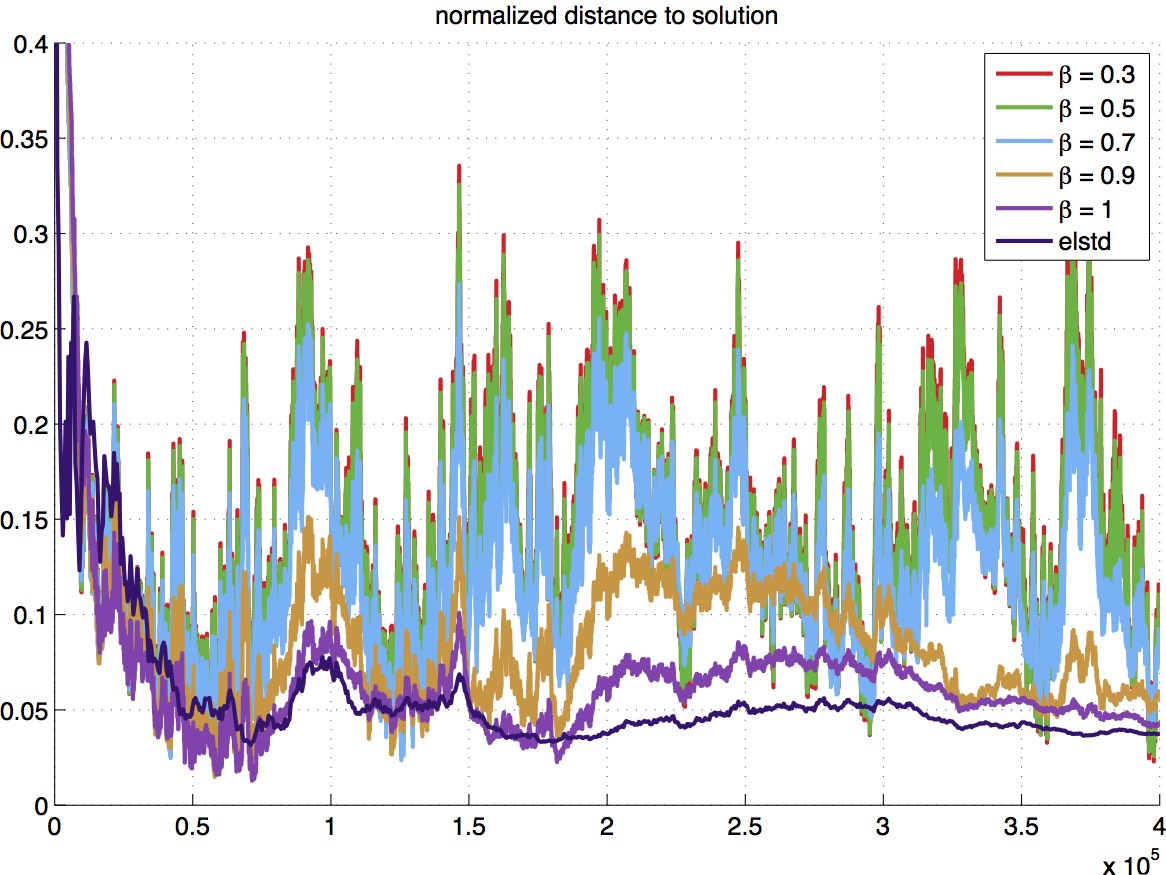} \hfill 
\includegraphics[width=0.45\linewidth]{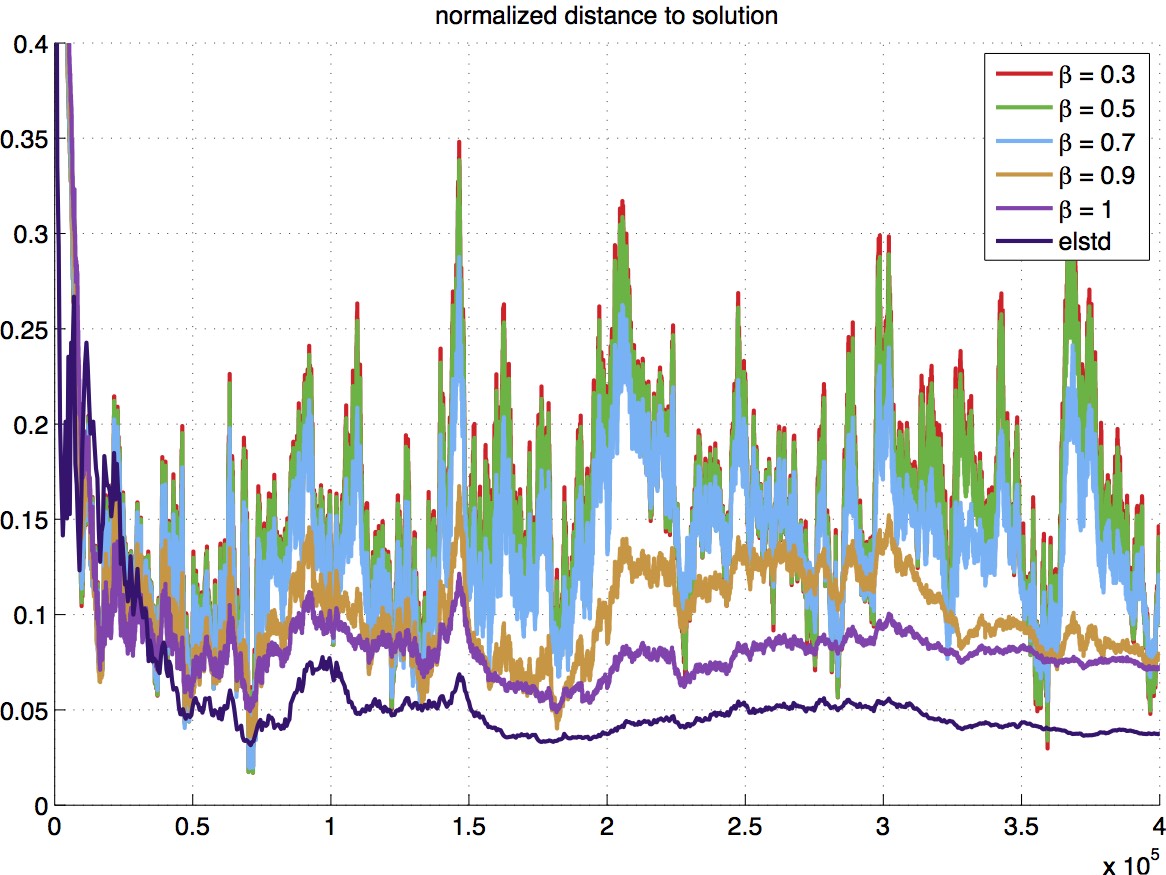} 
\caption{Variant I (left) and Variant II (right). Top: averaged iterates $\bar\theta_t$ for the entire run; bottom: iterates $\theta_t$ for the first half of the run. ELSTD is also included for comparison.} \label{fig-dim-ex2a}
\end{figure}

Similar to the previous subsection, in the first experiment for Problem II, we used five stepsize sequences that decrease at different rates $\beta$:
$$ \alpha_t = \frac{1}{2000 + (0.1 \, t)^\beta} \qquad \text{for} \ \ \ \beta \in \{0.3, 0.5, 0.7, 0.9, 1\}.$$
We ran the two algorithms with these five stepsize rules simultaneously for $8 \times 10^5$ iterations, using a common state trajectory.  
The results are plotted in Figure~\ref{fig-dim-ex2a}.
ELSTD (modified as in Section~\ref{sec-conststp}) is also included for comparison: the linear equations formed by ELSTD are solved every $500$ iterations to produce the ELSTD curve in the figure. The top row of Figure~\ref{fig-dim-ex1a} shows the normalized distances of the averaged iterates $\bar \theta_t$ for the entire run, and the bottom row shows the normalized distances of the iterates $\theta_t$ only for the first half of the run, in order to have a close-up view of the transient behavior. Once more, for $\beta < 1$, the advantages of averaging can be seen.

It can be seen that for the three largest stepsize rules ($\beta \in\{ 0.3, 0.5, 0.7\}$), the iterates behaved similarly to each other and did not settle in a small neighborhood of $\theta^*$ like the iterates generated with smaller stepsizes. This can again be understood by relating the situation here to the constant-stepsize case: With $\beta \in\{ 0.3, 0.5, 0.7\}$, the stepsizes decrease rather slowly. Even after $t=8 \times 10^5$ iterations, $\alpha_t$ is between $0.0004$ and $0.0005$ for $\beta=0.3, 0.5$ and about $0.0002$ for $\beta=0.7$. 
So we can expect the iterates to behave at best like the iterates with constant stepsize $0.0002$ (cf.\ the bottom row of Figure~\ref{fig-cnst-ex2f}).

\begin{figure}[htb] 
   \centering
\includegraphics[width=0.42\linewidth]{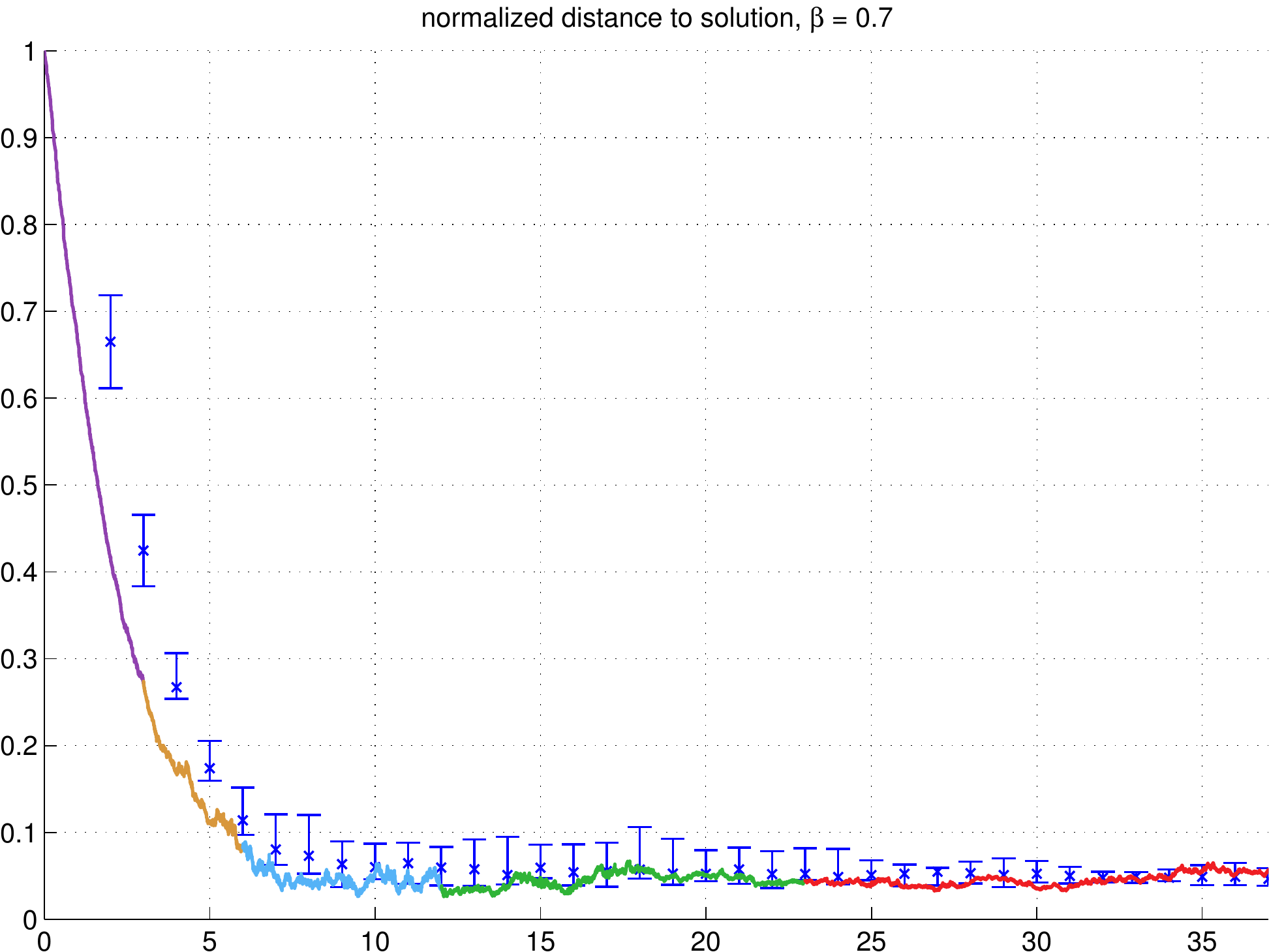} \qquad
\includegraphics[width=0.42\linewidth]{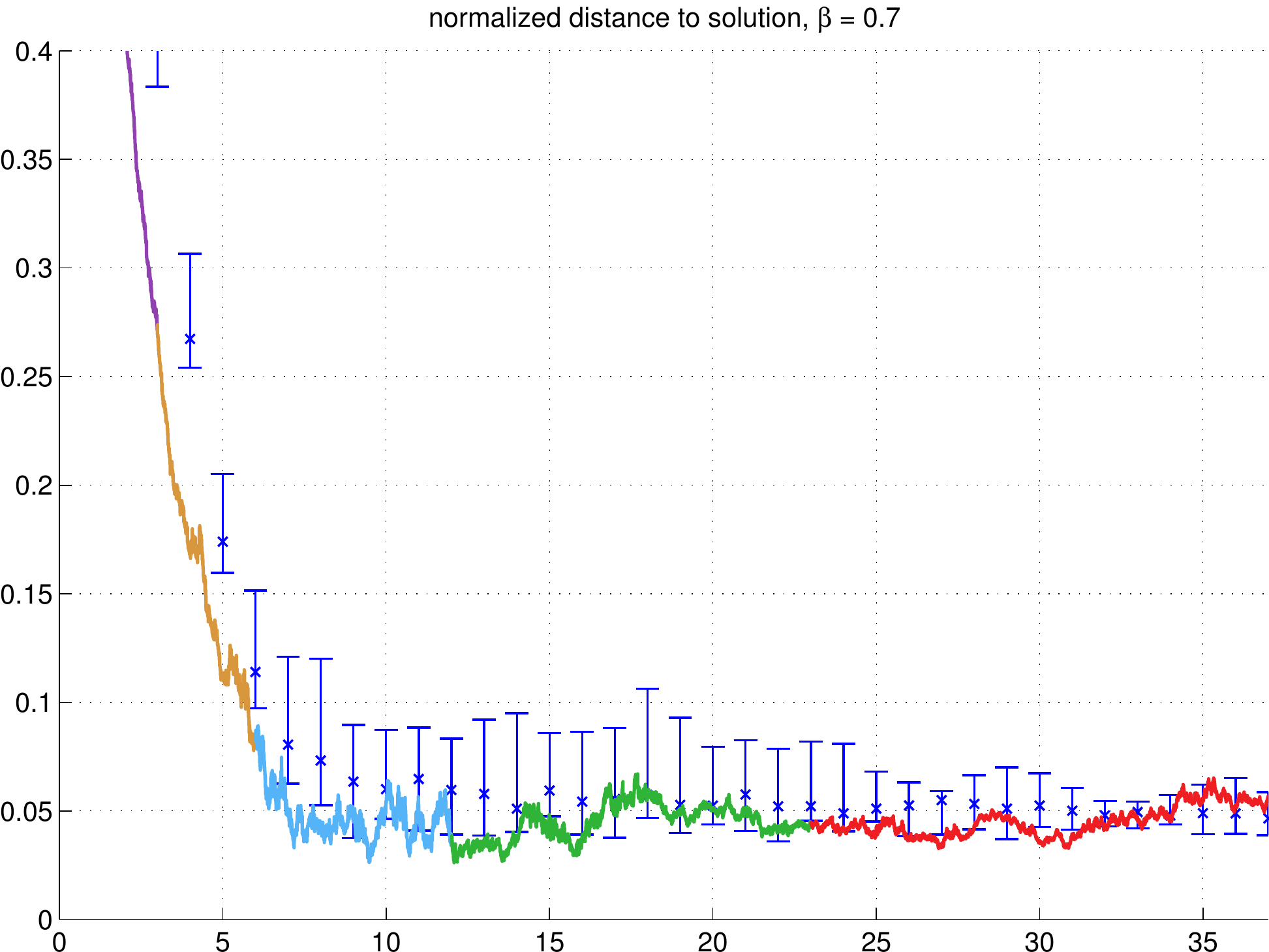}\\*[0.1cm] 
\includegraphics[width=0.42\linewidth]{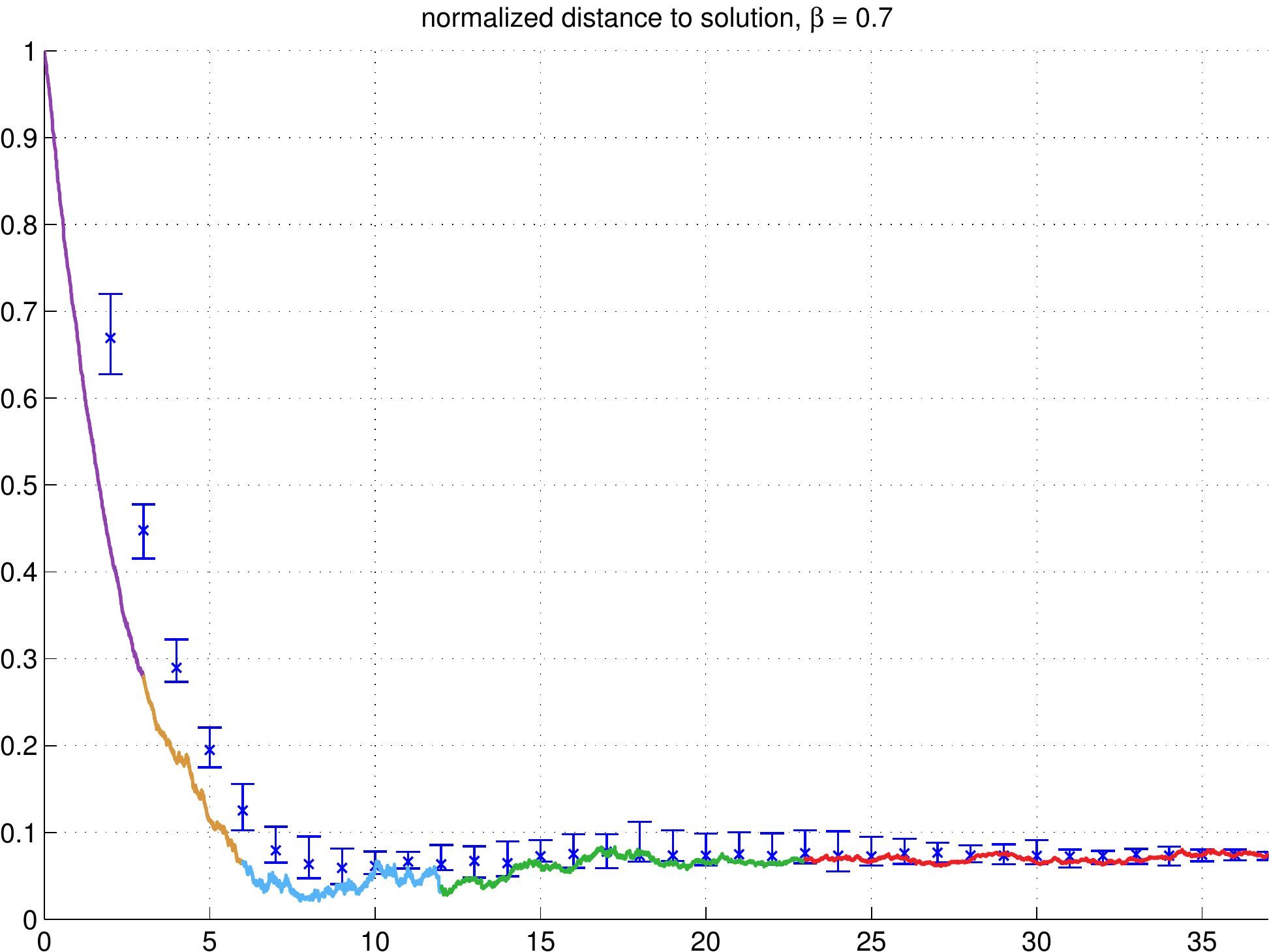} \qquad
\includegraphics[width=0.42\linewidth]{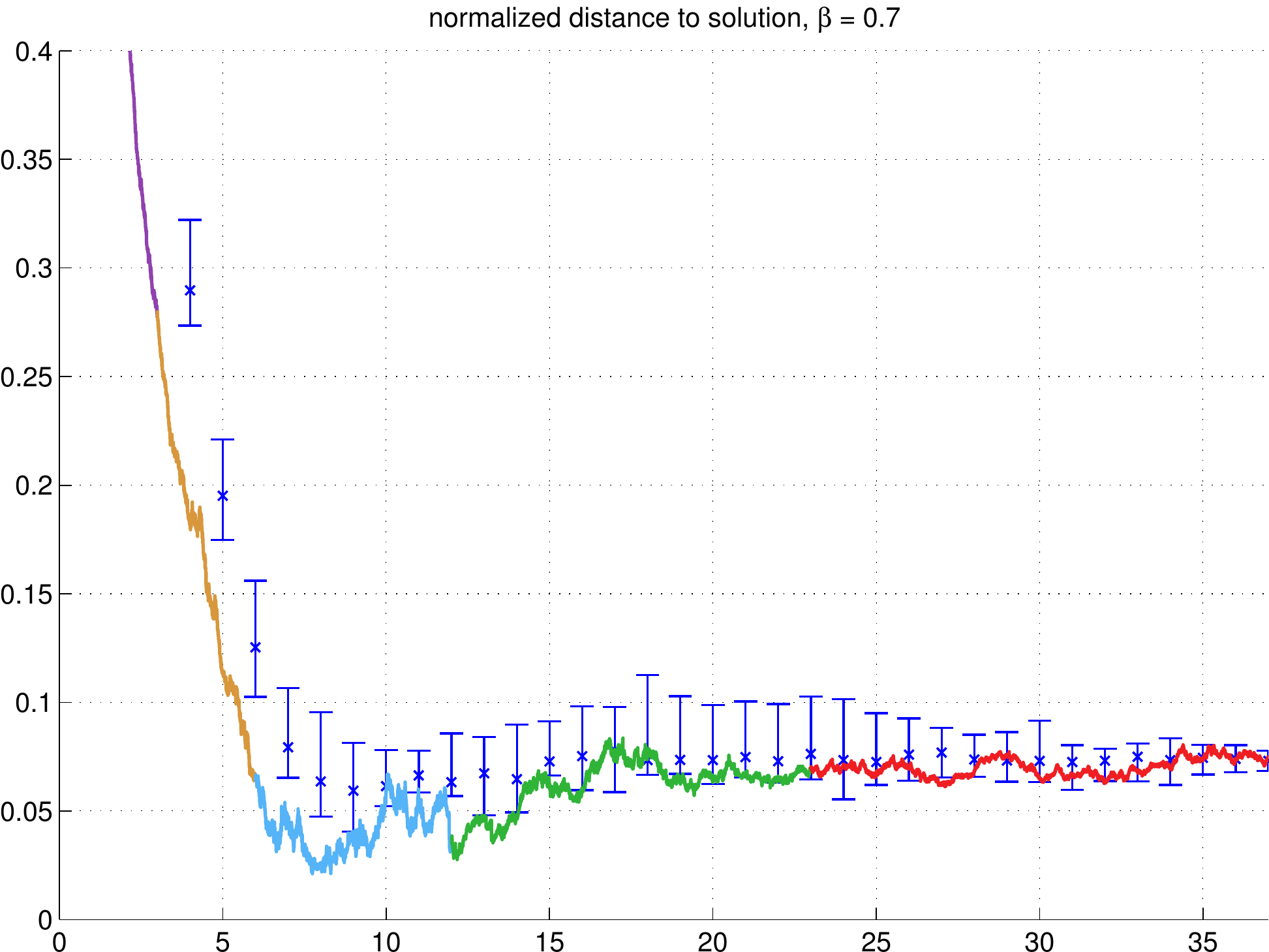} 
\caption{Variant I (top) and Variant II (bottom) with $\beta = 0.7$. The bottom portion of each plot on the left is enlarged and shown on the right. The solid curve corresponds to one run, with the $x$-component being $x=\sum_{k=0}^t \alpha_k$ and the $y$-component being the normalized distance of $\theta_t$ to $\theta^*$. (See the text for more details.)}\label{fig-dim-ex2b}
\end{figure}

\begin{figure}[!t] 
   \centering
\includegraphics[width=0.42\linewidth]{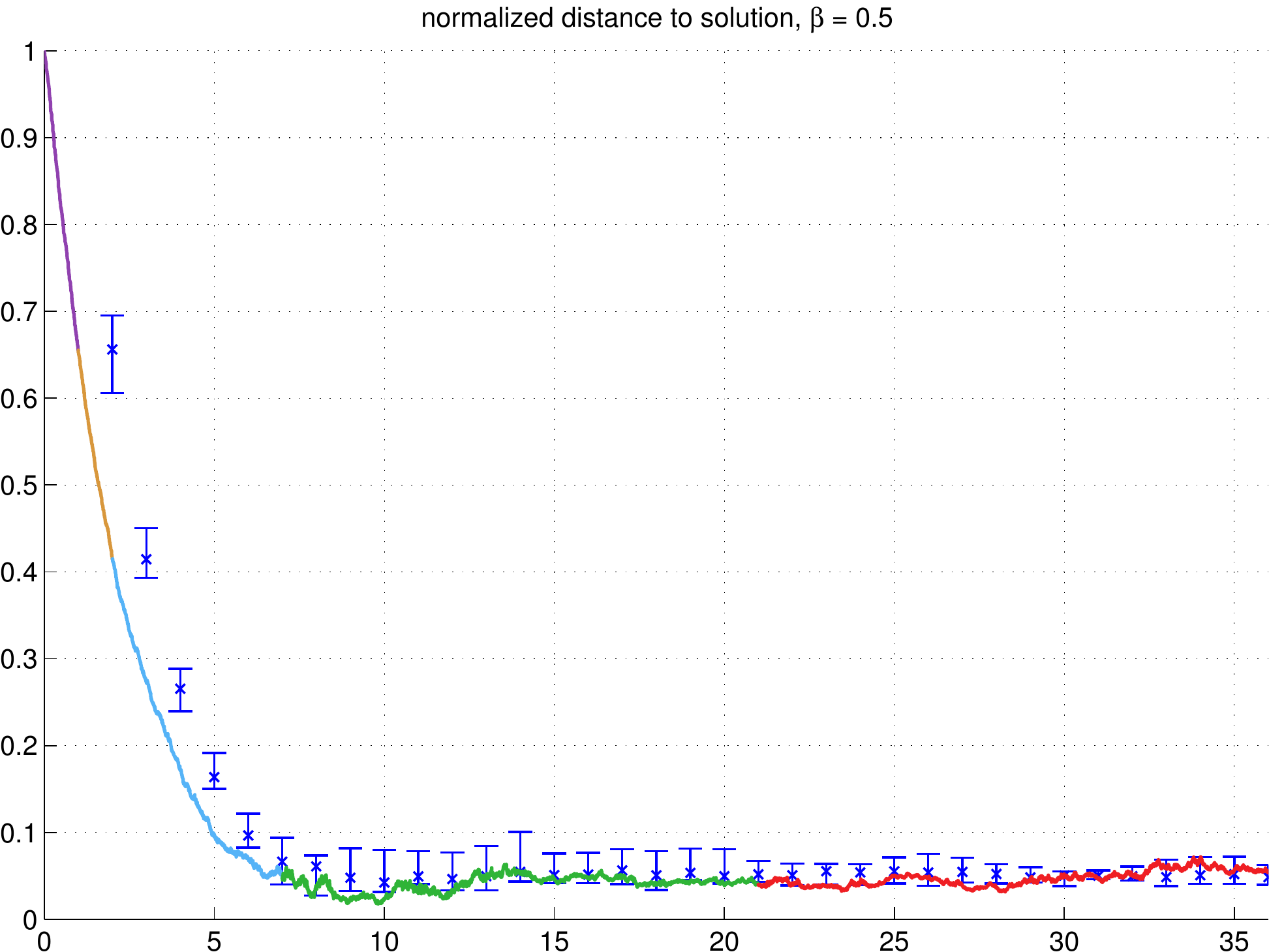} \qquad  
\includegraphics[width=0.42\linewidth]{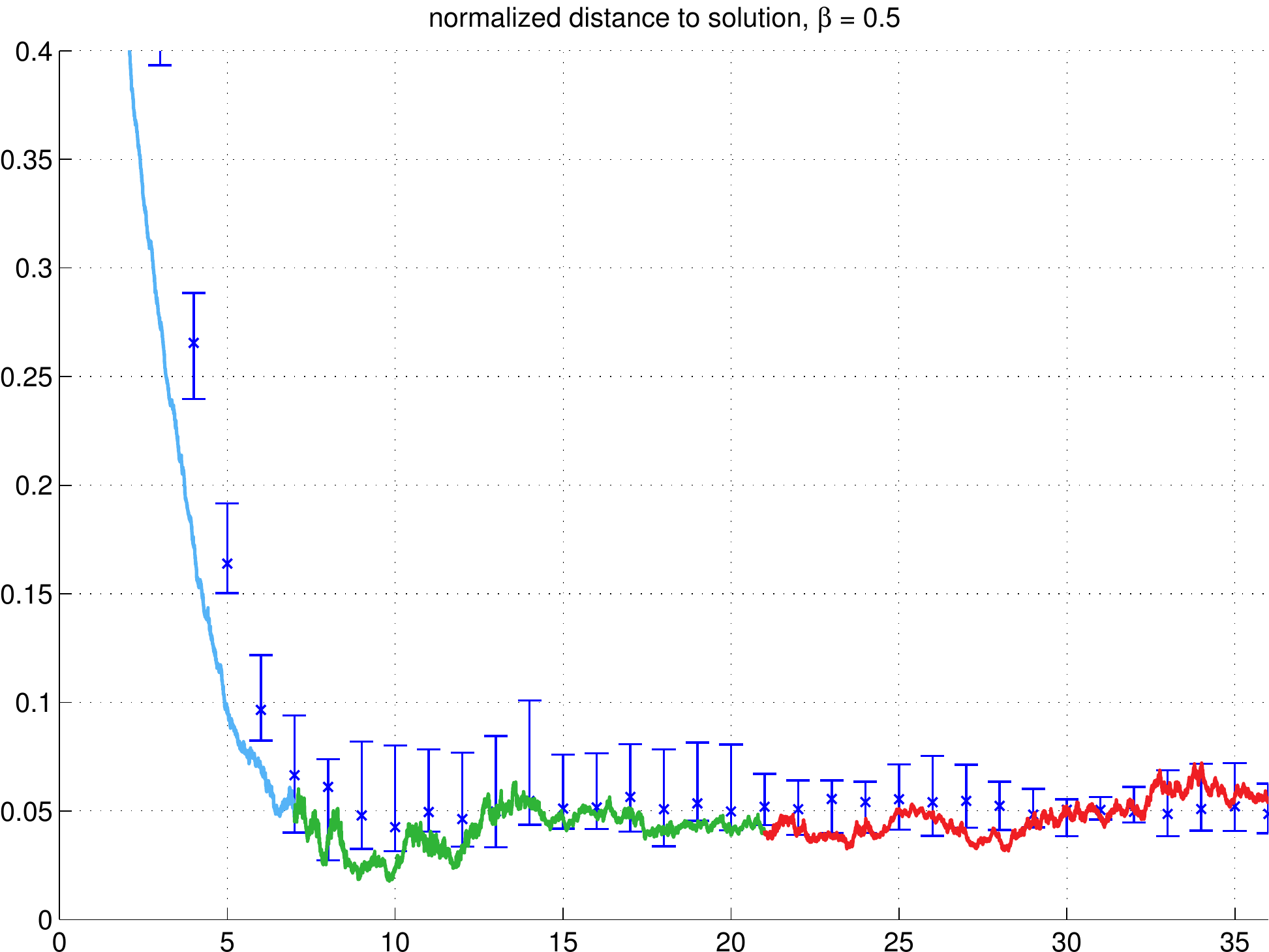}\\*[0.1cm] 
\includegraphics[width=0.42\linewidth]{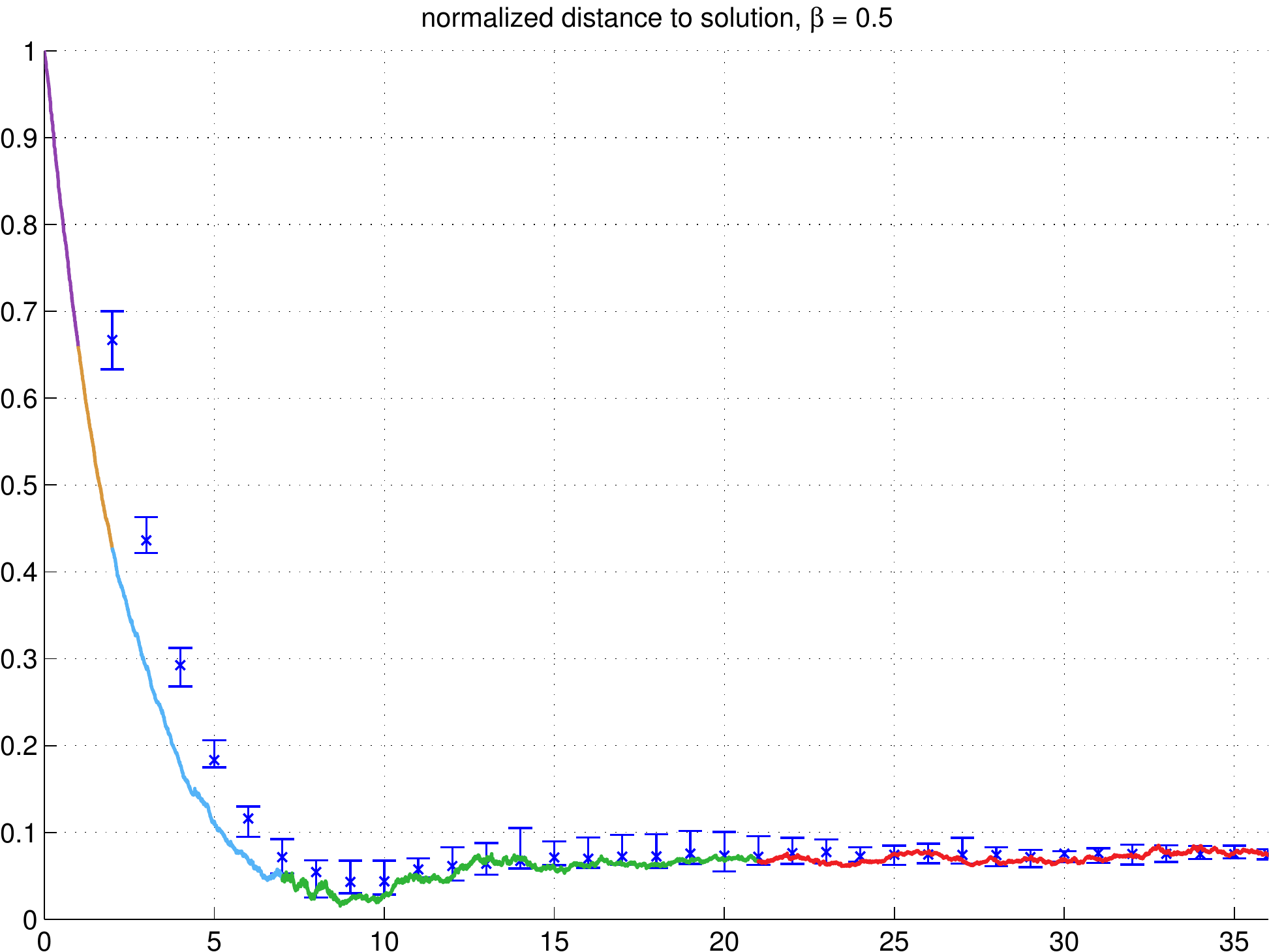} \qquad
\includegraphics[width=0.42\linewidth]{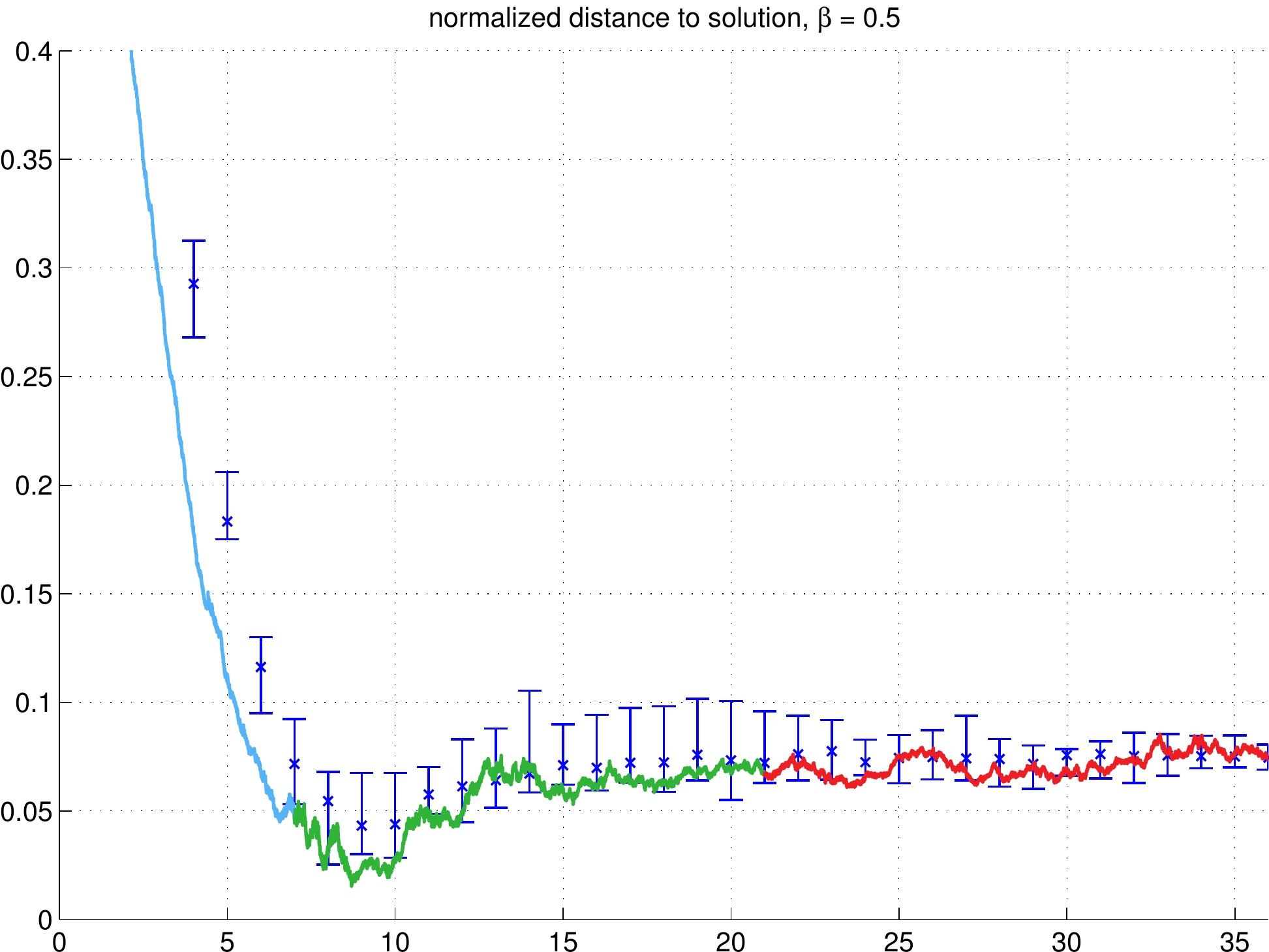} 
\caption{Variant I (top) and Variant II (bottom) with $\beta = 0.5$. The bottom portion of each plot on the left is enlarged and shown on the right. The solid curve corresponds to one run, with the $x$-component being $x=\sum_{k=0}^t \alpha_k$ and the $y$-component being the normalized distance of $\theta_t$ to $\theta^*$. (See the text for more details.)}\label{fig-dim-ex2c}
\end{figure}

In the next experiment, we did $10$ independent runs of $1.6 \times 10^6$ iterations each, for the two variant algorithms using two stepsize rules:
$$ \alpha_t = \frac{1}{2000 + (10 \, t)^\beta} \ \ \  \text{for} \ \beta = 0.7, \qquad \text{and} \quad \alpha_t = \frac{1}{2000 + (4000 \, t)^\beta} \ \ \ \text{for} \ \beta = 0.5. $$
As before we chose these stepsize rules to ensure that the stepsize becomes small enough later in the run (at $t=1.6 \times 10^6$, $\alpha_t$ is about $10^{-5}$ in both cases of $\beta$). The simulation results are plotted in Figure~\ref{fig-dim-ex2b} and Figure~\ref{fig-dim-ex2c} for $\beta=0.7$ and $\beta=0.5$ respectively.
The graphical objects in these figures have the same meanings as those in Figures~\ref{fig-dim-ex1b}-\ref{fig-dim-ex1c} for Problem~I, so we will only describe these objects briefly here.

In Figures~\ref{fig-dim-ex2b}-\ref{fig-dim-ex2c}, we plotted in solid lines the normalized distances (to $\theta^*$) of the iterates from one of the $10$ runs. The $x$-axis represents a continuous timeline, and a solid curve is made up of points $\big(\sum_{k=0}^t \alpha_k, |\theta_t - \theta^*|/|\theta^*| \big)$ from a single run of an algorithm. The whole curve is plotted on the left side of each figure, with a close-up view of its bottom portion shown on the right side.
We colored segments of the curves in different colors according to the range of stepsizes in each segment as follows:
$\alpha_t \in (0.0005, 0.0002]$ (purple), $\alpha_t \in (0.0002, 0.0001]$ (brown), 
$\alpha_t \in (0.0001, 0.00005]$ (blue), $\alpha_t \in (0.00005, 0.00002]$ (green),
$\alpha_t < 0.00002$ (red).

The blue error bars in the figures show the range of the maximal deviation from $\theta^*$ for multiple consecutive iterates from the $10$ experimental runs. They are formed in the same way as described in the earlier experiment for Problem I (see the descriptions for Figures~\ref{fig-dim-ex1b}-\ref{fig-dim-ex1c} in the previous subsection).  
The point with an `$\times$' mark inside the $x$-th error bar is the median, and the lower and upper ends of the error bar are the minimum and maximum, respectively, of the $10$ values of $\max_{\text{$x$-th segment}} |\theta_t - \theta^*|/|\theta^*|$ obtained from the  $10$ independent experimental runs.

The simulation results shown in Figures~\ref{fig-dim-ex1b}-\ref{fig-dim-ex1c} can be compared with the assertion in Theorem 3.3 of \cite{etd-wkconv} for the two variant algorithms with diminishing stepsizes.

\section{Mountain Car} \label{sec-mountaincar}

In this last set of experiments we test constrained ETD on a larger problem constructed from the Mountain Car problem \cite{SUB}. Mountain Car has continuous state and action spaces. As such it is actually beyond the finite-space model considered in \cite{etd-wkconv}, so the convergence theorems we proved therein for constrained ETD do not extend to the Mountain Car problem. Nevertheless, we observed empirically in our experiments that constrained ETD is well-behaved, and in this section we report some of these simulation results for Variant I with a constant stepsize. (Variant II behaves similarly but with a larger variance for this problem.)

\subsection{Experimental Setup}

We take the dynamics of the Mountain Car problem. The goal is to drive an underpowered car to reach the top of a steep hill.
A state consists of the position $p$ and velocity $v$ of the car, whose values are bounded as $p \in [-1.2, 0.5]$ and $v \in [-0.07, 0.07]$.
The position $0.5$ corresponds to the desired hill top destination, while the position $-\pi/6$ corresponds to the bottom of a valley. 
At each state three actions are available: $\{\texttt{back}, \texttt{coast},  \texttt{forward} \}$, designated by $\{-1, 0, 1\}$, respectively.
With $A_t$ denoting the action taken at time $t$ and with $\Pi_{[a,b]}(x) = \max \{a, \min\{b, x\}\}$ for an interval $[a,b]$ and scalar $x$, 
the dynamics of the car are defined as 
\begin{align*}
  v_{t+1} & =  \Pi_{[-0.07, \, 0.07]} \big( v_t + 0.001 A_t - 0.0025  \cos(3 p_t) \big), &
  p_{t+1} & = \Pi_{[-1.2, \, 0.5]} \big( p_t +v_{t+1} \big),
\end{align*}  
except that when $p_{t+1} = - 1.2$, the velocity is reset to zero: $v_{t+1}=0$. 
Before the destination $p=0.5$ is reached, the rewards depend only on the action taken and are given by $r(-1)=-1.5$, $r(1)=-1$, and $r(0)=0$. 
Once the destination $p=0.5$ is reached, the car enters a rewardless termination state permanently. 
We consider undiscounted expected total rewards, so the discount factor $\gamma=1$.

\begin{figure}[!t] 
   \centering
\includegraphics[width=0.9\linewidth]{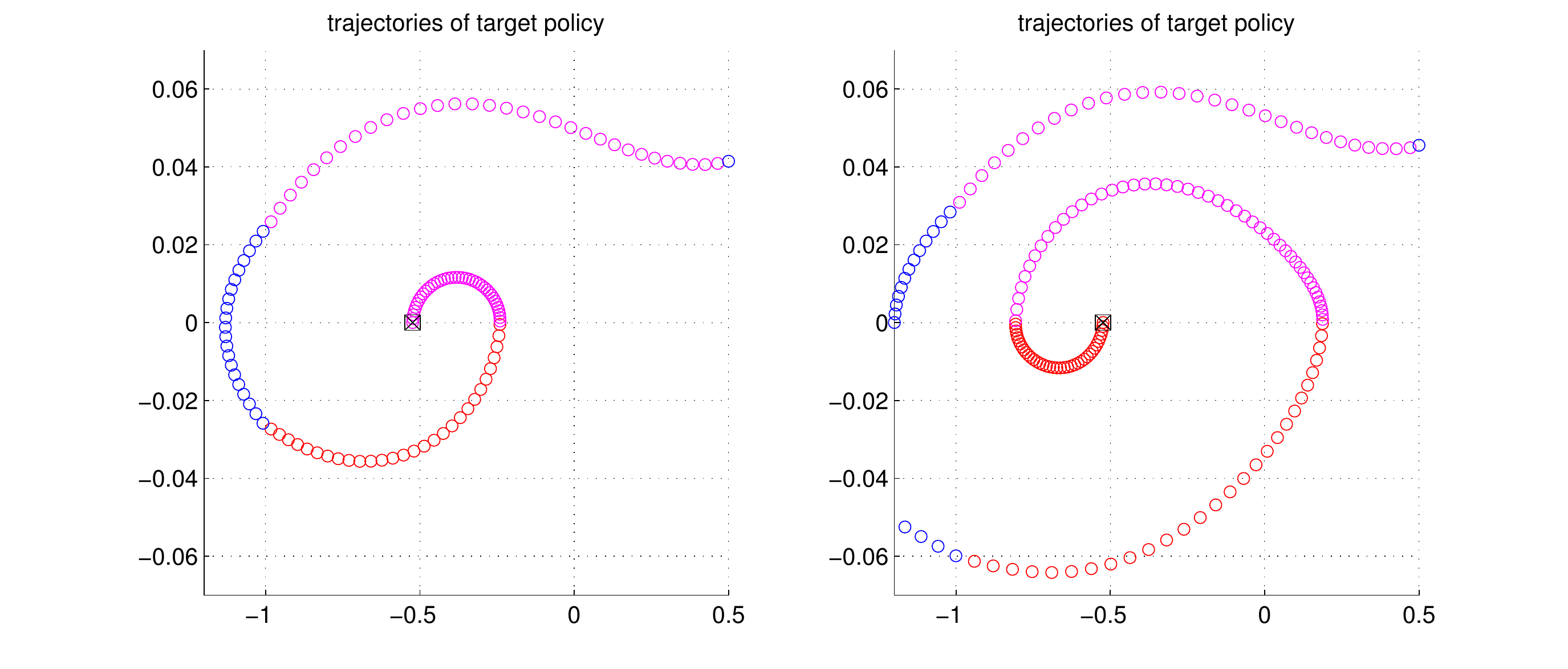}
\caption{Two state trajectories that can occur (with equal probability 0.5) under the target policy for the initial state $(-\tfrac{\pi}{6}, 0)$ (indicated by the marked square). The color of a state indicates the action taken at that state: red for \texttt{back}, blue for \texttt{coast}, and purple for \texttt{forward}.}\label{fig-policy}
\end{figure}

\medskip
\noindent {\bf Target policy:} 
The following policy will be our target policy $\pi$ throughout the experiments: at a state $(p,v)$,\vspace*{-0.2cm}
\begin{itemize}
\item if $p < -1$, then take action $0$ (coast);\vspace*{-0.1cm}
\item if $p \geq -1$, then take action $\text{sign}(v)$ unless $|v| \leq 10^{-6}$, in which case take action $\pm 1$ (forward or back) with equal probability.\vspace*{-0.2cm}
\end{itemize} 
This is a simple policy but behaves reasonably well. Figure~\ref{fig-valpolicy} (next page) shows the negative value function $-v_\pi$ and its contour map. The values of $v_{\pi}$ shown in this figure are estimated by simulating the policy $200$ times for each starting state $(p,v)$ in a set of  $171 \times 141$ points evenly spaced in the position-velocity space $[-1.2, 0.5] \times [-0.07, 0.07]$. In particular, the position (velocity) interval is evenly divided into subintervals of length $0.01$ ($0.001$). 
Figure~\ref{fig-policy} above shows the two trajectories that the car can traverse through in the state space if it is initially parked at the bottom of the valley ($p=-\pi/6$) and follows the target policy $\pi$.

{\samepage
\begin{figure}[!t] 
   \centering
\includegraphics[width=0.45\linewidth]{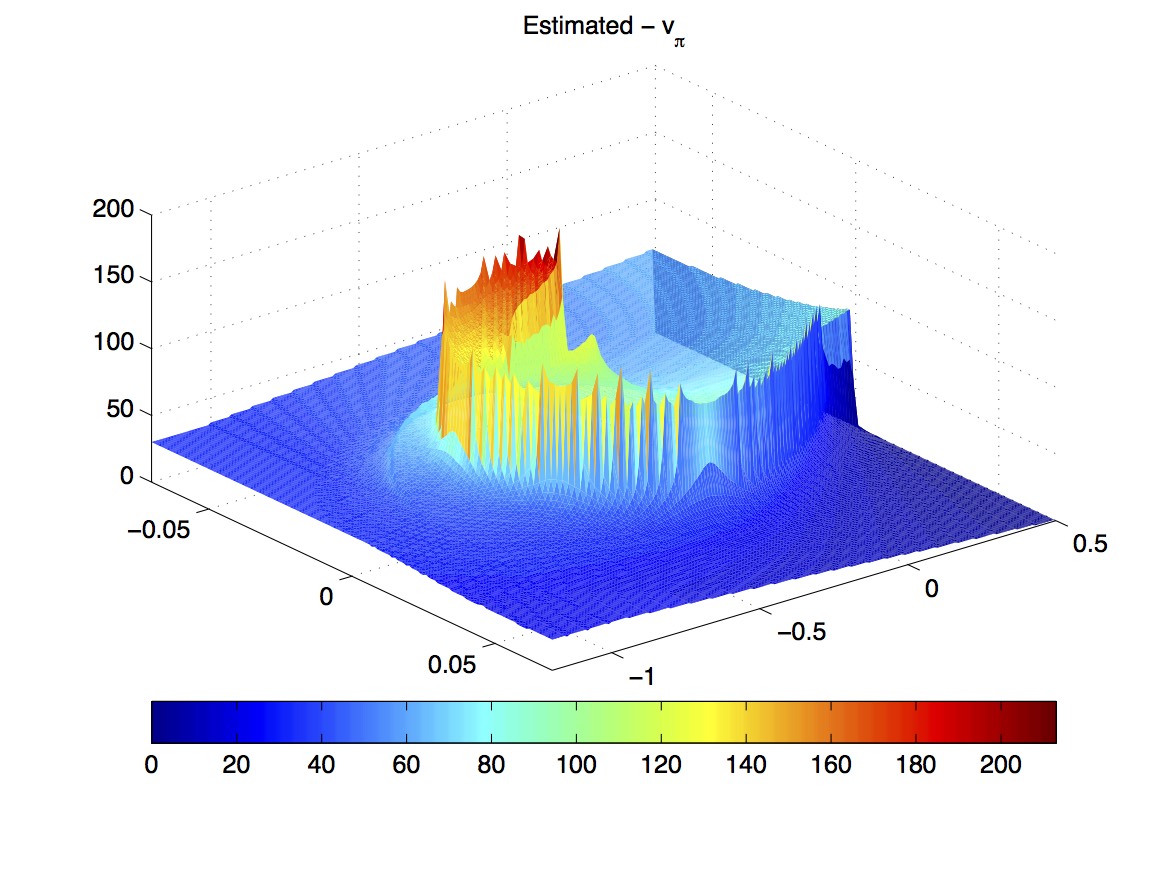} \hfill 
\raisebox{20pt}{\includegraphics[width=0.54\linewidth]{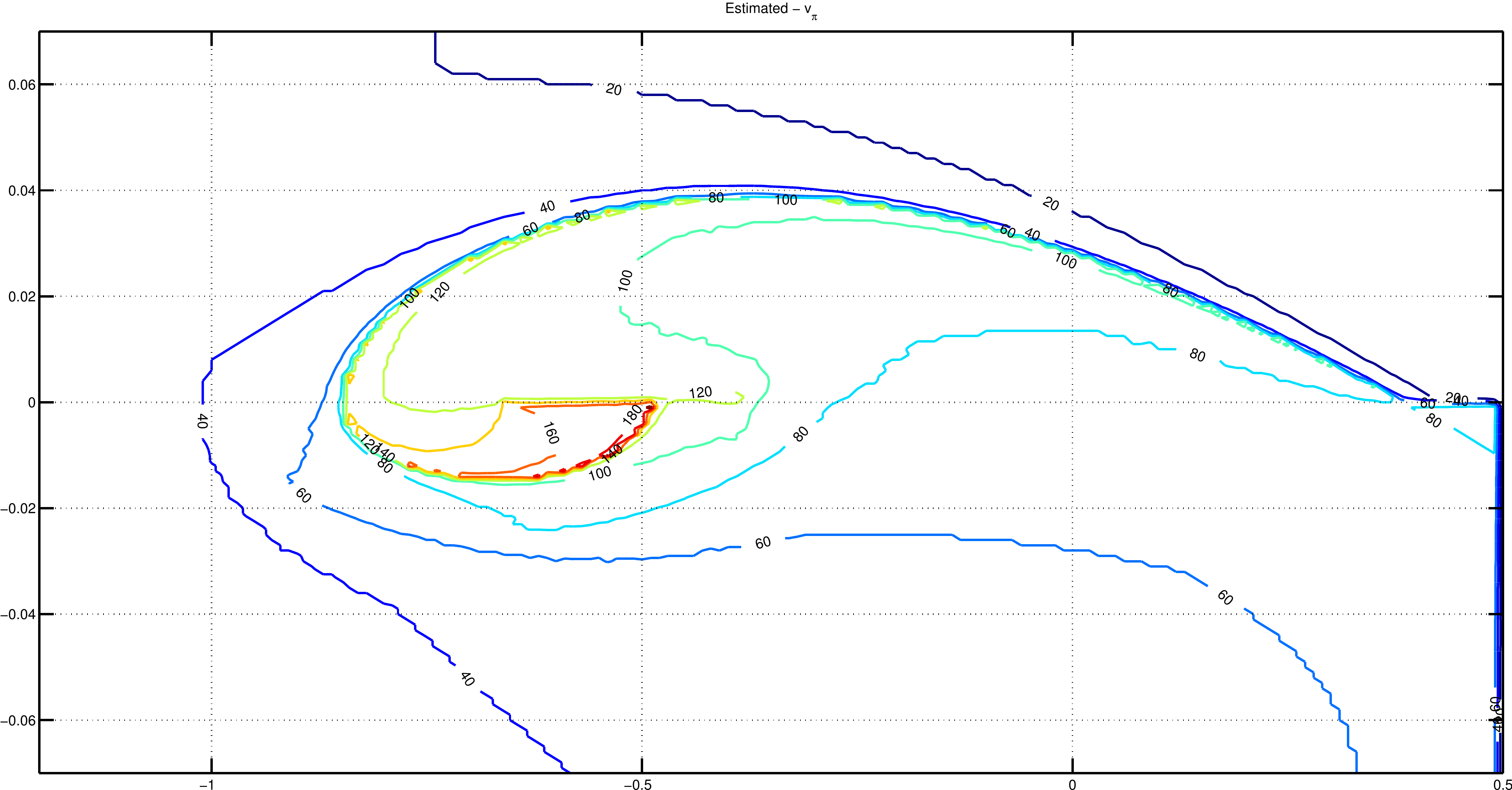}}
\caption{$-v_{\pi}$ estimated by simulating $\pi$ (left: 3D view; right: contour map).} \label{fig-valpolicy}
\end{figure}
\begin{figure}[!h] 
   \centering
\includegraphics[width=0.47\linewidth]{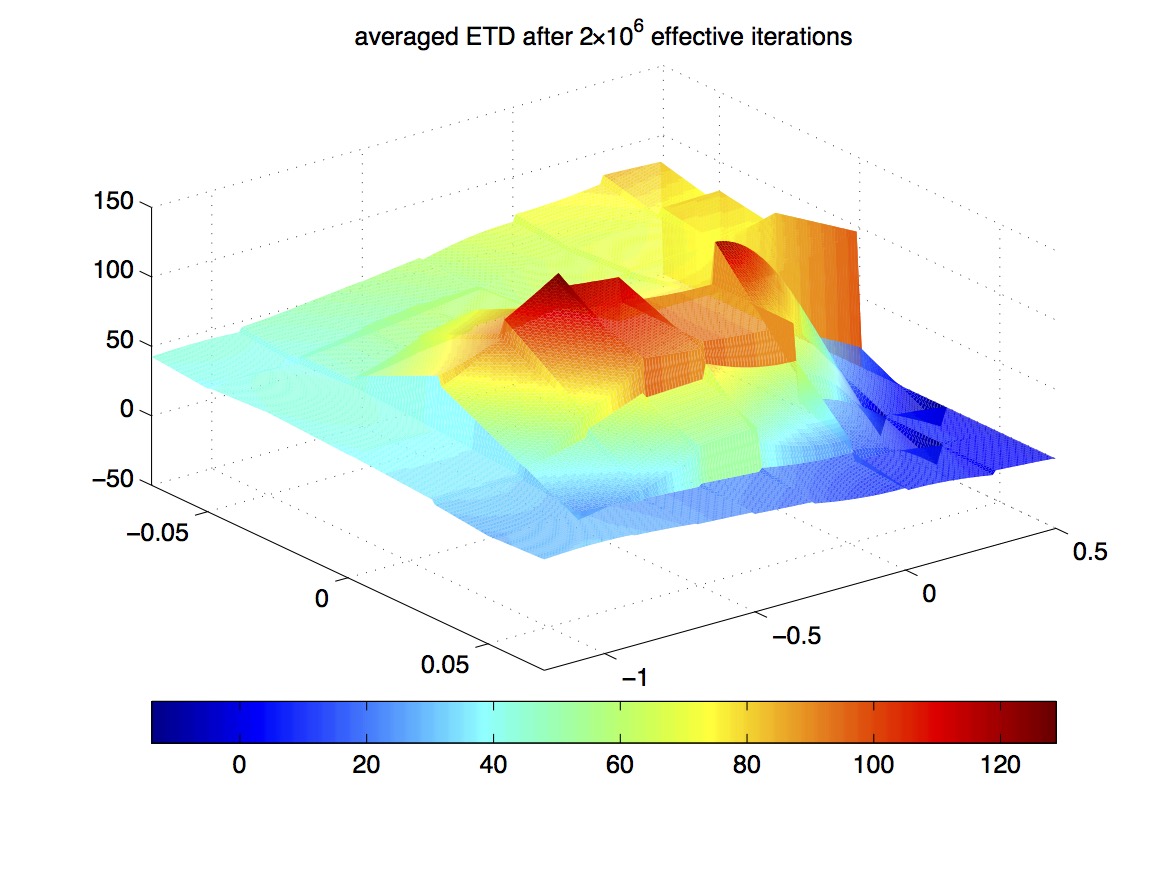} \quad
\includegraphics[width=0.47\linewidth]{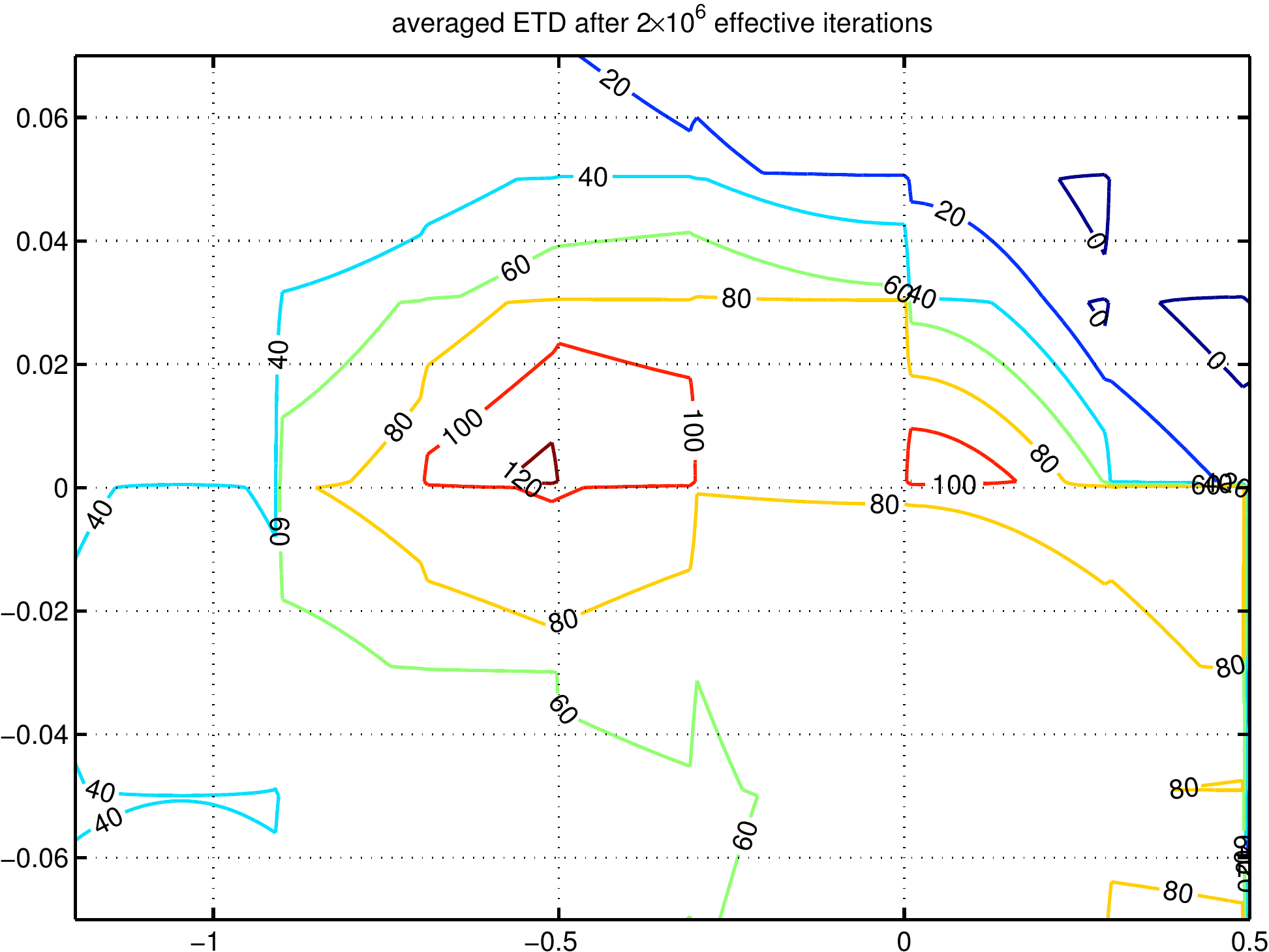}\\*[0.1cm]
\includegraphics[width=0.47\linewidth]{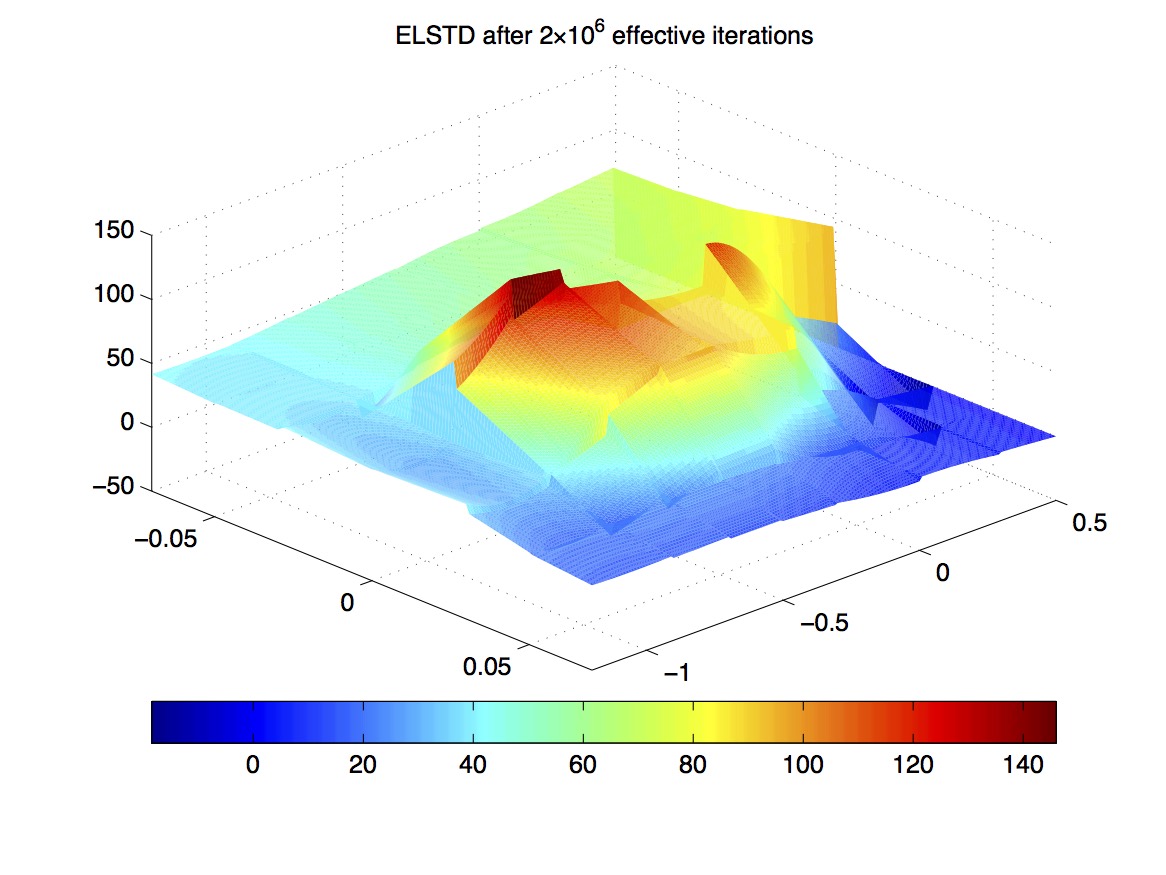} \quad
\includegraphics[width=0.47\linewidth]{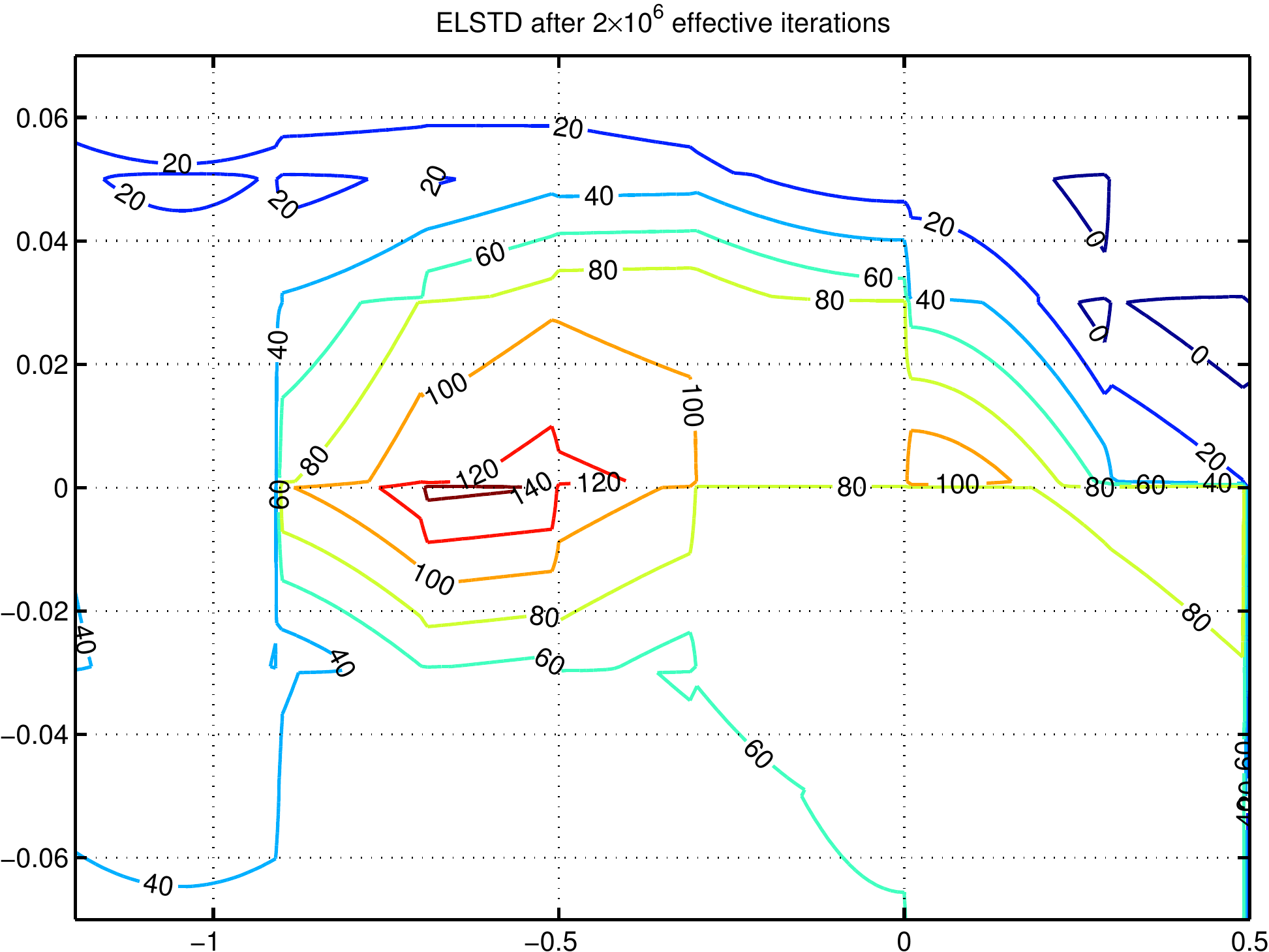}\\
\caption{Piecewise linear approximation of $-v_{\pi}$ (linear in $(\cos(3p), v)$ on each piece) calculated by Variant I (top) and  ELSTD (bottom). Left: 3D view; right: contour map.}\label{fig-car-ex1}
\end{figure}
}

\medskip
\noindent {\bf Behavior policy:} We use a fixed sampling scheme to generate states, actions and transitions for ETD learning. This scheme serves the role of the behavior policy $\pi^o$ and is defined as follows.\vspace*{-0.1cm}
\begin{itemize}
\item At a state $(p,v)$, if $p=0.5$ (the desired destination), then the next state is sampled uniformly from the state space $[-1.2, 0.5]\times[-0.07, 0.07]$.\vspace*{-0.2cm} 
\item For $p \not= 0.5$, two things can happen:\\
(1) With probability $0.9$, an action is chosen from the set $\{\texttt{back}, \texttt{coast},  \texttt{forward} \}$ randomly and uniformly, and the next state is determined by the state transition under that action.\\
(2) With probability $0.1$, a random state $(p', v')$ is chosen as the next state. In particular, either the velocity remains the same, $v'=v$, and the position $p'$ is uniformly sampled from the interval $[p, 0.5]$ or $[-1.2, p]$ (each of these two cases happens with probability $0.04$), or $(p',v')$ is uniformly sampled from the state space (this happens with probability $0.02$).\vspace*{-0.1cm}
\end{itemize}
The above scheme of generating data can be viewed as a valid behavior policy by enlarging the action space to include three more actions that correspond to the three different ways of randomly choosing $(p',v')$ described in step (2) above. This defines the importance sampling weights $\pi(s,a)/\pi^o(s,a)$ for the constrained ETD algorithms in the experiments.

It is worth mentioning that in the mathematical framework of off-policy learning, we need not restrict the behavior policy to be a physically feasible policy. Indeed, that would limit the use of off-policy learning in the goal-reaching type of problem such as Mountain Car, since in such problems, to find a policy that is able to reach the goal state can be tantamount to solving the problem itself. 
By defining the behavior policy in a broader way, one can apply off-policy learning methods to solving goal-reaching problems, at least in the context where the system dynamics of the problems can be simulated.

\medskip
\noindent {\bf Algorithmic parameters:} We will only show results for Variant I with a constant stepsize, as mentioned earlier.
The following algorithmic parameters are used throughout the experiments: 
a constant interest weight $0.5$ and a constant $\lambda = 0.5$ for all states; stepsize $\alpha = 0.003$; 
the radius parameter $r_\H = 2 \times 10^4$ for constraining $\theta$; and the truncation function $\psi_K$ with $K=50$ as given before. 
Since our purpose here is only to demonstrate that ETD can be applied beyond synthetic small problems, we did not optimize over these parameters. 
The stepsize we used is relatively large. As in the previous sections, we find that the use of a larger stepsize can make the algorithm progress faster initially, and together with averaging, it can yield useful approximation results in fewer iterations.

\subsection{Simulation Results}

\noindent {\bf First experiment:}
It can be seen from Figure~\ref{fig-valpolicy} (previous page) that $v_\pi$ is discontinuous and can change sharply between certain regions of the state space.
In the first experiment, we partition the position (and velocity) interval into $7$ (and $6$) subintervals to 
form $42$ rectangular regions to cover the space $[-1.2, 0.5) \times [-0.07, 0.07]$. 
In each region we approximate $v_{\pi}$ by an affine function of $(\cos(3p), v)$; the entire approximation is thus piecewise linear in $(\cos(3p), v)$. 
Specifically, to partition the position interval $[-1.2, 0.5)$ and the velocity interval $[-0.07, 0.07]$, 
we use the points given in the two vectors below as the mid points:
$$ \text{position:} \ \ (-0.9 \  -0.7  \  -0.5  \  -0.3  \  \ 0  \  \  0.3), \qquad  \text{velocity:} \ \    (-0.05 \ -0.03 \ \ 0 \ \ 0.03 \ \, 0.05).$$
Each of the $42$ regions is the product of two left-closed right-open intervals, except at the boundary of the state space, where the end points of an interval can be included or excluded in order to fill exactly the space $[-1.2, 0.5) \times [-0.07, 0.07]$. For each region, we used these $3$ features, $1$, $\cos(3p)$, and $15v$, to approximate the value function in that region.

The approximate value function obtained by Variant I after a single run of $2 \times 10^6$ effective iterations is shown in the top row of Figure~\ref{fig-car-ex1} (previous page).
Here an \emph{effective iteration} refers to an iteration in which the behavior policy takes an action that could also be taken by the target policy.
Plotted is the approximation corresponding to the averaged iterate $\bar\theta_t^\alpha$ at the end of the run, where the average is taken over the last $10^6$ iterations to avoid transient effects. 
We also ran ELSTD (modified as before) in the same run for comparison, and the approximate value function it obtained is plotted in the bottom row of Figure~\ref{fig-car-ex1}.  
It can be seen that both algorithms try to approximate $v_{\pi}$, and overall the contours of their approximations roughly match the contour of $v_{\pi}$ in shape (cf.~Figure~\ref{fig-valpolicy}). 

\begin{figure}[!htb] 
   \centering
\includegraphics[width=0.45\linewidth]{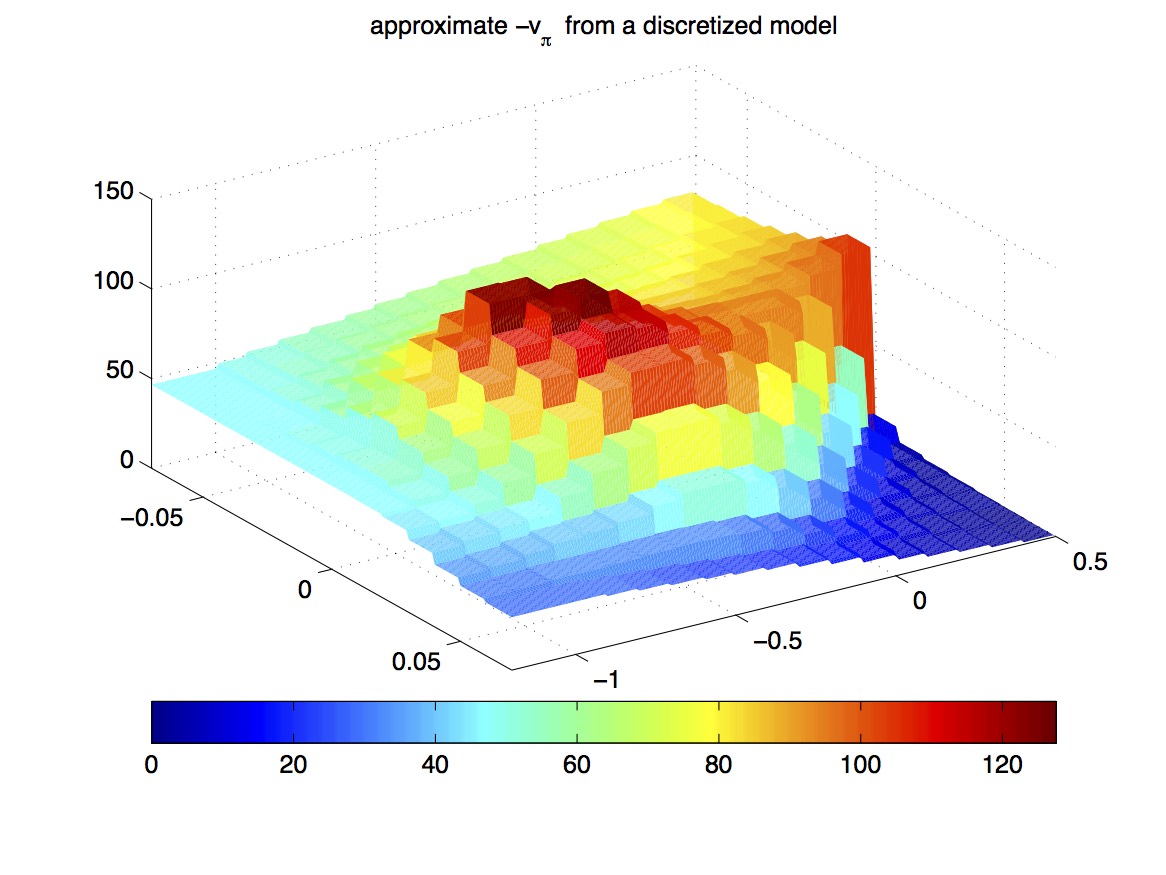} \hfill 
\raisebox{20pt}{\includegraphics[width=0.54\linewidth]{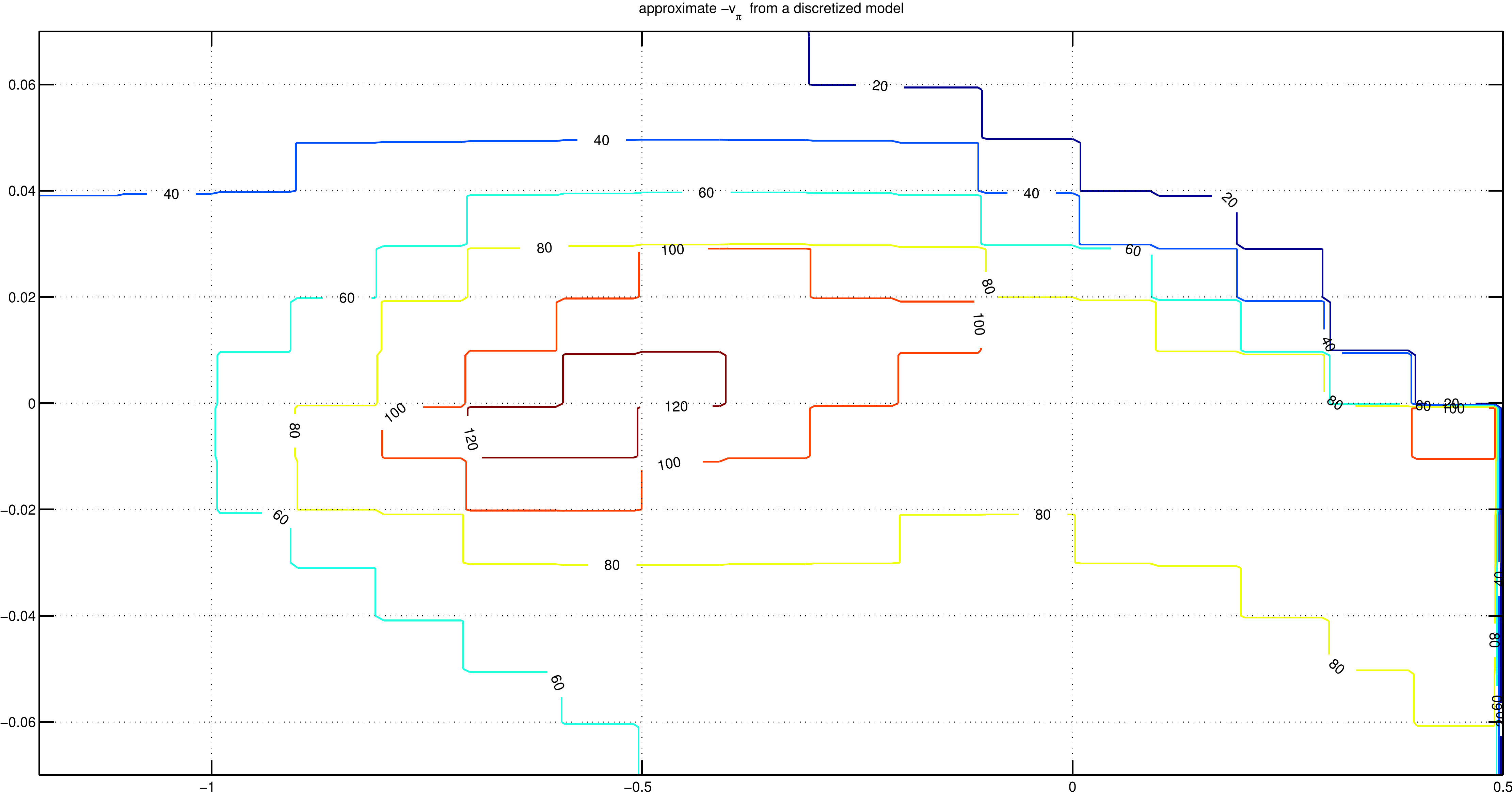}}
\caption{Approximate $-v_{\pi}$ from a discretized model (left: 3D view; right: contour map).}\label{fig-car-mdsol}
\end{figure}

\begin{figure}[!htb] 
   \centering
\includegraphics[width=0.47\linewidth]{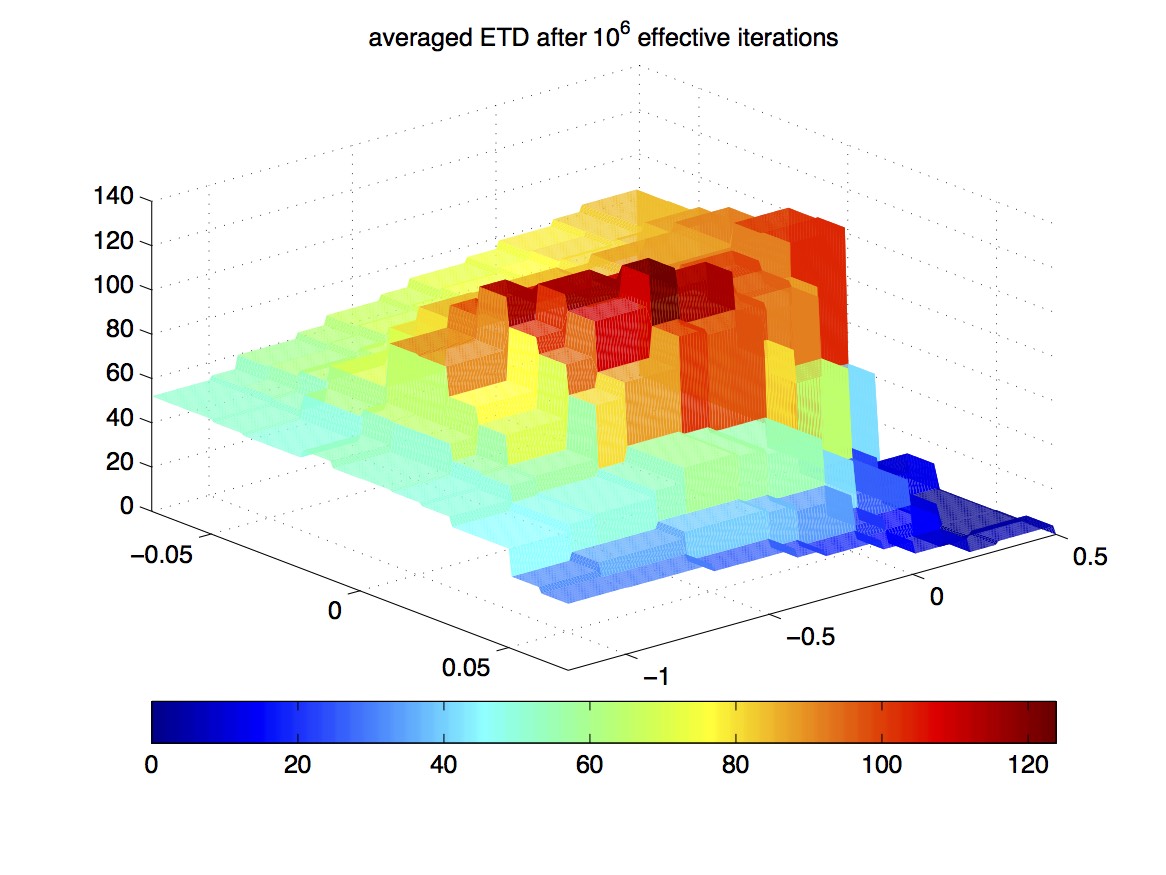} \quad
\includegraphics[width=0.47\linewidth]{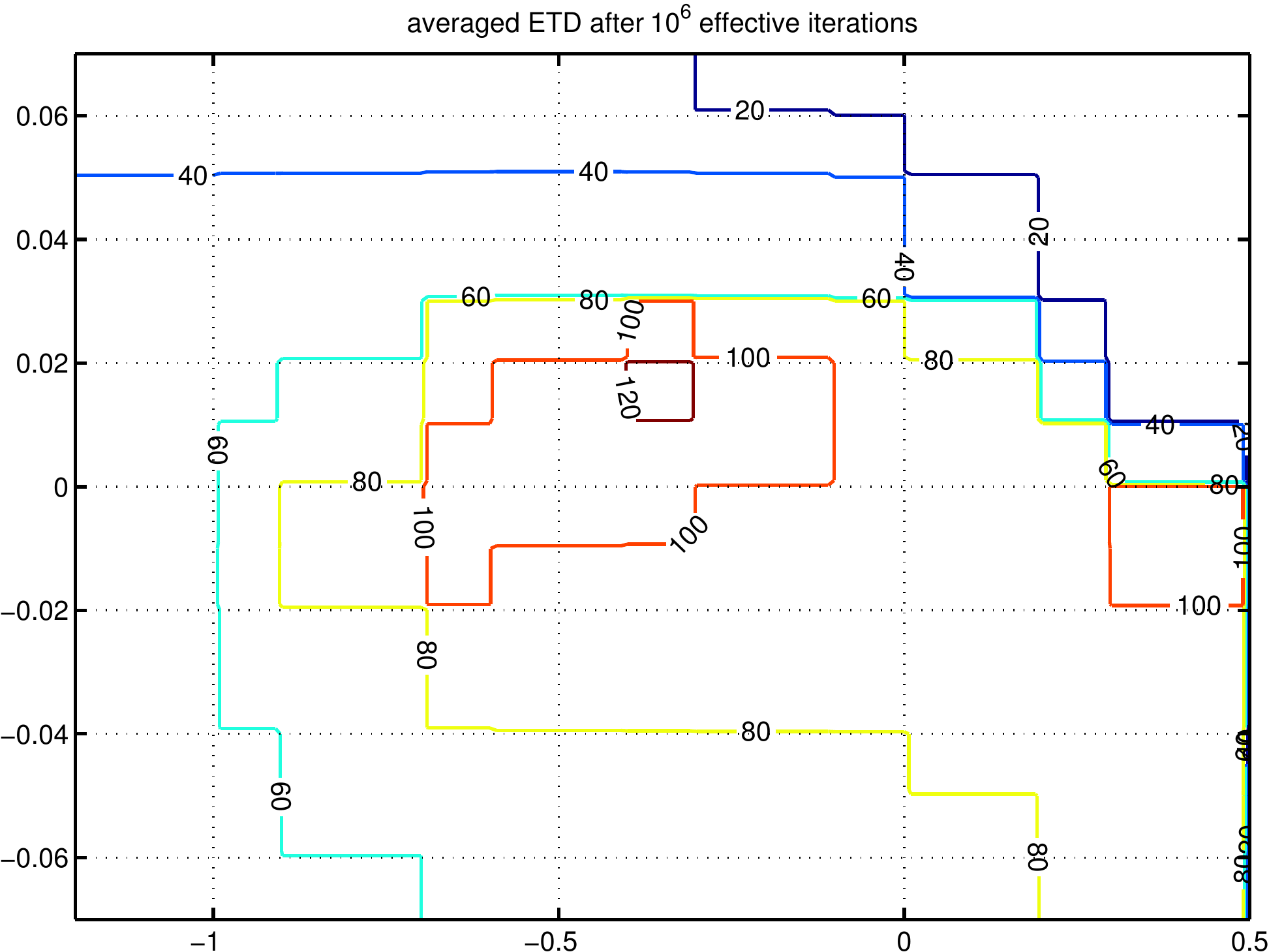}\\*[0.2cm]
\includegraphics[width=0.47\linewidth]{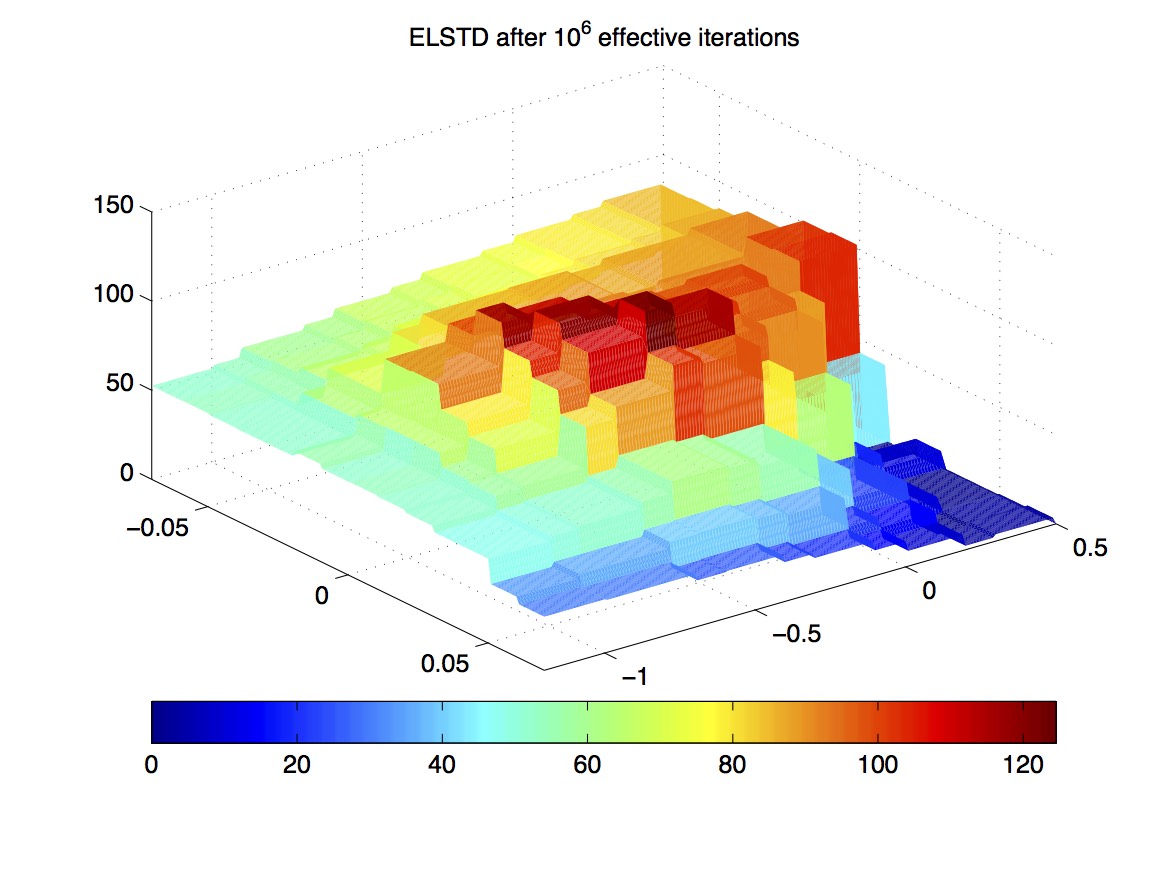} \quad
\includegraphics[width=0.47\linewidth]{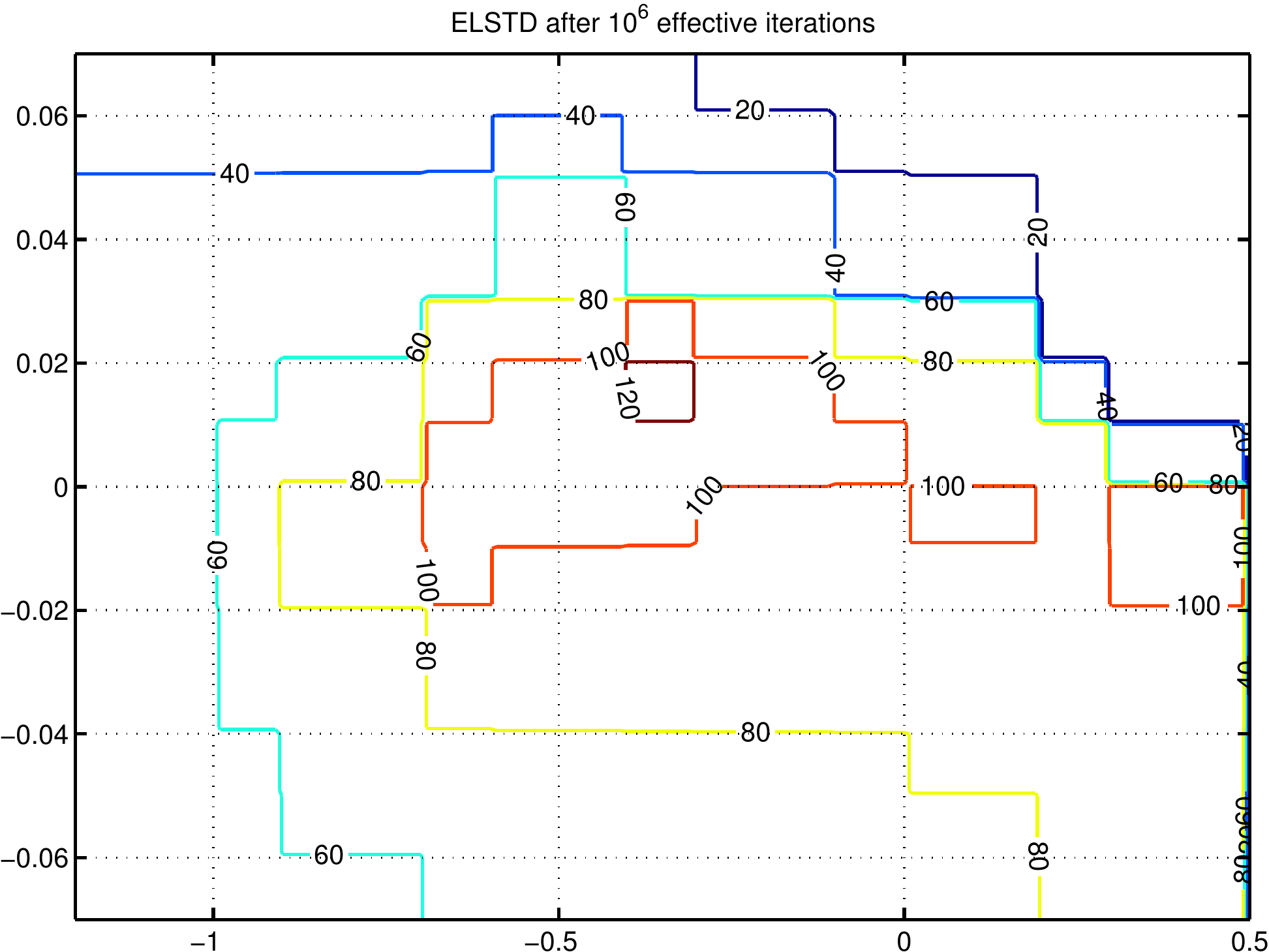}\\
\caption{Piecewise constant approximation of $-v_{\pi}$ calculated by Variant I (top) and  ELSTD (bottom) using tile-coding. Left: 3D view; right: contour map.}\label{fig-car-ex2}
\end{figure}

\begin{figure}[!htb] 
   \centering
\includegraphics[width=0.47\linewidth]{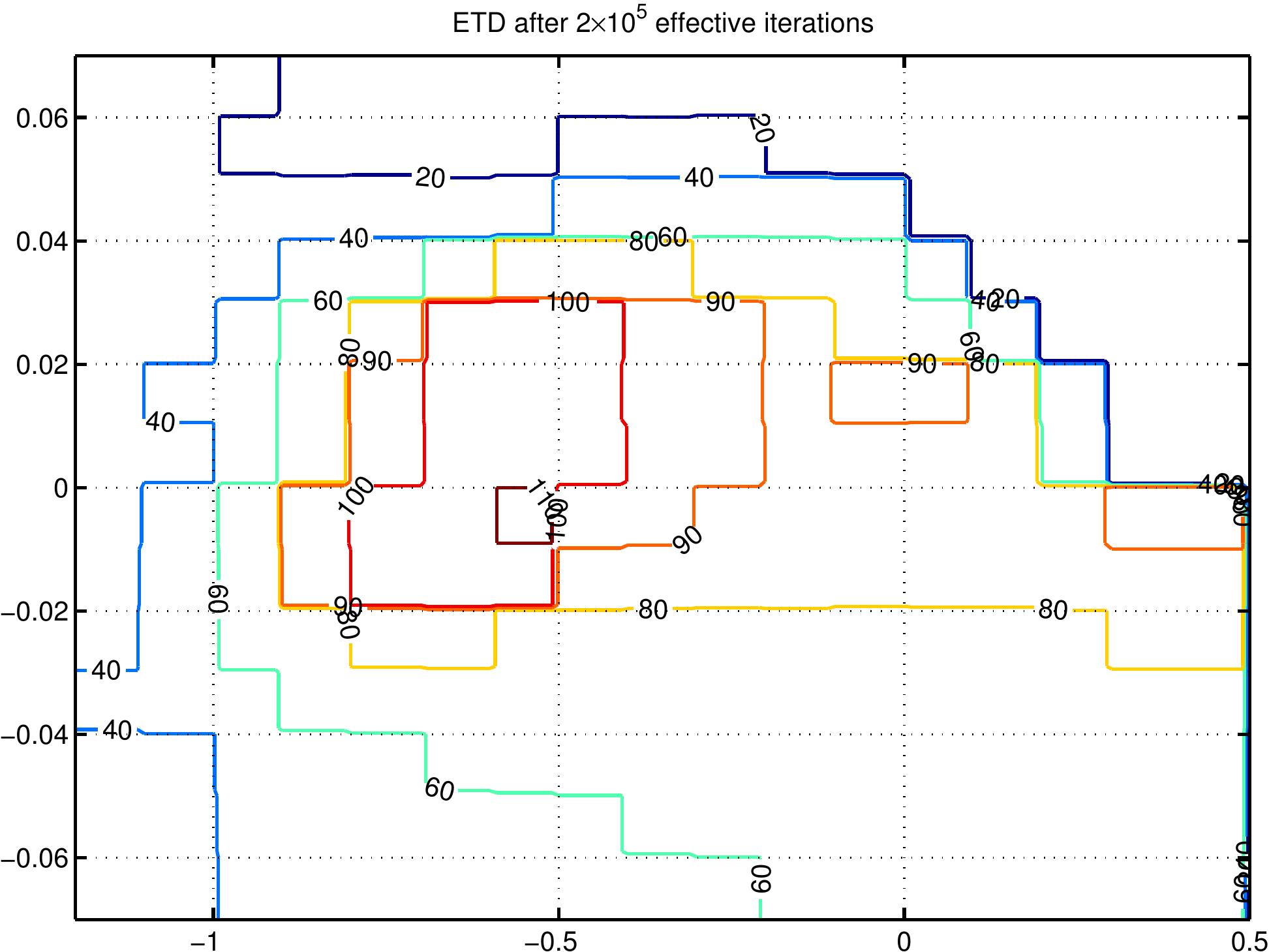} \quad
\includegraphics[width=0.47\linewidth]{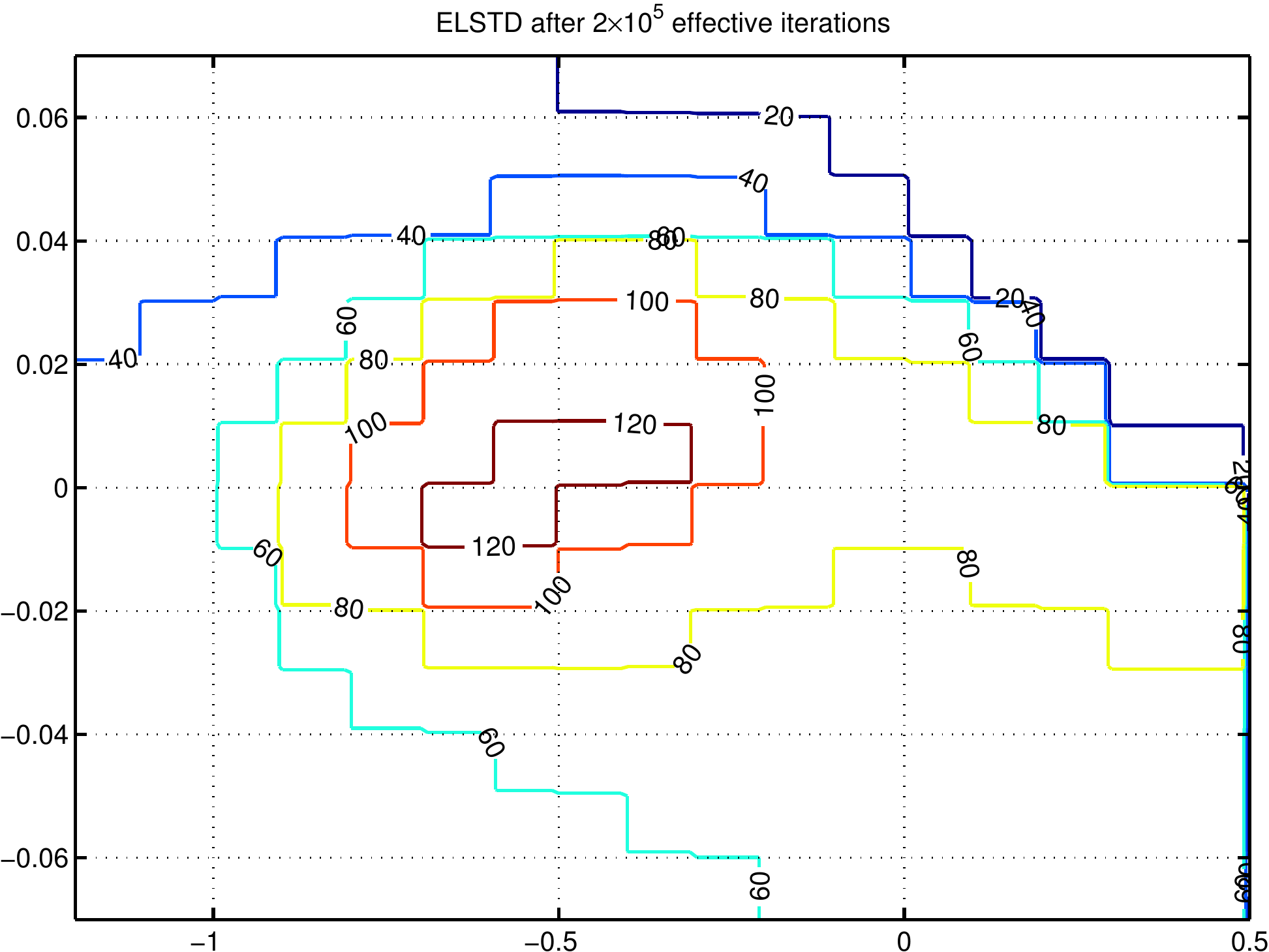}\\
\caption{Piecewise constant approximation of $-v_{\pi}$ calculated by Variant I (left) and  ELSTD (right) using tile-coding.}\label{fig-car-ex3a}
\end{figure}

\begin{figure}[!htb] 
   \centering
\includegraphics[width=0.47\linewidth]{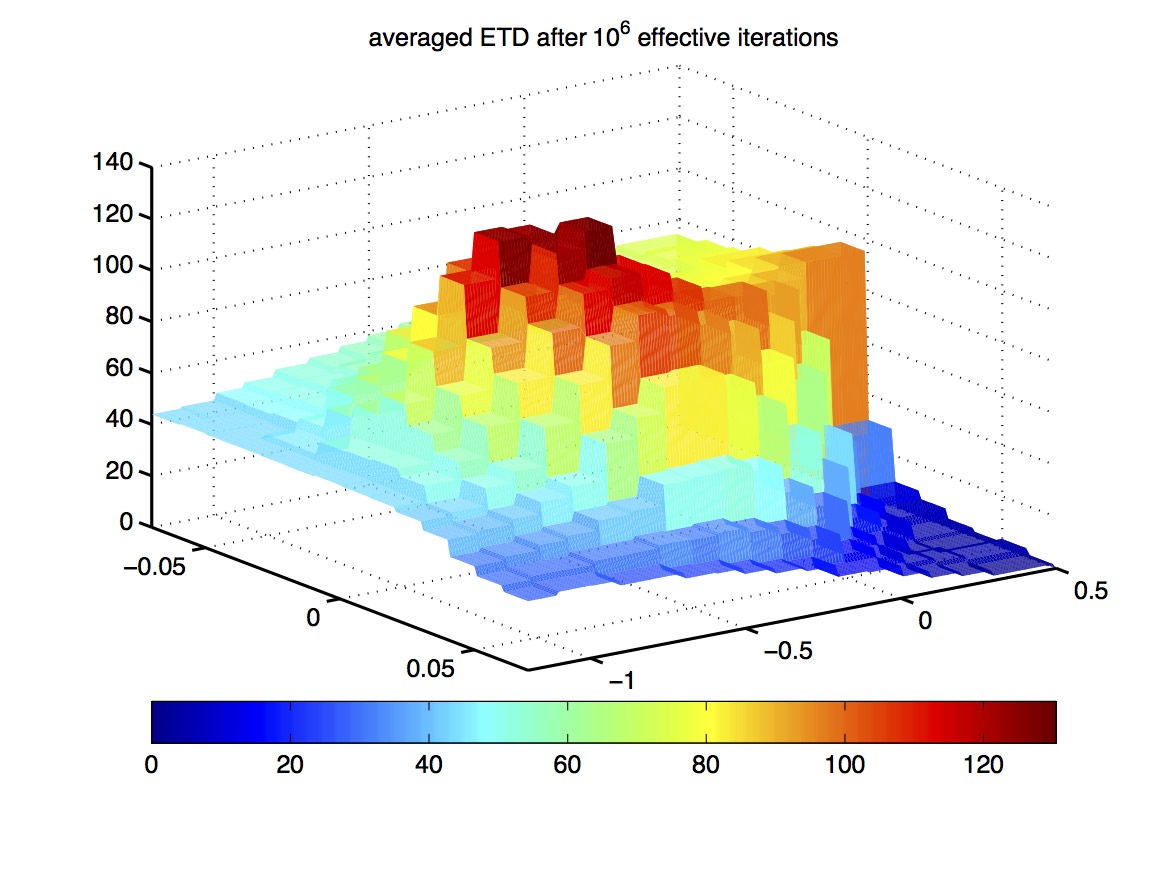} \quad
\includegraphics[width=0.47\linewidth]{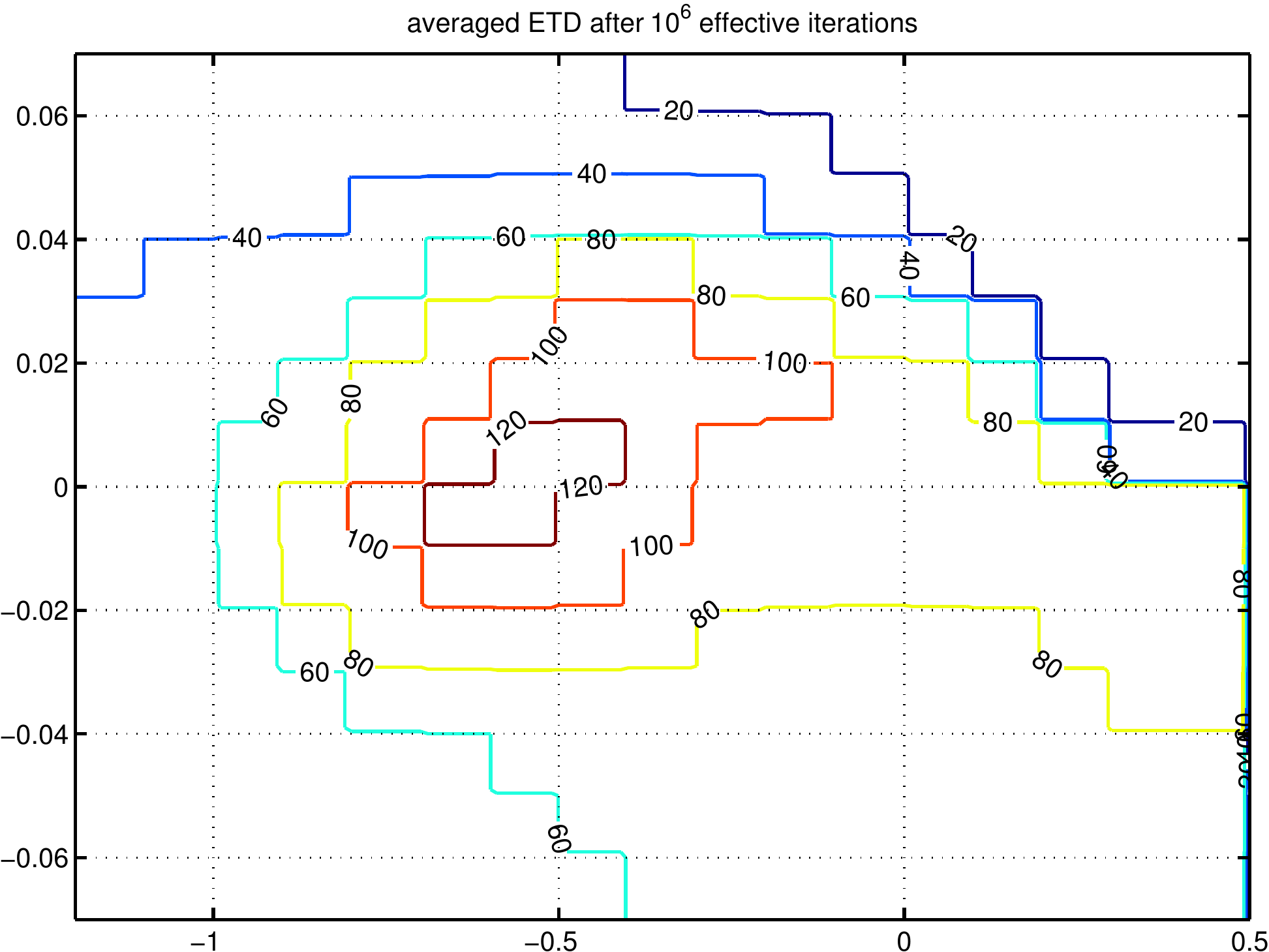}\\*[0.2cm]
\includegraphics[width=0.47\linewidth]{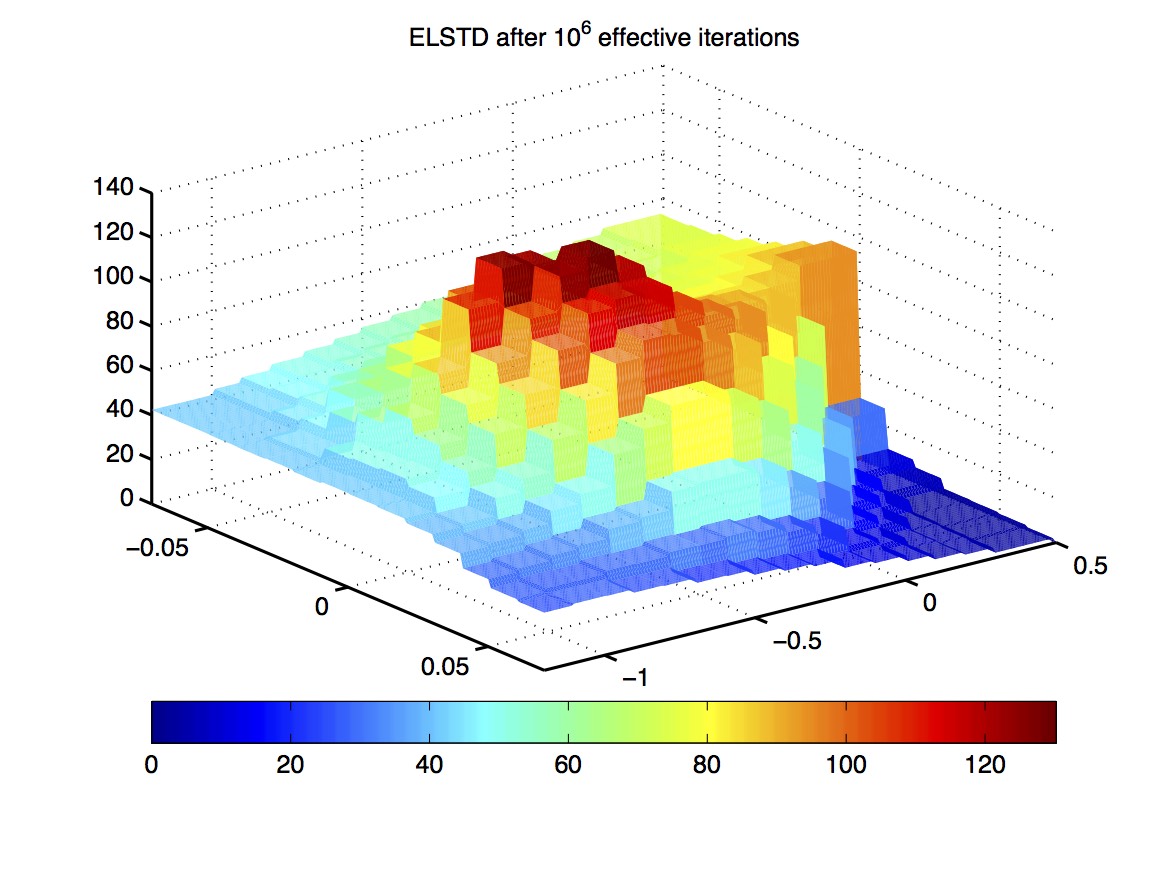} \quad
\includegraphics[width=0.47\linewidth]{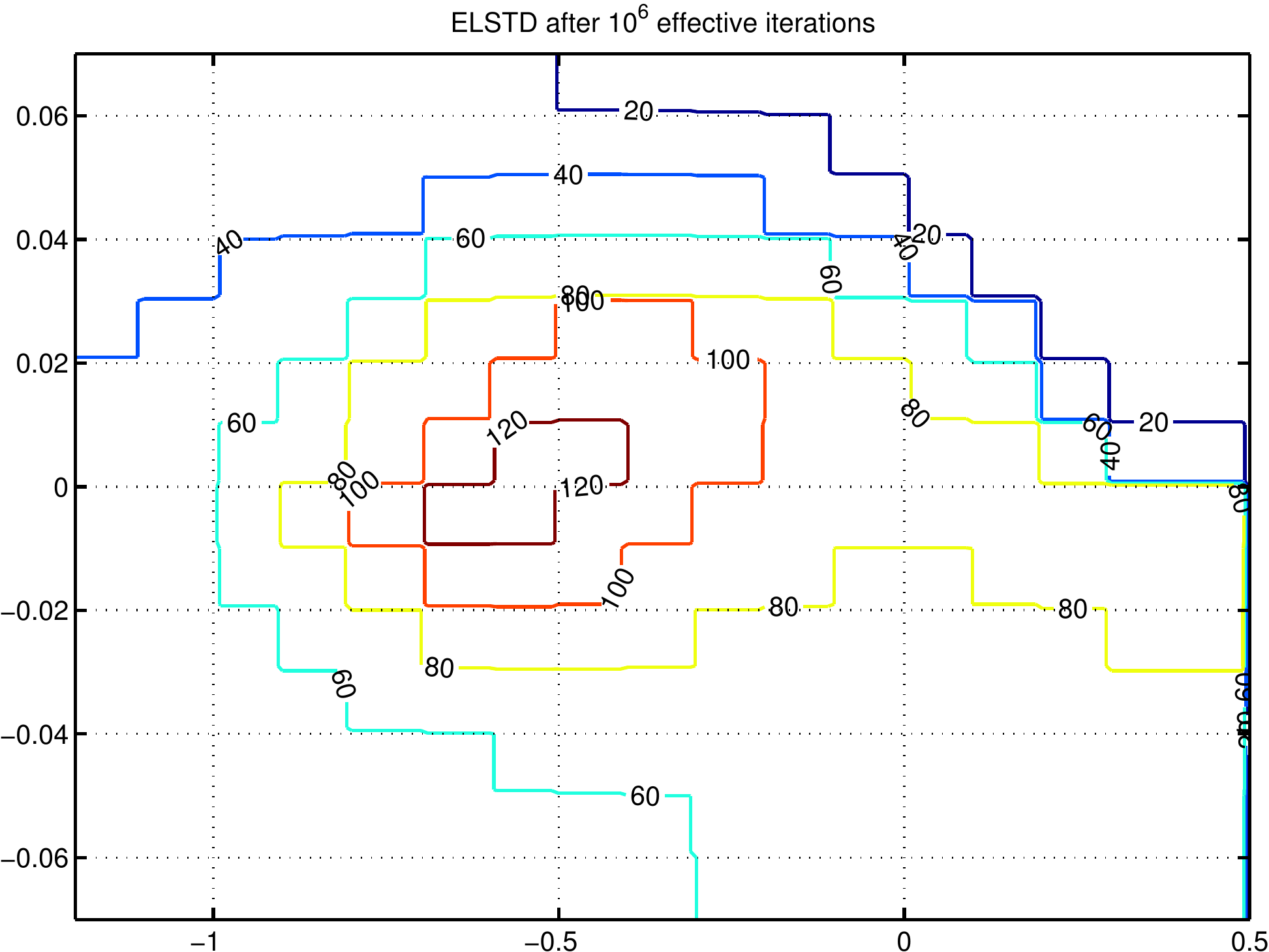}\\
\caption{Piecewise constant approximation of $-v_{\pi}$ calculated by Variant I (top) and  ELSTD (bottom) using tile-coding. Left: 3D view; right: contour map.}\label{fig-car-ex3b}
\end{figure}

\medskip
\noindent {\bf More experiments:} In the subsequent experiments we ran Variant I with features generated by tile-coding \cite{SUB}. This gives piecewise constant approximations of $v_{\pi}$, where the pieces are defined by the title-coding schemes we use. So instead of comparing the ETD approximations to $v_{\pi}$, which has jumps and curvy contours as shown in Figure~\ref{fig-valpolicy}, it seems better to compare the ETD approximations to an approximate solution from a discretized model that has a discretization resolution comparable to the resolution of the tile-coding schemes in the experiments.
Such a discretized model was constructed and it yields the approximate $v_{\pi}$ shown in Figure~\ref{fig-car-mdsol}.
The ETD approximation results for two tile-coding schemes are shown in Figures~\ref{fig-car-ex2}-\ref{fig-car-ex3b}. The details of these figures are as follows.

\medskip
\noindent Figure~\ref{fig-car-mdsol}: We first built a discretized finite-state model for the target policy as follows. 
We divided the position (velocity) interval evenly into subintervals of length $0.1$ ($0.01$), and thus obtained $270$ rectangular regions in total to fill the space $[-1.2, 0.5) \times [-0.07, 0.07]$. (As in the first experiment, except at the boundaries of the state space, each of these regions is the product of two left-closed right-open intervals.) The states in each region is treated as one aggregate state in the discretized model. To define the transition probabilities between the aggregate states for the target policy, we ran the behavior policy (the sampling scheme described earlier) for $10^7$ iterations, and used those effective iterations to calculate the transition frequencies between the aggregate states. These frequencies are taken to be the transition probabilities in the discretized model, and the per-stage rewards for the model are defined similarly. The Bellman equation for the discretized model is then solved, and the solution is used to define a piecewise constant approximation of $v_\pi$ (constant over each aggregate state).
Figure~\ref{fig-car-mdsol} plots the 3D view and contour map of the resulting approximation, which may be compared with the estimated $v_{\pi}$ shown in Figure~\ref{fig-valpolicy}.

\medskip
\noindent Figure~\ref{fig-car-ex2}: In this experiment a coarse tile-coding scheme is used to generate $78$ overlapping rectangular regions in total to cover the state space. Specifically, a first tiling comprises of $36$ rectangles, which are obtained by dividing the position interval $[-1.2, 0.5)$ and the velocity interval $[-0.07, 0.07]$ unevenly at the following points:
$$ \text{position:} \ \ (-0.9 \ -0.6 \ -0.3 \ \ 0 \ \ 0.3), \qquad \text{velocity:} \ \ (-0.05 \ -0.02 \ \ 0 \ \ 0.02 \ \, 0.05).$$
A second tiling is similarly formed by dividing the position and velocity intervals at these points:
$$ \text{position:} \ \ (-1.0 \ -0.7 \ -0.4 \ -0.1 \ \, 0.2), \qquad  \text{velocity:} \ \ (-0.06 \ -0.04 \ -0.01 \ \, 0.01 \ \, 0.03 \ \, 0.06).$$
This tiling comprises of $42$ rectangles (it corresponds to offsetting the first tiling by $(0.2, 0.01)$ and then covering the exposed sides of the state space with extra rectangles).  
Correspondingly, we used $36+42=78$ binary features for each state, to indicate the two rectangles containing that state. (As before, each rectangle is taken to be the product of two left-closed right-open intervals except on the boundaries of the state space.)

We ran Variant I and ELSTD with these features for $10^6$ effective iterations. Plotted in the top row of Figure~\ref{fig-car-ex2} is the approximate value function corresponding to the averaged iterate $\bar \theta_t^\alpha$ produced by Variant I at the end of the run. The average here is taken over the last $5\times 10^5$ iterations to reduce transient effects. 
The approximation obtained by ELSTD is plotted in the bottom row of Figure~\ref{fig-car-ex2} for comparison.
The resolutions of the two tilings used in this experiment are lower than the resolution used to build the discretized model. Nevertheless, comparing Figure~\ref{fig-car-ex2} with Figure~\ref{fig-car-mdsol}, one can recognize the similarities between the ETD/ELSTD approximations here and the approximate $v_{\pi}$ from the discretized model shown.

\medskip
\noindent Figures~\ref{fig-car-ex3a}-\ref{fig-car-ex3b} (next page): In this experiment we ran Variant I with a finer tile-coding scheme. 
Similarly to the previous case, we made two tilings of the state space by dividing the position and velocity intervals unevenly, first at these points:
$$ \text{position:} \ \ (-1.0 \ -0.8 \ -0.6 \ -0.4 \ -0.2 \ \ 0 \ \ 0.2), \qquad \text{velocity:} \ \  (-0.05 \  -0.03 \ -0.01 \ \ 0 \ \ 0.01 \ \, 0.03 \ \, 0.05),$$
and then at these points:
\begin{align*}
   \text{position:} \ \ & (-1.1 \ -0.9 \ -0.7 \ -0.5 \ -0.3 \ -0.1 \ \, 0.1 \ \, 0.3), \\
    \text{velocity:} \ \ & (-0.06 \ -0.04 \ -0.02 \ \ 0 \ \ 0.01 \ \, 0.02 \ \, 0.04 \ \, 0.06).
\end{align*}    
The first (second) tiling comprises of $64$ ($81$) rectangles. 
Correspondingly, each state has $64+81=145$ binary features to indicate the two rectangles that contain the state.

We ran both Variant I and ELSTD with these features. 
Figure~\ref{fig-car-ex3a} shows the approximations obtained after $2 \times 10^5$ effective iterations.
As can be seen, Variant I is in the process of building up the approximate value function, while ELSTD generally converges faster.
Figure~\ref{fig-car-ex3b} shows the approximations obtained after $10^6$ effective iterations, 
where for Variant I (top row of Figure~\ref{fig-car-ex3b}), plotted is the approximate value function corresponding to the averaged iterate $\bar \theta_t^\alpha$ at the end of the run, with averaging taken over the last $5\times 10^5$ iterations as before to reduce transient effects.
It can be seen that the results from Variant I and ELSTD (bottom row of Figure~\ref{fig-car-ex3b}) are now much closer to each other than in Figure~\ref{fig-car-ex3a}. 
Furthermore, both approximations can be compared with the approximate solution from the discretized model shown in Figure~\ref{fig-car-mdsol}.

\clearpage

\bibliographystyle{apa}
\let\oldbibliography\thebibliography
\renewcommand{\thebibliography}[1]{%
  \oldbibliography{#1}%
  \setlength{\itemsep}{0pt}%
}
{\fontsize{9}{11} \selectfont
\bibliography{etdexp-log}
}

\end{document}